\documentclass[a4paper,11pt]{report}  

\usepackage[left=3.25cm,right=3.25cm,top=2.5cm,bottom=3.5cm]{geometry}  

\usepackage[margin=\the\parindent,small,bf,sf]{caption} 
\usepackage{graphicx}
\usepackage{pdfpages}
\usepackage[afrikaans,english]{babel}
\usepackage{microtype}
\usepackage{setspace}
\newcommand{\myemph}[1]{{\sffamily\bfseries#1}}
\usepackage[raggedright,sf,bf]{titlesec}
\titlelabel{\thetitle.\ }
\titleformat{\chapter}[display]{\huge\bfseries\sffamily}{\chaptertitlename\ \thechapter}{15pt}{\Huge \raggedright}
\titlespacing*{\chapter}{0pt}{0pt}{40pt}  

\makeatletter
\let\originall@chapter\l@chapter
\def\l@chapter#1#2{\originall@chapter{{\sffamily #1}}{#2}}
\makeatother
\let \savenumberline \numberline
\def \numberline#1{\savenumberline{#1.}}

\usepackage[cmex10]{amsmath}
\usepackage{amssymb}
\usepackage{cancel}
\DeclareMathOperator*{\argmax}{arg\,max}
\newcommand{\T}{^\textrm{T}}

\renewcommand{\vec}[1]{\boldsymbol{\mathbf{#1}}}
\newcommand{\defeq}{\triangleq}

\usepackage{booktabs}
\usepackage{tabularx}
\usepackage{multirow}
\newcommand{\mytable}{
    \centering
    \small
    \renewcommand{\arraystretch}{1.2}
    }

\newcolumntype{C}{>{\centering\arraybackslash}X}
\newcolumntype{L}{>{\raggedright\arraybackslash}X}

\usepackage{fancyhdr}
\fancypagestyle{plain}{\fancyhead{}
                       
                       \fancyfoot[C]{\thepage}}

\usepackage{algorithm}  

\usepackage{hyperref}
\hypersetup{colorlinks=true,linktoc=all,citecolor=black,linkcolor=black}
\usepackage[nottoc]{tocbibind}

\usepackage{algpseudocode}  
 
\usepackage{natbib}
\title{\sffamily \bfseries Unsupervised neural and Bayesian models for zero-resource speech processing}
\author{Herman Kamper}
\date{2016}

\usepackage{xcolor}
\definecolor{orange}{HTML}{FF6600}
\newcommand{\edit}[1]{#1}

\begin{document}

\graphicspath{{frontmatter/fig/}}
\pagenumbering{Alph}

\begin{titlepage}
\begin{center}

\null\vspace{2cm}

{\sffamily \bfseries \huge Unsupervised neural and Bayesian models for zero-resource speech processing \par}\vfill

{\Large\itshape Herman Kamper \par}\vspace{1cm}\vfill

\includegraphics[width=35mm]{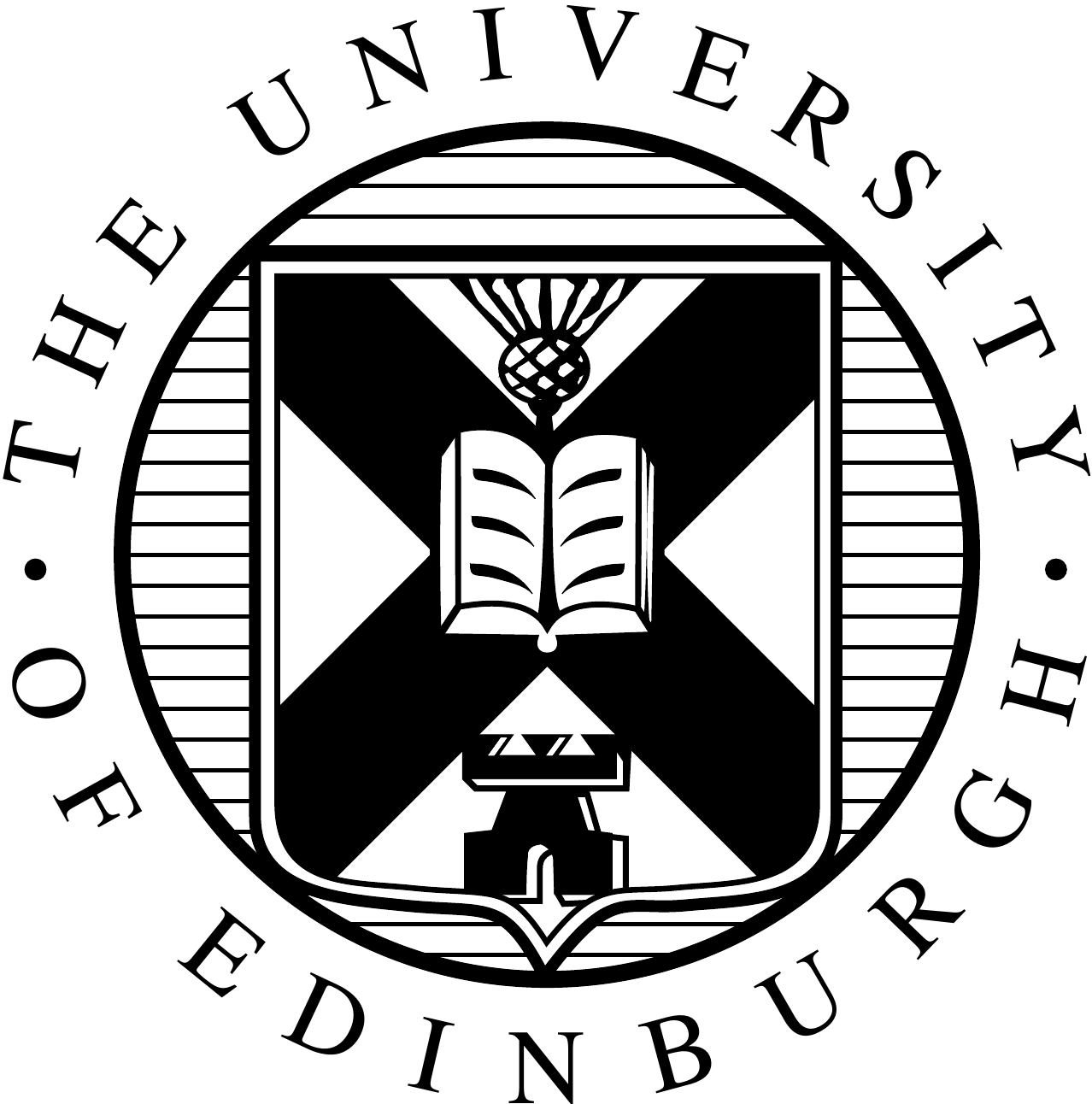}\vfill

{\large
\setstretch{1.35}
Doctor of Philosophy\\
Institute for Language, Cognition and Computation\\
School of Informatics\\
University of Edinburgh\\
2016\par
}
\end{center}
\end{titlepage}

\thispagestyle{empty}
\cleardoublepage
\thispagestyle{empty}

\vspace*{\fill}
\begin{center}
\setstretch{1.35}
\large
\textit{To Helena,}\\
\textit{I expect you to read the whole thing.}
\end{center}

\vspace{5cm}

\vspace*{\fill}
\cleardoublepage
\pagenumbering{roman}
\chapter*{Abstract}
\addcontentsline{toc}{chapter}{Abstract}
\makeatletter\@mkboth{}{Abstract}\makeatother

Zero-resource speech processing is a growing research area which aims to develop methods that can discover linguistic structure and representations directly from unlabelled speech audio.
Such unsupervised methods would allow speech technology to be developed in settings where transcriptions, pronunciation dictionaries, and text for language modelling are not available.
Similar methods are required for cognitive models of language acquisition in human infants, and for developing robotic applications that are able to automatically learn language in a novel linguistic environment.

There are two central problems in zero-resource speech processing: (i) finding frame-level feature representations which make it easier to discriminate between linguistic units (phones or words), and (ii) segmenting and clustering unlabelled speech into meaningful units.
The claim of this thesis is that both top-down modelling (using knowledge of higher-level units to to learn, discover and gain insight into their lower-level constituents) as well as bottom-up modelling (piecing together lower-level features to give rise to more complex higher-level structures) are advantageous in tackling these two problems.

The thesis is divided into three parts. The first part introduces a new autoencoder-like deep neural network for unsupervised frame-level representation learning.
This \textit{correspondence autoencoder} (cAE) uses 
weak top-down supervision from an unsupervised term discovery system that identifies noisy word-like terms in unlabelled speech data.
In an intrinsic evaluation of frame-level representations, the cAE outperforms several state-of-the-art bottom-up and top-down approaches, achieving a relative improvement of more than 60\% over the previous best system.
This shows that the cAE is particularly effective in using top-down knowledge of longer-spanning patterns in the data;
at the same time, we find that the cAE is only able to learn useful representations when it is initialized using bottom-up pretraining on a large set of unlabelled speech.

The second part of the thesis presents a novel unsupervised segmental Bayesian model that segments unlabelled speech data and clusters the segments into hypothesized word groupings.
The result is a complete unsupervised tokenization of the input
speech in terms of discovered word types---the system essentially performs unsupervised speech recognition.
In this approach, a potential word segment (of arbitrary length) is embedded in a fixed-dimensional vector space. The model, implemented as a Gibbs sampler, then builds a whole-word acoustic model in this embedding space while jointly performing segmentation.
We first evaluate the approach in a small-vocabulary multi-speaker connected digit recognition
task, where we report unsupervised word error rates (WER) by mapping the unsupervised decoded output to ground
truth transcriptions. 
The model achieves around 20\% WER, outperforming a previous HMM-based system by about 10\% absolute.
To achieve this performance, the acoustic word embedding function (which maps variable-duration segments to single vectors) is refined in a top-down manner by using terms discovered by the model in an outer loop of segmentation.



The third and final part of the study extends the small-vocabulary system in order to handle larger vocabularies in conversational speech data.
To our knowledge, this is the first full-coverage segmentation and clustering system that is applied to large-vocabulary multi-speaker data.
To improve efficiency, the system incorporates a bottom-up syllable boundary detection method to eliminate unlikely word boundaries. 
We compare the system on English and Xitsonga datasets to several state-of-the-art baselines. 
We show that by imposing a consistent top-down segmentation while also using bottom-up knowledge from detected syllable boundaries, both single-speaker and multi-speaker versions of our system outperform a purely bottom-up single-speaker
syllable-based approach. We also show that the discovered clusters can be made less speaker- and gender-specific by using features from the cAE (which incorporates both top-down and bottom-up learning).
The system's discovered clusters are still less pure than those of two multi-speaker unsupervised term discovery systems, but provide far greater~coverage.

In summary, the different models and systems presented in this thesis show that both top-down and bottom-up modelling can improve representation learning, segmentation and clustering of unlabelled speech data.

\selectlanguage{afrikaans}
\chapter*{Opsomming \\ {\Large (in Afrikaans)}}
\addcontentsline{toc}{chapter}{Opsomming}
\makeatletter\@mkboth{}{Opsomming}\makeatother

\textit{Nul-hulpbron-spraakverwerking} is 'n nuwe navorsingsarea wat poog om strukture en voorstellings van taal direk uit ongemerkte spraakdata te ontgin.
Sulke modelle van spraak wat sonder toesig afgerig kan word 
sal dit moontlik maak om spraaktegnologie te ontwikkel in omgewings waar transkripsies, uitspraakwoordeboeke en teks vir taalmodellering nie beskikbaar is nie.
Soortgelyke tegnieke is ook nodig in kognitiewe modelle wat die wyse naboots waarop baba's taal aanleer, asook in robotte wat vanself 'n nuwe taal kan aanleer in 'n onbekende taalomgewing.

Daar is twee groot probleme in nul-hulpbron-spraakverwerking: (i) kenmerkvektorvoorstellings 
moet gevind word wat dit makliker maak om tussen taaleenhede (fone of woorde) te onderskei, en (ii) rou, ongemerkte spraak moet in sinvolle eenhede gesegmenteer en gegroepeer 
word.
Hierdie proefskrif neem die standpunt in dat beide \textit{bo-na-onder modellering} 
(waar ho\"{e}rvlak-eenhede gebruik word om insig van laervlak-eenhede te verkry) asook \textit{onder-na-bo modellering} 
(waar laervlak-voorstellings saamgestik word sodat meer komplekse ho\"{e}rvlak-eenhede tevoorskyn kom)  nuttig is om hierdie twee probleme in nul-hulpbron-spraakverwerking aan te spreek.

Die proefskrif bestaan uit drie dele.
In die eerste deel stel ons 'n nuwe outo-enkoderende diep neurale netwerk 
bekend, wat gebruik kan word om raamvlak kenmerkvektorvoorstellings sonder toesig aan te leer.
Hierdie \textit{korrespondensie-outo-enkodeerder} (kOE) 
gebruik 'n swak vorm van bo-na-onder toesig wat deur 'n outomatiese termontdekkingstelsel verkry word (so 'n stelsel vind sonder toesig herhalende terme wat rofweg met woorde ooreenstem).
In 'n direkte evaluering van raamvlak voorstellings wys ons dat die kOE beter vaar as verskeie bo-na-onder en onder-na-bo tegnieke: die kOE verbeter die vorige beste resultaat met meer as 60\%.
Dit wys dat die kOE effektief gebruik maak van ho\"{e}rvlak-kennis van langer, herhalende patrone in die data; terselfdertyd vind ons dat dit slegs goeie voorstellings aanleer as dit op 'n onder-na-bo wyse ge\"{i}nitialiseer word.

Die tweede deel van die proefskrif beskryf 'n nuwe toesiglose Bayes\"{i}ese model wat ongemerkte spraakdata segmenteer en groepeer in eenhede wat (rofweg) met woorde ooreenstem.
Die uittree van die stelsel is 'n volledige dekodering van die spraak in terme van ontdekte woordeenhede---die stelsel voer dus 'n tipe spraakherkenning uit, sonder enige toesig.
In hierdie nuwe benadering word 'n potensi\"{e}le woordsegment (wat enige lengte kan wees) gekarteer na 'n vektorruimte met 'n vaste dimensionaliteit.
Ons model, wat as 'n Gibbs-monster-algoritme ge\"{i}mplementeer word, 
bou dan 'n akoestiese model oor volledige woorde op in hierdie vektorruimte terwyl dit terselfdertyd die data segmenteer.
Ons evalueer eerstens hierdie benadering deur dit toe te pas op 'n spraakdatabasis van gesproke syferreekse (so die data bevat slegs 'n klein woordeskat).
Deur die gedekodeerde spraak te vergelyk met korrekte transkripsies, kan ons 'n woordfouttempo (WFT) bereken.
Ons model behaal 'n WFT van 20\%, 'n absolute verbetering van rondom 10\% oor 'n vorige benadering wat verskuilde Markov-modelle gebruik.
Om hierdie resultaat te behaal moes die akoestiese karteringsfunksie (wat 'n segment van arbitr\^{e}re lengte projekteer na 'n enkele vektor) verfyn word deur gebruik te maak van terme wat ontdek is in 'n buitelus van bo-na-onder segmentasie.

In die derde en laaste deel van die proefskrif brei ons die kleinwoordeskatselsel uit sodat dit toegepas kan word op realistiese natuurlike spraak met 'n groter woordeskat.
Sover ons weet is dit die eerste keer dat 'n segmentasie-en-groeperingstelsel wat volle dekking bied op data met 'n realistiese woordeskat en veelvoudige sprekers toegepas word.
Om effektiwiteit te verbeter gebruik die stelsel 'n onder-na-bo tegniek wat outomaties grense tussen lettergrepe identifiseer en sodoende onwaarskynlike woordgrense elimineer.
Ons vergelyk die stelsel met verskeie vorige benaderings op data van twee tale: Engels en Xitsonga.
Deur 'n konsekwente volledige bo-na-onder segmentasie te kombineer met onder-na-bo identifikasie van lettergrepe, vaar beide die enkel- en multi-spreker weergawes van ons stelsel beter as 'n suiwer onder-na-bo lettergreep-gebaseerde tegniek.
Ons wys ook dat die ontdekte woordgroepe minder geslags- en spreker-spesifiek gemaak kan word deur kenmerkvektore van die kOE te gebruik (wat self 'n kombinasie van bo-na-onder en onder-na-bo modellering gebruik).
Ons volledige stesel se groeperings is nie so suiwer soos twee multi-spreker outomatiese termontdekkingstelsels nie, maar lewer veel beter dekking.

Om saam te vat: die verskillende modelle en stelsels wat in hierdie proefskrif beskryf word wys dat beide bo-na-onder en onder-na-bo modellering die kenmerkvektorvoorstellings, segmentasie en groepering van ongemerkte spraakdata verbeter.


\selectlanguage{english}

\chapter*{Lay summary}
\addcontentsline{toc}{chapter}{Lay summary}
\makeatletter\@mkboth{}{Lay summary}\makeatother

Automatic speech recognition is becoming part of our daily lives through applications like Google Now and Apple's Siri.
Ranging from assistive technologies for the disabled to automatic meeting transcription systems for the corporate workplace, future applications could improve the lives of many.
However, current methods require thousands of hours of \textit{transcribed} speech data for developing robust systems. This is why most commercial companies are focusing only on the first few hundred most common languages. However, there are about 7000 languages spoken in the world today.
If we only rely on current methods, speech technology will never be developed for many under-resourced languages.

The emerging area of \textit{zero-resource speech processing} seeks to address this problem.
Specifically, since it is often much easier to obtain speech recordings than transcriptions, zero-resource techniques aim to learn the structure of language directly from unlabelled raw speech audio.
This would allow speech technology to be developed in settings where it is impossible to get transcriptions.
As an example, such methods could be used by a linguist to analyze audio recordings of a previously undocumented language. 
The same methods could be used in a robot which is required to learn a new language directly from speech audio in an unknown environment.
Cognitive scientists have also long been interested in how infants acquire their native language using speech in their surroundings; since this problem is so similar to that of zero-resource speech processing, these methods could lead to new insights into language acquisition in humans.

This thesis makes contributions in both of the central problem-areas of zero-resource speech processing.
The first problem is how speech signals should be represented for an algorithm to make sense of it.
Raw speech is a complex signal, so the original waveform needs to be transformed into a \textit{representation} which makes it easier for a machine to process (in humans, our ears apply several such transforms before passing sound information on to the brain).
To address this problem, we propose a new deep neural network model which takes a small snippet of speech and transforms it so that it looks similar to another speech snippet containing the same speech sound.
The idea is that this network should capture the core information of the common sound that is being produced, while normalizing out aspects of the sounds which are not common (for example the two snippets could come from two different speakers).
Using this approach, we outperform the previous best representation learning method by more than 60\%.

The second problem in zero-resource speech processing is to solve the related tasks of \textit{segmentation} and \textit{clustering}.
Although it might not seem that way, there are no pauses between words in fluent speech: to see this, quickly say `stuffy nose' and then `stuff he knows', or compare `wreck a nice beach' to `recognize speech'. 
Segmentation is the task of predicting where words start and end in the stream of speech.
Even if we know where words start and end, we still don't know the types of words that are used in a language; clustering is the task of grouping together different segmented instances of the same word.
Clustering is difficult since the same word may sound very different when spoken by different people due to differences in pitch, timing, and accent: these cause little difficulty for humans but confuse computer algorithms.
Without transcriptions, the problems of segmentation and clustering are very hard to solve.

We propose a new algorithm which simultaneously segments and clusters raw speech into words.
We first apply our approach to a corpus of English digit sequences containing only a small number of unique words. An  algorithm can be evaluated by determining how accurately it discovers the true words in the data.
Compared to a previous study, our approach reduce errors by more than a third, giving an 80\% accuracy on this small-vocabulary task.
We then extend the approach in order to handle larger vocabularies in conversational speech data in two languages: English and Xitsonga (an under-resourced Bantu language spoken in southern Africa).
This is the first time that a zero-resource system such as ours is applied to large-vocabulary data from multiple speakers.
Previous work focused on the easier task of processing speech from individual speakers; we show that our approach outperforms earlier work even though it simultaneously processes data from all the speakers.
When we incorporate representations from the deep neural network developed in the first part of the thesis, our approach deals with multiple speakers even better.

This thesis looks at speech recognition from a very different perspective compared to the traditional view.
This led to new ways to combine different sources of knowledge in raw speech data.
We hope that the principles developed in this work would be applied in future zero-resource speech processing systems, and ultimately lead to speech technology in under-resourced settings where users can benefit from it directly.

\chapter*{Acknowledgements}
\makeatletter\@mkboth{}{Acknowledgements}\makeatother

I had an absolutely awesome team of supervisors.
To Sharon Goldwater, I am especially grateful for her thoughtful guidance and mentorship, for steering me in the right direction while giving me freedom to explore, and for teaching me the importance of identifying and asking the key (and sometimes awkward) questions.
Aren Jansen's passion was a constant source of encouragement leading to an appreciation for the interesting problems in our field---it is amazing how someone I've never met (in person) could have such an impact. 
To Simon King I am grateful for helping me settle into CSTR-life at the start and for making sure I was always able to attend conferences. 

\edit{I would like to thank my thesis examiners Steve Renals and Jim Glass for making my defence an enjoyable experience.}
Many thanks also to the Commonwealth Scholarship Commission for funding my studies.


It was a privilege to work with Micha Elsner, Daniel Renshaw, Adam Lopez and Sameer Bansal in some great partnerships.
I want to thank Karen Livescu from TTI in Chicago for all her support; I learnt a lot from her, Weiran Wang and Hao Tang during my summer visit to this amazing institution.
Colleagues in ILCC and CSTR provided many stimulating (and often highly entertaining) discussions.
Several researchers within the zero-resource community assisted with questions and requests: Okko R{\"a}s{\"a}nen, Shreyas Seshadri, Roland Thiolli{\`e}re and Maarten Versteegh all gave help and input for the experiments in Chapter~\ref{chap:bucktsong}.
The familiar face of Oliver Walter was a constant source of encouragement at conferences.

Without those that helped make Edinburgh my home, I would have long ago given up this battle (mainly against the Scottish weather).
My Edinburgh family is large, but I need to thank by name Ben, Carolyn, Michael, Carina, Andy and Moira; the many mosque meals, squash games, board games, coffees and kitchen-chats were a blessing.
The gospel family at Chalmers Church, and especially the guys at Cord, helped me grow and kept me on track with their many prayers.



Although I was in Edinburgh, South Africa went to great lengths to make sure I was okay. \selectlanguage{afrikaans}Wikus, sonder ons Skype-sessies sou ek baie vroeg gedood het.
Freek, Barend, Est{\'e}, Simon, Carla: julle het baie baie verlange veroorsaak.
Baie dankie vir al die lekker Oxford- en Engeland-kuiers, Jan! \selectlanguage{english}I cannot give enough thanks to my parents, brother and sisters. \selectlanguage{afrikaans}Paps en Mams, die Sondag Skype-sessies het my deur baie weke gedra; dankie vir julle baie raad en gebede.
Femke, jou WhatsApp boodskappe was baie keer die ligpunt van my dag.
Miens, jou kaaskoek is die grootste rede vir my om vakansies huis toe te kom.
Franna, dankie dat jy altyd daar was, selfs al was dinge rof vir jou.
Dr{\'e}, ek is s{\'o} bly ek het 'n sussie bygekry wat so mooi na my boetie kyk, hey.
Aan Pa Klaas en Ma Coramine, dankie vir die baie re{\"e}lings en opofferings, en dankie aan Oom TK dat hy die Kaap-seun aanvaar het. Helena, jy is 'n wonder.

\selectlanguage{english}``So the \textsc{Lord} God formed from the ground all the wild animals and all the birds of the sky. He brought them to the man to see what he would call them, and the man chose a name for each one.''  Thank You for giving us this to explore, and for language.











\chapter*{Declaration}

I declare that this thesis was composed by myself, that the work contained herein is
my own except where explicitly stated otherwise in the text, and that this work has not
been submitted for any other degree or professional qualification except as specified.\par
\vspace{1in}
\begin{flushright}
    Herman Kamper
\end{flushright}

\tableofcontents
\cleardoublepage
\pagenumbering{arabic}

\chapter{Introduction}
\label{chap:introduction}

The last few years have seen great advances in speech recognition. Much of this progress is due to the resurgence of neural networks; most speech systems now rely on deep neural networks (DNNs) with millions of parameters~\citep{dahl+etal_taslp12,hinton+etal_spm2012}.
However, as the complexity of these models has grown, so has their reliance on labelled training data. Currently, system development requires large corpora of transcribed speech audio data, texts for language modelling, and pronunciation dictionaries.
Despite speech applications becoming available in more languages, it is hard to imagine that resource collection at the required scale would be possible for all 7000 languages spoken in the world today.

A major stumbling block for system development is the transcription of speech audio, which (compared to audio data collection itself) is extremely expensive and time consuming.
To address this problem, we need unsupervised methods that are able to discover the latent structure of language directly from speech audio.
Human infants excel at this task during early language learning: using speech from their surroundings, infants learn the phonetic contrasts, lexicon and grammar of their native language with minimal guidance~\citep{kuhl_nature04}.
Unfortunately, although cognitive scientists have long been interested in modelling this process, most cognitive models of language acquisition use transcribed symbolic sequences as input rather than continuous speech~\citep{goldwater+etal_cognition09}.

The emerging area of \textit{zero-resource speech processing} seek to address the problems 
of learning meaningful representations and linguistic structures directly from unlabelled speech audio~\citep{jansen+etal_icassp13,versteegh+etal_interspeech15}.
Successful zero-resource methods would make it possible to develop speech applications in severely under-resourced settings where transcribed speech data is simply not available.
As a practical example, such methods could allow linguists to investigate and analyze speech audio where it is difficult or impossible to get transcribed speech, for instance when dealing with languages without a written form~\citep{besacier+etal_speechcom14}.
Zero-resource systems could also be used as a new way to model infant language acquisition from naturalistic speech input~\citep{rasanen_speechcom12}.
A related problem occurs in robotics, where a robot is required to learn language in an unfamiliar linguistic environment~\citep{sun+vanhamme_csl13,taniguchi+etal_arxiv15}.
Seeing zero-resource speech processing as a (very difficult) machine learning problem, work towards this goal could lead to new insights and modelling approaches for speech processing in general---a field where models often become standardized (e.g.\ HMMs and DNNs have dominated the last two decades).
It has also already been shown that purely unsupervised methods can be used to improve the performance of supervised~systems~\citep{jansen+etal_icassp13}.

\section{Goals and methodology}

Interest in zero-resource speech processing has grown considerably in the last few years, with two central problem areas emerging: (i)~finding speech representations (often at the frame level) that make it easier to discriminate between meaningful linguistic units and (ii)~segmenting and clustering unlabelled speech audio into meaningful units. 
This thesis makes contributions in both these areas.
Before describing these problems and our proposed solutions in more detail, we outline the main claim of the thesis.

\subsection{Top-down and bottom-up modelling}
\label{sec:intro_topdown_bottomup}

The overarching claim of this thesis is that both top-down and bottom-up modelling are beneficial for zero-resource speech processing.
We use \textit{top-down modelling} to refer to a process where knowledge of higher-level units (typically the units of ultimate interest) are used to learn, discover and gain insight into their lower-level constituents.
Conversely, \textit{bottom-up modelling} uses 
knowledge obtained directly from the lower-level features to guide learning and discovery of more complex higher-level structures.
Although it is sometimes difficult to strictly place a particular method into one of these categories, 
thinking in these terms often illuminates how the method is approaching its task.
We claim that a combination of both methodologies are beneficial in order to learn better representations and linguistic structures directly from raw speech.

This claim is analogous to observations made in studies of human language acquisition:
\cite{feldman+etal_ccss09} note that infants are still learning about phonetic contrasts in their native language even after starting to segment words from continuous speech.
This suggests that infants could use top-down knowledge of discovered words to assist phonetic category acquisition. 
Conversely, successful bottom-up distinctions between phonetic categories could assist in disambiguating different word types.
In analogy, the computational models and systems presented in this thesis benefit from both methodologies.
As an example, we show in Chapter~\ref{chap:cae} that top-down knowledge from discovered word segments can be used to find bottom-level speech features that are more discriminative.
Conversely, in Chapter~\ref{chap:bucktsong} we show that by using these improved bottom-level features within a top-down segmentation and clustering model, the model discovers clusters that are purer and more speaker- and gender-independent.
Below we expand on how our proposed methods incorporate both top-down and bottom-up modelling in addressing the two main problems in zero-resource speech processing.




\subsection{Unsupervised representation learning}


The first problem is to use unlabelled data to learn speech representations or features that make subsequent unsupervised discovery tasks easier.
Stated more formally, the task of \textit{unsupervised representation learning} involves finding
speech features (often at the frame level) that make it easier to discriminate between meaningful linguistic units (normally phones or words) while being robust to irrelevant information (such as speaker and gender)~\citep{versteegh+etal_interspeech15}.
The task has been described as `phonetic discovery', `unsupervised acoustic modelling' and `unsupervised subword modelling', depending on the type of feature representations that are produced.

Many early studies used {bottom-up} methods, where representations are learnt directly from the low-level acoustic features.
Approaches include unsupervised Gaussian mixture models (GMMs)~\citep{zhang+glass_icassp10,chen+etal_interspeech15} and 
bottom-up trained unsupervised hidden Markov models (HMMs)~\citep{varadarajan+etal_acl08,lee+glass_acl12,siu+etal_csl14}.
In~\citep{jansen+church_interspeech11} and~\citep{jansen+etal_icassp13b}, the authors found that these purely bottom-up approaches could be significantly improved by using {top-down} word-level constraints.
They proposed that a first-pass discovery system could be used to automatically find reoccurring word- or phrase-like patterns in the data; these longer-spanning patterns could then be used as weak top-down supervision for subsequent frame-level representation learning.
In an intrinsic evaluation, features from top-down constrained HMMs~\citep{jansen+church_interspeech11} and GMMs~\citep{jansen+etal_icassp13b} were shown to significantly outperform their purely bottom-up counterparts.
These studies are described in more detail in Section~\ref{sec:background_phondisc}.

The recent success of DNNs in supervised speech recognition has naturally prompted subsequent studies in the zero-resource area~\citep{zhang+etal_icassp12,zeiler+etal_icassp13,badino+etal_icassp14}.
In the supervised case, DNNs 
implicitly perform representation learning (to great effect): lower layers can be interpreted as a deep feature extractor which is learnt jointly with a supervised classifier~\citep{yu+etal_iclr13}. 
In the unsupervised case, however, we do not have access to the phone class targets required for fine-tuning standard feedforward DNNs. 
Other types of neural networks like restricted Boltzmann machines~\citep{zhang+etal_icassp12} or autoencoders~\citep{zeiler+etal_icassp13} can be trained on unlabelled speech data without explicit supervision, using data likelihood or reconstruction error to define a loss function. 
However, these are purely bottom-up models, operating directly on the acoustics without regard to longer-spanning patterns in the data (as in the early unsupervised GMM and HMM approaches described above).

In this thesis, our proposal for unsupervised frame-level representation learning 
is to incorporate top-down knowledge of longer-spanning patterns, as was done for HMMs and GMMs in~\citep{jansen+church_interspeech11,jansen+etal_icassp13b}, but to do so within the neural network regime.
Concretely, we propose the \textit{correspondence autoencoder} (cAE), an autoencoder-like unsupervised DNN 
which uses aligned feature frames from top-down discovered words as input-output pairs.
Apart from using this top-down signal, the cAE is initialized using bottom-up pretraining on a large corpus of unlabelled speech.
In an intrinsic evaluation, we show that the cAE effectively combines these top-down and bottom-up signals in order to achieve major improvements over previous bottom-up and top-down methods.
In the last part of the thesis we also evaluate features from the cAE within a segmentation and clustering system (described next), and show that these features also extrinsically improve cluster purity and speaker- and gender-independence in this downstream task.

%
%
%

%

\subsection{Unsupervised segmentation and clustering of speech}
\label{sec:intro_unsup_seg_clust}

The second area of zero-resource research deals with unsupervised segmentation and clustering.
Here the aim is to find the boundaries of meaningful linguistic units within the stream of unlabelled speech, and to cluster these units into groups of the same (unknown) type.
Typically the units of interest are longer-spanning word- or phrase-like patterns.
One version of this task involves segmenting and clustering isolated repeated word-like patterns that are spread out over the speech data~\citep{park+glass_taslp08,jansen+vandurme_asru11}.
We refer to this task as \textit{unsupervised term discovery} (UTD); it is also referred to as `lexical discovery' or `spoken term discovery' in the literature.
It is such a system that is used to obtain the word pairs used for the cAE described above.
The isolated discovered segments in UTD are spread out sparsely over the data, leaving much of the input speech as background.

In contrast, \textit{full-coverage speech segmentation and clustering} aims to completely segment speech into a sequence of word-like units, proposing word boundaries and cluster assignments for the entire input~\citep{sun+vanhamme_csl13,chung+etal_icassp13,walter+etal_asru13,lee+etal_tacl15,rasanen+etal_interspeech15}.
Such systems essentially perform a type of unsupervised speech recognition.
This has several benefits over sparse discovery (as in UTD).
As a practical example, a linguist analyzing unlabelled speech data might be interested in a particular portion of speech not covered by a sparse discovery method.
More generally, since successful full-coverage segmentation would provide a complete tokenization of its input (as traditional speech recognition systems do), it would allow downstream applications to be developed in a manner similar to when supervised systems are available. This includes tasks like query-by-example search (finding a spoken query in a speech collection) and speech indexing (grouping together related utterances in a corpus).
From a cognitive modelling perspective, humans also perform full-coverage segmentation. Most existing models of infant language acquisition (taking symbolic input) therefore perform full-coverage segmentation~\citep{goldwater+etal_cognition09}, and the same would be useful in a cognitive model of language acquisition from continuous speech input.

In this thesis, our second aim is to develop such an unsupervised full-coverage segmentation and clustering system.
A few recent studies~\citep{sun+vanhamme_csl13,chung+etal_icassp13,walter+etal_asru13,lee+etal_tacl15,rasanen+etal_interspeech15}, summarized in detail in Section~\ref{sec:background_fullcoverage}, share this goal.
Almost all of these follow an approach of bottom-up phone-like subword discovery with subsequent or joint word discovery, working directly on the frame-wise acoustic speech features.
We instead propose a very different approach in which whole words are modelled directly using a segmental Bayesian model.
Concretely, a fixed-dimensional representation of whole segments is used: any potential word segment consisting of an arbitrary number of speech frames is mapped to a single fixed-length vector, its \textit{acoustic word embedding}.
Ideally, segments of the same type should be mapped to similar areas in the embedding space.
Using this representation, the model jointly segments unlabelled speech data into word-like segments and then clusters these segments using a whole-word acoustic model.
The result is a complete unsupervised tokenization of the input
speech in terms of discovered clusters, each cluster representing a discovered word type.\footnote{`Word type' refers to distinct words, such as the entries in a lexicon. `Word token' refers to different realizations of a particular word.}
Because the model has no subword level of representation and models whole segments directly, we refer to the model as \textit{segmental}~\citep{zweig+nguyen_interspeech10}.

We first evaluate this model on a small-vocabulary unsupervised speech recognition task using a multi-speaker corpus of English digit sequences.
Compared to a more traditional subword-based HMM approach (representative of many other full-coverage methods), our approach achieves about 10\% absolute better unsupervised word error rate (WER), calculated by mapping the
unsupervised decoded output to ground truth transcriptions.
Subsequently, we evaluate our approach on large-vocabulary multi-speaker data from two languages: English and Xitsonga.
To our knowledge, this is the first time that a full-coverage method is evaluated on large-vocabulary data from multiple speakers.
Although the model imposes a consistent top-down segmentation and clustering of entire utterances, it is flexible in that bottom-up constraints can be easily incorporated into the segmentation algorithm; 
in the large-vocabulary system, a bottom-up syllable boundary detection method is used to eliminate unlikely word boundaries.
We show that the combination of top-down segmentation with bottom-level syllable-based constraints results in consistent improvements over a purely bottom-up single-speaker syllable-based approach.
Further improvements are achieved by using features from the cAE as input (incorporating both top-down and bottom-up learning to obtain better frame-level representations) instead of traditional acoustic features.

\section{Contributions}
\label{sec:intro_contributations}

Using a combination of top-down and bottom-up modelling, this thesis makes contributions in both unsupervised representation learning, and in segmenting and clustering unlabelled speech.
We highlight the following main contributions:


\begin{itemize}
    \item The cAE is the first neural network model to incorporate top-down constraints from a term discovery system for unsupervised frame-level representation learning.
    Furthermore, earlier HMM- and GMM-based approaches~\citep{jansen+church_interspeech11,jansen+etal_icassp13b} all used ground truth words from forced alignments to simulate UTD, while we use word pairs from a real UTD system, making the overall approach truly unsupervised.
    Since first publication of the cAE~\citep{kamper+etal_icassp15}, it has been applied and extended by other researchers, both at the University of Edinburgh~\citep{renshaw+etal_interspeech15,renshaw_masters16} and elsewhere~\citep{yuan+etal_interspeech16}.
    \item We propose the first whole-word segmental model for unsupervised full-coverage speech segmentation and clustering.
    In contrast to other studies, the model does not perform any explicit subword modelling, but is still flexible enough to handle bottom-up constraints in a principled manner.
    We do not argue that direct whole-word modelling is necessarily superior (although there are several merits as outlined in Section~\ref{sec:tidigits_related_studies_full_coverage}). Rather, we see our approach as a new contribution to zero-resource speech processing, and show that whole-word modelling (specifically using acoustic word embeddings) is an attractive and sensible research direction.
    \item To our knowledge, we present the first zero-resource full-coverage system that is evaluated on large-vocabulary multi-speaker data. Previous systems have either focused on identifying isolated terms~\citep{park+glass_taslp08,jansen+vandurme_asru11,lyzinski+etal_interspeech15}, were speaker-dependent~\citep{lee+etal_tacl15,rasanen+etal_interspeech15}, or used only a small vocabulary~\citep{walter+etal_asru13}.
    We perform evaluation on both English and Xitsonga data, showing that the approach generalizes across languages.
    \item To our knowledge, this large-vocabulary system is also the first full-coverage segmentation and clustering system to incorporate unsupervised representation learning (using the cAE). We show that this unsupervised representation learning method improves cluster purity as well as speaker- and gender-independence.
\end{itemize}

\section{Thesis outline}
\label{sec:thesis_outline}


\noindent \myemph{Chapter~\ref{chap:background}:\ \nameref{chap:background}.} 
Two research communities in particular share an interest in zero-resource speech processing: we review studies from both the speech engineering and the scientific cognitive modelling communities. 
From this review, 
we conclude that segmental modelling of whole word-like units is an attractive approach for full-coverage segmentation and clustering of raw speech.
Furthermore, such a segmentation system should ideally incorporate unsupervised representation learning, specifically using top-down constraints to guide representation learning. 
Finally, higher-level context (language modelling) could prove useful, but previous studies have found this challenging.\vspace{1em}

\noindent \myemph{Chapter~\ref{chap:cae}:\ \nameref{chap:cae}.}
This chapter introduces the {correspondence autoencoder} (cAE). 
In a word discrimination task which intrinsically evaluates the quality of frame-level features, we compare the cAE to state-of-the-art bottom-up and top-down GMM-based methods, and to a purely bottom-up stacked autoencoder.
We show that the cAE achieves a relative improvement of more than 60\% over the previous best system. 
This shows that the cAE makes effective use of the weak top-down supervision from a first-pass UTD system, while using bottom-up pretraining on a large corpus of unlabelled speech for initialization.\vspace{1em}

\noindent \myemph{Chapter~\ref{chap:tidigits}:\ \nameref{chap:tidigits}.}
This chapter introduces the novel segmental Bayesian model for full-coverage segmentation and clustering of unlabelled speech.
We first give an intuitive overview of the model, and then give complete mathematical and algorithmic details.
We evaluate the model on a multi-speaker small-vocabulary connected digit recognition task. 
The model achieves around 20\% unsupervised WER, outperforming an HMM-based approach by about 10\% absolute.
To achieve this performance, the acoustic word embedding approach is refined using top-down discovered terms obtained by running our system in an outer loop of segmentation.
On this small-vocabulary task, the model does not require a pre-specified vocabulary size, in contrast to the HMM baseline.\vspace{1em}

\noindent \myemph{Chapter~\ref{chap:bucktsong}:\ \nameref{chap:bucktsong}.}
This chapter presents our large-vocabulary segmental Bayesian model. 
To improve efficiency, the model incorporates a bottom-up syllable boundary detection method to eliminate unlikely word boundaries.
The embedding method used to map variable-duration word segments to fixed-length vectors is also simplified.
After describing these changes, we also give details for including a bigram model of word predictability (up to this point a unigram model is used). 
Both speaker-dependent and speaker-independent evaluations are performed on data from two languages: English and Xitsonga.
From a comparison with other state-of-the-art systems, we conclude that the improvements of our system on several metrics are due to the consistent top-down segmentation that it imposes over entire utterances while simultaneously adhering to bottom-level constraints.
Another finding is that cAE features (Chapter~\ref{chap:cae}) result in clusters that are purer and less speaker- and gender-specific than when using traditional features.
Because of the peaked nature of the acoustic model component of the model, the bigram extension is not able to take advantage of word-word dependencies in the data.\vspace{1em}

\noindent \myemph{Chapter~\ref{chap:conclusion}:\ \nameref{chap:conclusion}.}
In the conclusion, we return to the goals and overarching claim of the thesis.
We summarize and highlight the main findings from the previous chapters, and explain how both bottom-up modelling (e.g.\ in the segmental representations, syllable boundary detection) and top-down modelling (e.g.\ in feature learning, segmentation) are shown to be benificial for zero-resource speech processing.
Finally, recommendations for future work are discussed. 

\section{Published work}

Chapter~\ref{chap:cae} is largely based on the paper presented at \textit{ICASSP 2015}~\citep{kamper+etal_icassp15}.
Preliminary work for Chapter~\ref{chap:tidigits} was presented at \textit{SLT 2014}~\citep{kamper+etal_slt14} and at \textit{Interspeech 2015}~\citep{kamper+etal_interspeech15}, with the final model and evaluation published in the \textit{IEEE/ACM Transactions on Audio, Speech and Language Processing}~\citep{kamper+etal_taslp16}.
Chapter~\ref{chap:bucktsong} is based on the \textit{arXiv} journal publication~\citep{kamper+etal_arxiv16}.

\graphicspath{{background/fig/}}

\chapter{Background}
\label{chap:background}

The problem of discovering linguistic knowledge directly from raw speech audio has sparked recent interest in two communities, which is summarized in this chapter.
In the speech engineering community, successful unsupervised modelling techniques would allow rapid development of zero-resource speech technology for severely under-resourced languages.
Studies in this area have considered discovery of reoccurring word- or phrase-like patterns in speech data, as well as unsupervised representation learning for obtaining better speech features at the phone or frame level.
In the scientific cognitive modelling community, unsupervised speech processing is very relevant since it is similar to the problem faced by infants during early language learning. 
Here, previous studies have considered full-coverage word segmentation and lexicon discovery, but have done so using transcribed symbolic input. 
Recent studies, spanning both communities, have attempted full-coverage segmentation and clustering of raw speech, but only on small-vocabulary data or data from a single speaker.
The chapter is concluded with a discussion on useful and essential aspects of a successful zero-resource segmentation and clustering system.




\section{Unsupervised discovery of words in speech}
\label{sec:background_utd}

Unsupervised term discovery (UTD), sometimes referred to as `lexical discovery' or `spoken term discovery', is the speech processing task of finding meaningful repeated word- or phrase-like patterns in raw speech audio.
Typically, UTD systems aim to find and cluster repeated isolated acoustic segments within utterances, and the rest of the data is treated as background~\citep{park+glass_taslp08}.
A task closely related to UTD is unsupervised query-by-example search, where a spoken query is given as input and an unsupervised system needs to return all the utterances in a corpus containing that query~\citep{metze+etal_icassp13}.

\subsection{Segmental dynamic time warping}
\label{sec:background_sdtw}

Dynamic time warping (DTW) is a dynamic programming method for finding the optimal alignment between two time series.
For speech it can be used to obtain an overall measure of similarity between two vectorized utterances.
However, since DTW aligns entire sequences, similar segments within two utterances are not identified.
In order to perform UTD, \cite{park+glass_taslp08} therefore developed a variant of DTW, called segmental DTW, which allows pairs of similar audio segments within utterances to be discovered and then clustered into hypothesized word types.
Their algorithm could find most of the frequent words in the MIT lecture corpus.

Segmental DTW now forms part of most state-of-the-art UTD and unsupervised query-by-example search systems.
Follow-up work has built on Park and Glass' original method in various ways, for example through improved
feature representations~\citep{zhang+glass_asru09,zhang+etal_icassp12}, by greatly improving its efficiency by using randomized hashing algorithms~\citep{jansen+vandurme_asru11}, and by investigating the cognitive plausibility of the algorithm~\citep{mcinnes+goldwater_ccss11}.

As is done in the models of Chapters~\ref{chap:tidigits} and~\ref{chap:bucktsong} in this thesis, most UTD systems operate
on whole-word representations, with no subword level of representation.
However, each word is represented as a vector time series with variable lengths (number of frames), which requires DTW for comparisons.  Despite the elegant solution provided by segmental DTW for finding similar sub-sequences, each DTW comparison has quadratic time complexity in the duration of the segments being compared. Each utterance in the corpus also needs to be compared to each other utterance, which itself has quadratic complexity in the number of utterances.
This means that DTW-based approaches are not scalable for many applications and constraints are often used.
For example, in the UTD system of \cite{jansen+vandurme_asru11}, 
a coarse hashing technique
is first used to limit the search space for the subsequent segmental DTW.

\subsection{Embedding speech segments in a fixed-dimensional space}
\label{sec:background_segmental}

Because of the time complexity of segmental DTW (which is expensive even when using some approximate pre-processing technique), \cite{levin+etal_asru13} proposed an alternative approach, in which an arbitrary-length speech segment is embedded in a fixed-dimensional space such that segments of the same word type have similar embeddings.
Segments can then be compared by simply calculating a distance in the embedding space, a linear time operation in the embedding dimensionality.
Standard clustering approaches can also be applied directly in this space.

Several embedding approaches were proposed and compared in~\citep{levin+etal_asru13}, based on 
the idea of using a {\it reference vector} to construct the mapping from variable-length vector time series to \edit{a} fixed-dimensional vector.
For a target speech segment, a reference vector 
consists of the DTW alignment cost to every exemplar in a reference set.
Applying dimensionality reduction to the reference vector yields the desired embedding.
In this thesis, we will refer to such embedding vectors as \textit{acoustic word embeddings}, or simply \textit{embeddings}.
The intuition of the reference vector approach is that the content of a speech segment should be characterized well through its similarity to other segments.
Although this approach still requires the calculation of several DTW alignment costs, the number of calculations is linear in the number of segments to embed if the reference set size is fixed.
In the most relevant setup for us, Levin et al.\ assumed that a set of pre-segmented word exemplars is available, but that their identities are unknown.
Several dimensionality reduction approaches were evaluated; 
it was found that Laplacian eigenmaps with a kernel-based out-of-sample extension (a non-linear graph embedding technique) performed similarly to DTW for capturing word similarity.
We use this approach as embedding function for the segmental model presented in Chapter~\ref{chap:tidigits}, with complete details given in Section~\ref{sec:tidigits_embeddings}.

Several studies have since used acoustic word embeddings.
In their own follow-up work, \cite{levin+etal_icassp15} developed a complete embedding-based query-by-example search system.
\cite{chung+etal_arxiv16} employed the same framework as in~\citep{levin+etal_icassp15}, but used an autoencoding encoder-decoder neural network as embedding function, and achieved improvements over DTW.
The encoder-decoder neural network encodes a variable-length sequence into a single acoustic word embedding vector, and is trained to reconstruct its variable-length input given the embedding vector.


As in segmental DTW, these acoustic word embedding approaches are attractive since they do not require explicit subword modelling. But they are (typically) more efficient than DTW.
They also allow segments to be compared directly in a fixed-dimensional space, meaning that word discovery can be performed using standard clustering methods.
Furthermore, segmental embedding approaches do not make the frame-level independence assumptions of many speech processing systems, which have long been argued against~\citep{zweig+nguyen_interspeech10,gillick+etal_asru11}.

\section{Unsupervised phonetic discovery and representation learning}
\label{sec:background_phondisc}

In speech processing, {phonetic discovery} is the task of discovering the categorical set of subword units that make up a language and relating these to the underlying acoustic features (so it is sometimes called `unsupervised acoustic modelling').
{Unsupervised representation learning}, in this context, is the task of learning a frame-level mapping from the original features to a new representation that make it easier to discriminate between different linguistic units (normally subwords or words).
It is sometimes difficult to make a precise distinction between these two tasks~\citep{versteegh+etal_interspeech15}, and so these are discussed together here.
Below we describe approaches that learn purely bottom-up (directly from the acoustics), and then those that use top-down knowledge to guide discovery.
Before that, however, we remark on how these systems are typically evaluated.

\subsection{Evaluation of frame-level speech representations}
\label{sec:background_phondisc_eval}

The evaluation of zero-resource speech processing methods is a research problem in itself~\citep{ludusan+etal_lrec14}.
Early studies on phonetic discovery and unsupervised representation learning used extrinsic evaluations in downstream tasks such as query-by-example search~\citep{zhang+glass_asru09,zhang+glass_icassp10} and topic classification~\citep{gish+etal_interspeech09,siu+etal_interspeech10}. Systems that perform explicit phonetic discovery and segmentation can be evaluated intrinsically in terms of their accuracy in detecting phone boundaries~\citep{lee+glass_acl12}. This is not possible, however, for continuous vector representations such as those from some unsupervised representation learning methods.

In order to compare different feature representations without the need to build a full search or recognition system, \cite{carlin+etal_icassp11} developed the \textit{same-different task}.
This task is general in that it allows both vector time series representations (such as those from continuous representation learning methods) and tokenized representations (such as those from phonetic discovery) to be compared.
For every pair of word tokens in a test
set of pre-segmented words, the DTW distance (for vector time series representations) or the
edit distance (for tokenized representations) is calculated using the representation under evaluation.
Two words can then be classified as belonging to the same or different type based on whether the distance is below some threshold,
and a precision-recall curve is obtained by varying the threshold.
To evaluate representations across different operating points, the area under the precision-recall curve is calculated
to yield the final evaluation metric, referred to as the average precision (AP). \cite{carlin+etal_icassp11} found perfect correlation between AP and phone error rate in a supervised setting, justifying it as an effective way to evaluate different representations of speech in unsupervised settings.
Note that, apart from using this task to evaluate phonetic discovery or unsupervised representation learning at the frame-level, this task can also be used to directly evaluate whole-word fixed-dimensional vector representations, such as those described in Section~\ref{sec:background_segmental}; in this case, instead of calculating the DTW distance over a vector time series, the cosine or Euclidean distance between single acoustic word embedding vectors would be used.\footnote{We use this approach to evaluate different acoustic word embedding approaches in Section~\ref{sec:downsampling}.}

Another recent evaluation method for phonetic discovery and frame-level representation learning is the \textit{ABX task}~\citep{schatz+etal_interspeech13}. This task measures the discriminability of representations by asking whether a speech segment $X$ is more similar to segments $A$ or $B$, where the segments $A$ and $X$ are different realizations of the same type, while $B$ is different. The task is typically performed on minimal \edit{phone trigram} pairs, so $A$ and $X$ would be realizations of the same \edit{phone trigram} sequence (e.g.\ `bag'), while $B$ is different from $A$ and $X$ in its middle phone (e.g.\ `bug').
The final metric is an error rate over all $(A, B, X)$ triplets in a test set.
Again, as in the same-different task, both vectorized and tokenized representations can be evaluated using DTW or edit distance for segment comparison.
Both the same-different and ABX tasks perform an intrinsic evaluation of the discriminability of a particular feature representation: same-different does so at the word level, while ABX does so (typically) at the \edit{phone trigram} level using minimal pairs.

\subsection{Bottom-up approaches}
\label{sec:background_bottomup}

Since 2008, several researchers have worked on training unsupervised hidden Markov models (HMMs) directly from unlabelled audio.
Approaches include the successive state-splitting algorithm of \cite{varadarajan+etal_acl08}, and the more traditional approach of Gish, Siu and colleagues~\citep{gish+etal_interspeech09,siu+etal_interspeech10,siu+etal_interspeech11,siu+etal_csl14} in which HMMs are refined through an iterative re-estimation and unsupervised decoding procedure.
Since most of these systems were either developed on very small corpora~\citep{varadarajan+etal_acl08} or evaluated in downstream tasks such as topic classification~\citep{gish+etal_interspeech09,siu+etal_interspeech10}, it is unclear whether the discovered units truly correspond to phone-like units.
\edit{Nevertheless, this early work clearly showed the applicability and promise of unsupervised subword modelling in a range of speech processing tasks.}
As a precursor to their full word segmentation system (complete details in Section~\ref{sec:background_fullcoverage}), \cite{lee+glass_acl12} developed a non-parametric Bayesian HMM which automatically infers the number of subword HMMs.
Their system achieved a 76.3\% phone boundary detection $F$-score on TIMIT, and they showed qualitatively that the discovered units mapped well to ground truth phones.
As was the case for the work by Gish et al., their system relied on a presegmentation method to eliminate unlikely phone boundaries (based on spectral change) in order to reduce computational~load.

A simpler approach than using HMMs is to train a large universal background model (UBM) on unlabelled speech data.
A UBM is typically a Gaussian mixture model (GMM) trained directly on the acoustic features; although it ignores ordering information (in contrast to HMMs), it requires fewer heuristics than many of the above approaches.
The idea is that, given a sufficiently large GMM, every component would correspond to some 
subword unit.
Instead of using standard acoustic features, \cite{zhang+glass_asru09,zhang+glass_icassp10} performed segmental DTW on posteriorgrams from a GMM-UBM and obtained significant improvements in query-by-example search and term discovery (extrinsic evaluations) compared to using traditional acoustic features directly.
In follow-up work, \cite{anguera_icassp12} incorporated a discriminative clustering objective, while \cite{chen+etal_interspeech15} obtained improvements in an ABX evaluation by using an infinite GMM, a non-parametric extension of the GMM which infers its number of components automatically.
\cite{jansen+etal_icassp13b} also found that GMM-UBM posteriorgrams improved performance over standard acoustic features in the intrinsic same-different word discrimination task; however, when introducing additional top-down word-level information, much larger gains were achieved (details in the next section).\footnote{Although GMMs are categorical, the soft GMM posteriorgrams can be seen as a new distributed feature representation of the acoustic input. This is one reason why some authors~\citep{versteegh+etal_interspeech15} don't make a strict distinction between `phonetic discovery' and `unsupervised representation learning'.}

Neural network (NN) approaches have also been used for bottom-up phonetic discovery and representation learning.
An autoencoder (AE) is a feedforward NN where the target output of the network is equal to its input~\citep[$\S$4.6]{bengio_ftml09}, so it can be trained unsupervised (see Section~\ref{sec:cae_ae}).
AEs can be stacked to form deep networks, and this has proved useful for general unsupervised machine learning tasks like dimensionality reduction~\citep{hinton+salakhutdinov_science06} and pretraining~\citep{bengio+etal_nips07}. 
\cite{zeiler+etal_icassp13} was the first to apply stacked AEs directly to unlabelled speech; they showed that many of the AE filters correspond visually to ground truth phonetic units.
\cite{badino+etal_icassp14,badino+etal_interspeech15} also used stacked AEs, but with the explicit aim of finding a discrete categorical representation (i.e.\ phonetic discovery).
Their approach involves thresholding hidden activations to get a binary representation which is then used to find discrete clusters of subword units.
Such a discrete tokenized representation is useful since many down-stream tasks require categorical input.
However, based on an evaluation using the ABX task, \cite{badino+etal_interspeech15} found that discrete representations are often less accurate at phonetic discrimination
than some of the NN-based continuous vector representations described in the next section.

\subsection{Top-down approaches}
\label{sec:background_topdown}

The above approaches aim to discover phonetic units or representations without regard to longer-spanning word- or phrase-like patterns in the data.
Knowledge of such patterns could be used as a weak top-down supervision signal to guide subword discovery.
In several recent studies, UTD has been used to provide such top-down constraints.

An early approach from \cite{jansen+church_interspeech11} involved training whole-word HMMs on discovered words; similar HMM states are then clustered to automatically find subword unit models.
A useful property of this approach is that speaker-independence at the whole word level implies speaker-independence at the subword level.\footnote{This implies, of course, that the UTD system is tasked with finding speaker independent clusters. Fortunately, this problem has received significant attention in the UTD literature~\citep{zhang+glass_icassp10,jansen+etal_interspeech10,jansen+vandurme_asru11}.}
Their approach outperformed standard perceptual linear prediction (PLP) features in a multi-speaker same-different evaluation.
Since their approach only uses the discovered word examples for parameter estimation, much of the input speech data is disregarded.

\begin{figure}[tbp]
    \centering
    \includegraphics[trim=4cm 21cm 4cm 2.5cm,clip=true,scale=1]{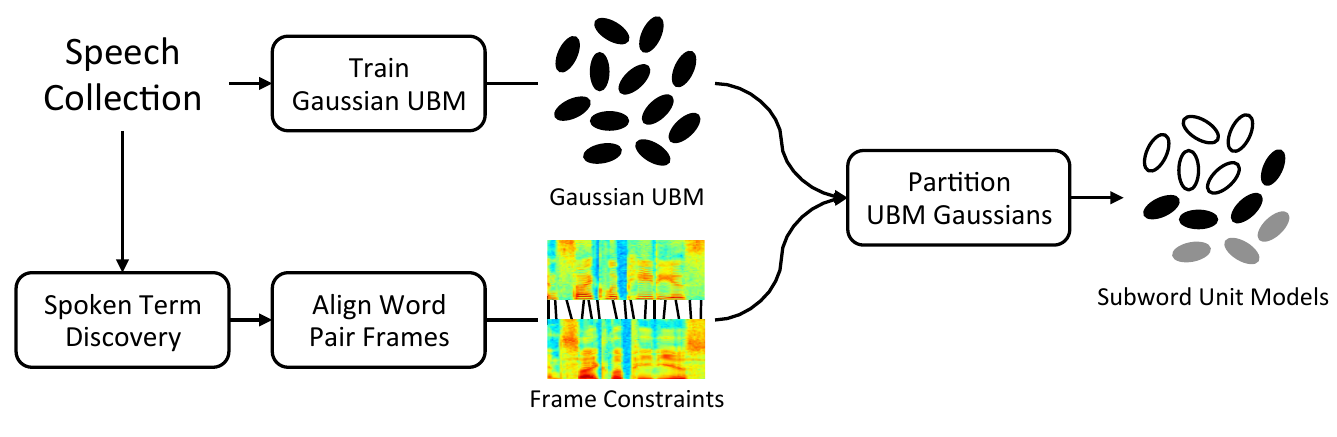}
    \caption{The unsupervised subword acoustic model training algorithm of~\citep{jansen+etal_icassp13b}. Top-down information is gained from frame-level alignments of different instances of the same discovered word type, and this is used to partition a GMM-UBM.}
    \label{fig:jansen+etal_icassp13b}
\end{figure}

This was was addressed in~\citep{jansen+etal_icassp13b}, using the approach illustrated in Figure~\ref{fig:jansen+etal_icassp13b}.
First, a GMM is trained bottom-up on a speech corpus, providing a UBM that takes into account all the speech data. UTD then
finds reoccurring words in the corpus. For each pair of word segments of the same type, frames are aligned using DTW. Based on the idea that different realizations of the same word should have a similar underlying subword sequence, UBM components in matching frames are attributed to the same subword unit. The resulting partitioned UBM is a type of unsupervised acoustic model where every partition corresponds to a subword unit. In the same-different task, posteriorgrams calculated over the partitioned UBM significantly outperformed the original features.\footnote{In both~\citep{jansen+church_interspeech11} and~\citep{jansen+etal_icassp13b}, UTD was simulated by sampling segments from ground truth forced alignments.}
This clearly illustrates the benefit of combining bottom-up background modelling (or initialization) on the lower-level frame-wise features with top-down knowledge of longer-spanning word patterns in the data.

This same idea has since been used in several zero-resource speech studies.
This includes our own work in Chapter~\ref{chap:cae}, where we introduce the {correspondence autoencoder}: an AE-like NN that takes input-output pairs of matching frames from word pairs discovered through UTD~\citep{kamper+etal_icassp15}.
At the same time that this model was being developed, \cite{synnaeve+etal_slt14} were developing another NN approach based on Siamese networks: tied networks that take in pairs of speech frames and minimize or maximize a distance depending on whether a pair comes from the same or different word classes (as predicted by UTD).
In~\citep{renshaw+etal_interspeech15} and~\citep{thiolliere+etal_interspeech15}, both approaches performed well in a multi-speaker minimal-pair phone trigram discrimination task, with the Siamese approach performing better in most cases.
For both models, gains over traditional acoustic features were particularly high when evaluating representations across speakers.
These results are discussed in more detail at the closing of Chapter~\ref{chap:cae}.


In summary, the studies discussed in this section indicate that there is much to gain from using top-down knowledge for unsupervised representation learning.
At the same time, the approach of~\cite{jansen+etal_icassp13b} would not have been possible without bottom-up background modelling; in our own work in Chapter~\ref{chap:bucktsong} we also find that bottom-up initialization is crucial.
The results from~\citep{jansen+church_interspeech11,renshaw+etal_interspeech15,thiolliere+etal_interspeech15}, together with the results of Chapter~\ref{chap:bucktsong}, also suggest that top-down constraints are especially useful for dealing with data from multiple speakers.

\section{Cognitive models of language acquisition}

Most of the studies described so far come from the speech processing community.
This section describes studies from the cognitive modelling community, where researchers are interested in how human infants acquire their native language.

\subsection{Word segmentation of symbolic input}
\label{sec:background_symbolic_wordseg}

Cognitive scientists have long been interested in how infants
learn to segment words and discover the lexicon of their native
language, with computational models seen as one way to specify and
test particular theories; see \citep{goldwater+etal_cognition09,rasanen_speechcom12}  for reviews.
In this community, most computational models of word segmentation
perform full-coverage segmentation of the data, breaking up the entire input into a sequence of
words (in contrast to UTD systems).
However,
these models generally take phonemic or phonetic strings as input,
rather than continuous~speech.

Early word segmentation approaches using phonemic input include those based
on transition probabilities~\citep{brent_ml99}, neural networks~\citep{christiansen+etal_lcp98} and
probabilistic models~\citep{venkataraman_cl01}.
\cite{goldwater+etal_cognition09} proposed a non-parametric Bayesian approach which outperformed previous work.
Their approach learns a language model over the tokens in
its inferred segmentation, incorporating priors that favour predictable
word sequences and a small vocabulary.
They experimented with learning either a unigram or bigram language model, and found that the proposed boundaries of both models were very accurate, but the unigram model proposed too few boundaries.
The original method uses a Gibbs sampler to sample individual boundary
positions; in follow-on work, \cite{mochihashi+etal_acl09} developed a blocked sampler that uses dynamic programming
to resample the segmentation of a full utterance at once (the sampler in Chapters~\ref{chap:tidigits} and~\ref{chap:bucktsong} is based on this work).
Goldwater et al.'s original model assumed that every instance of a
word is represented by the same sequence of phonemes; several later
studies proposed noisy-channel extensions using finite state transducers in order to deal
with variation in word pronunciation~\citep{neubig+etal_interspeech10,elsner+etal_emnlp13,heymann+etal_asru13}. 

Although these word segmentation models take discrete symbols as input, they do perform full-coverage segmentation.
One approach for dealing with continuous speech input would be to perform categorical phonetic discovery (Section~\ref{sec:background_phondisc}) and then apply a symbolic segmentation model to the tokenized output.
This was exactly the approach followed at the 2012 JHU CSLP workshop on zero-resource speech technology~\citep{jansen+etal_icassp13}.
Different combinations of phonetic discovery and word segmentation models were considered.
The attendees found, however, that the unsupervised tokenized speech was too noisy for subsequent word segmentation: the segmentation models struggled to discover word categories because of the large variation in the tokenization of different instances of the same word type.
Models making a unigram assumption of word predictability therefore performed poorly, with the more complex bigram models performing even worse since they attempted to learn dependencies between words without having discovered the word categories in the first place.
This highlights the necessity for a form of joint modelling of phonetic discovery (or representation learning), word category assignment (clustering) and segmentation, and that it is crucial to solve these tasks before introducing a more complex (possibly joint) language model.

\subsection{Multi-modal language acquisition}
\label{sec:background_multimodal}

During early language development, infants have access to more than just the speech modality.
This has prompted researchers to consider discovery across multiple modalities.
\cite{aimetti+etal_interspeech10}, for example, considered audio-visual language acquisition using a variant of segmental DTW.
In an engineering setting, \cite{renkens+etal_slt14} and \cite{renkens+vanhamme_interspeech15} considered the problem where a robot is shown actions paired with spoken commands; the robot is then required to learn the command-vocabulary and map these to appropriate actions (without any prior supervision).
Although not exactly the same as the task we are interested in, we mention this line of research since it serves as further motivation for our own work; in Section~\ref{sec:conclusion_zerospeech} of the concluding chapter we note that the extension of our models to the multi-modal case should be considered in future work.
Typically, the extra grounding information makes multi-modal learning easier.
Improvements and ideas from zero-resource single-modal speech processing could therefore be carried over to the multi-modal case, as has already been illustrated in~\citep{sun+vanhamme_csl13} and \citep{renkens+vanhamme_interspeech15} where a zero-resource word discovery method was extended with weak grounding information.

\section{Full-coverage segmentation and clustering of speech}

UTD systems (Section~\ref{sec:background_utd}) aim to find isolated, repeated word segments, leaving much of the data as background.
Cognitive models (Section~\ref{sec:background_symbolic_wordseg}) perform full-coverage segmentation, but take symbolic sequences as input instead of continuous speech.
We are interested in full-coverage segmentation and clustering of raw continuous speech, where word boundaries and lexical categories are predicted for the entire input.
Several researchers share this goal, and recent studies are summarized below.
This is not an exhaustive review, but these studies in particular share characteristics and served as inspiration for the work in this thesis.

\subsection{Previous approaches}
\label{sec:background_fullcoverage}

\cite{sun+vanhamme_csl13} developed an approach based on non-negative matrix factorization (NMF).
NMF is a technique which allows fixed-dimensional representations of speech utterances (typically co-occurrence statistics of acoustic events) to be factorized into lower-dimensional parts, corresponding to phones~\citep{ogrady+pearlmutter_neurocomp08} or words~\citep{stouten+etal_ieee08}. 
In standard NMF, however, the ordering of these parts are not retained.
To capture temporal information, \cite{sun+vanhamme_csl13} incorporated NMF in a maximum likelihood training procedure for discrete-density HMMs.
They applied this approach to an eleven-word unsupervised connected digit recognition task using the TIDigits corpus.
They learnt 30 unsupervised HMMs, each representing a discovered word type. 
They found that the discovered word clusters corresponded to sensible words or subwords:
average cluster purity 
was around 85\%.
Although NMF itself relies on a fixed-dimensional representation (as the systems of Section~\ref{sec:background_segmental} do) the final HMMs of their approach still perform frame-by-frame modelling (as also in  the studies below).

\cite{chung+etal_icassp13} used an HMM-based approach which alternates between subword and word discovery.
Their system models discovered subword units as continuous-density HMMs and learns a lexicon in terms of these units by alternating between unsupervised decoding and parameter re-estimation.
For evaluation, the output from their unsupervised system was compared to the ground truth transcriptions and every discovered word type was mapped to the ground truth label that resulted in the smallest error rate.
This allowed their system to be evaluated in terms of unsupervised word error rate (WER); on a four-hour Mandarin corpus with a small vocabulary size of about 400, they achieved WERs of around 60\%.

In \citep[Ch.~3]{lee_phd14} and \citep{lee+etal_tacl15}, a non-parametric hierarchical Bayesian model for full-coverage speech segmentation was developed.
Using adaptor grammars (a generalized framework for defining such Bayesian models), an unsupervised subword acoustic model developed in earlier work~\citep{lee+glass_acl12}, described in Section~\ref{sec:background_bottomup}, was extended with syllable and word layers, as well as a noisy channel model for capturing phonetic variability in word pronunciations.
When applied to speech from single speakers in the MIT Lecture corpus, most of the words with highest TF-IDF scores were successfully discovered,
and Lee et al.\ showed that joint modelling of subwords, syllables and words improved term discovery performance and word boundary detection accuracy (reported in terms of $F$-score).
Although Bayesian models are useful for incorporating prior knowledge and for finding sparser solutions~\citep{goldwater+griffiths_acl07}, Lee et al.'s model still makes frame-level independence assumptions.


\cite{walter+etal_asru13} developed a fully unsupervised system for connected digit recognition using the TIDigits corpus.
As in~\citep{chung+etal_icassp13}, they followed a two-step iterative approach of subword and word discovery.
For subword discovery, speech is partitioned into subword-length segments and clustered based on DTW similarity.
For every subword cluster, a continuous-density HMM is trained.
Word discovery takes as input the subword tokenization of the input speech.
Every word type is modelled as a discrete-density HMM with multinomial emission distributions over subword units, accounting for noise
and pronunciation variation.
HMMs are updated in an iterative procedure of parameter estimation and decoding.
Eleven of the whole-word HMMs were trained, one for each of the digits in the corpus.
Using a random initialization, their system achieved an unsupervised WER of 32.1\%; using UTD~\citep{park+glass_taslp08} to provide initial word identities and boundaries, 18.1\% was achieved.
In a final improvement, the decoded output was used to train from scratch standard continuous-density whole-word HMMs. 
This led to further improvements by leveraging the well-developed HMM tools used for supervised speech recognition.
This study of Walter et al.\ shows that unsupervised multi-speaker speech
recognition on a small-vocabulary task is possible. It also provides
useful baselines on a standard dataset, and gives a reproducible evaluation method in terms of the standard WER.
We use their system as a baseline in Chapter~\ref{chap:tidigits}.

The full-coverage word segmentation system of \cite{rasanen+etal_interspeech15} relies on an unsupervised method that predicts boundaries for syllable-like units.
These syllable tokens are then $K$-means clustered using averaged Mel-frequency cepstral coefficients (MFCCs) as segmental acoustic embeddings (see Section~\ref{sec:downsampling}), and reoccurring syllable clusters are treated as words.
Clustering is performed on a per-speaker basis, and the number of clusters is set as a fixed proportion of the proposed syllable tokens.
Their system performed well in the lexical discovery track of the Zero Resource Speech Challenge (ZRS) at Interspeech 2015~\citep{versteegh+etal_interspeech15}, where a whole suite of evaluation metrics were used.
The explicit use of automatically discovered syllables as the minimal unit in their overall approach can be seen as one way to incorporate prior knowledge of speech into a zero-resource system.
We use their system as a baseline in Chapter~\ref{chap:bucktsong}.

\begin{figure}[tbp]
    \centering
    \begin{minipage}{0.36\textwidth}
        \centering
        \includegraphics[scale=0.6]{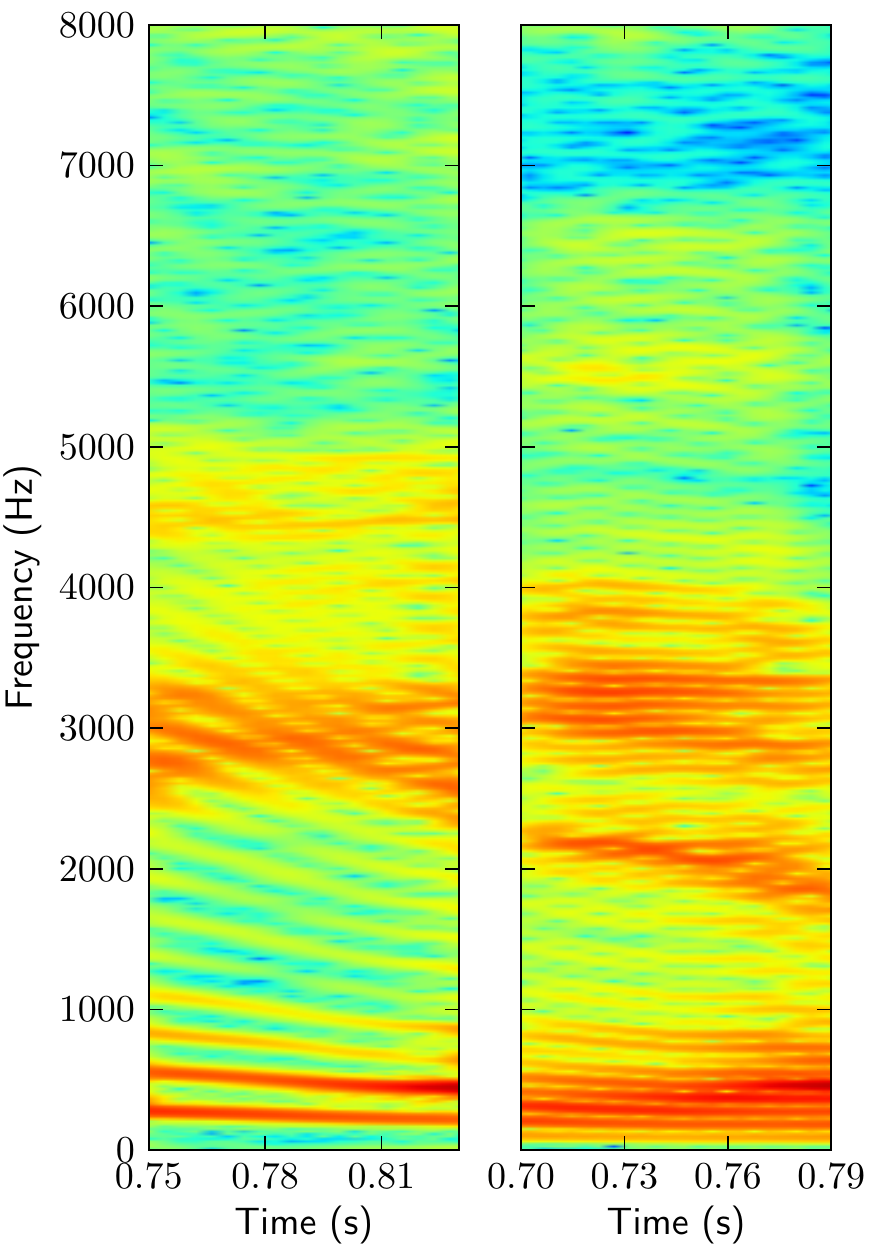} \\
        {\small (a)}
    \end{minipage}
    \qquad
    \begin{minipage}{0.56\textwidth}
        \centering
        \includegraphics[scale=0.6]{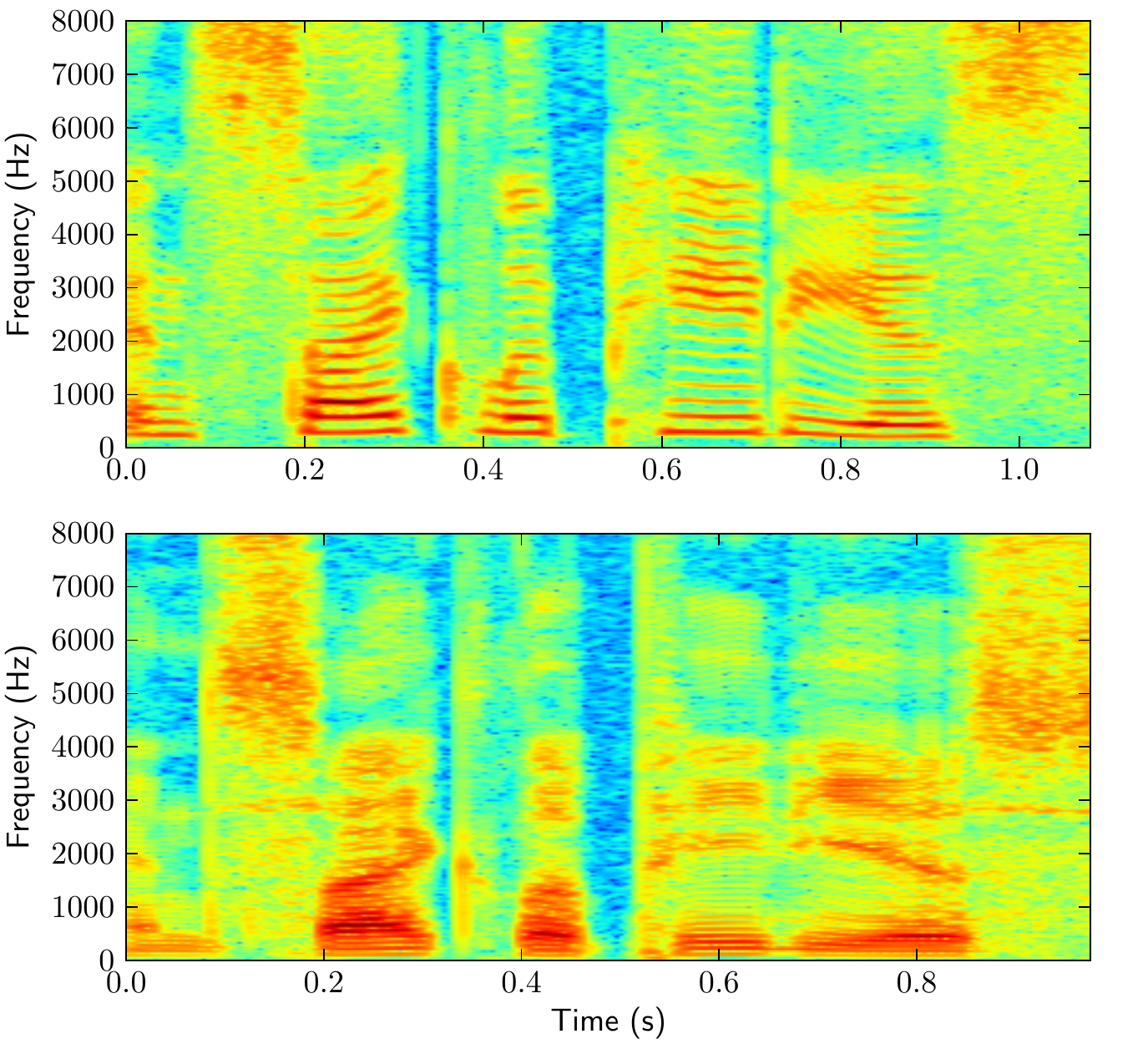} \\
        {\small (b)}
    \end{minipage}
     \caption{The spectrograms of (a) the phoneme /iy/ spoken by a female and male speaker and (b) the word `encyclopedias' spoken by the same female and male speakers.  The phoneme in (a) is the first /iy/ in the two word examples in (b).
     It is easier to identify cross-speaker similarities in the word examples than in the phoneme examples.}
     \label{fig:spectrogram}
\end{figure}

Apart from~\citep{sun+vanhamme_csl13}, the studies above all perform some form of explicit subword modelling, while the acoustic word embedding approaches of Section~\ref{sec:background_segmental} (which we also use) operate on fixed-dimensional embeddings of whole segments.
Direct modelling of larger segments has both advantages and disadvantages.
On the positive side, it is often easier to identify cross-speaker similarities between words than between subwords~\citep{jansen+etal_icassp13b}, which is why most UTD systems focus on longer-spanning patterns; Figure~\ref{fig:spectrogram}, for example, shows that it is much more difficult to find cross-speaker similarities between the two phones /iy/ in Figure~\ref{fig:spectrogram}(a) than between the words in which they occur, shown in Figure~\ref{fig:spectrogram}(b).
From a cognitive perspective, there is also evidence that infants are able to segment whole words from continuous speech before phonetic contrasts in their native language have been fully learned~\citep{bortfeld+etal_psychol05,feldman+etal_ccss09}.
On the other hand, direct segmental modelling makes it more difficult to explicitly include intermediate modelling layers (phones, syllables, morphemes) as in~\citep{lee_phd14,lee+etal_tacl15}.
Furthermore, segmental approaches are completely reliant on the
quality of the acoustic embeddings; for example, in the analysis of Section~\ref{sec:tidigits_scaling} we see that
the embedding function of \cite{levin+etal_asru13}, described in Section~\ref{sec:background_segmental}, deals poorly with very short speech segments.

\subsection{Evaluation of full-coverage segmentation and clustering systems}
\label{sec:background_eval}

In the above review of full-coverage systems, the evaluation method used in each of the studies was noted.
Metrics include average cluster purity, unsupervised WER, word boundary detection $F$-scores, and the set of 17 metrics used as part of the ZRS.
A lack of a standard set of metrics makes it very hard to compare different systems.
On the other hand, this lack is understandable since it is difficult to know upfront what the desired output of a zero-resource system should be.
As an example, we use WER in Chapters~\ref{chap:tidigits} and~\ref{chap:bucktsong}, because it is easily interpretable and well-known in the speech community.
However, WER makes the assumption that the discovered units correspond to words, and also penalizes results if multiple clusters contain tokens from the same ground truth word type, even if no other word types are found in those clusters.
Most of the other metrics have similar pros and cons (see the discussions in Sections~\ref{sec:tidigits_evaluation} and~\ref{sec:metrics}).
The suite of metrics used as part of the ZRS, which measures different aspects of zero-resource discovery systems~\citep{versteegh+etal_interspeech15}, is a step in the right direction (these are also used in Chapter~\ref{chap:bucktsong}), but because there are so many metrics, it becomes difficult to understand the relative performance of one system compared to another.
We note this issue of evaluation since it is an important aspect of zero-resource research, and we conclude in Chapter~\ref{chap:conclusion} that it warrants further investigation in future~work.

\section{Summary and conclusions}
\label{sec:background_conclusion}

From this review of previous work, we make three main conclusions that relate specifically to the task of full-coverage segmentation and clustering of speech.
Many of these arguments are also made in Daniel Renshaw's thesis~\citep{renshaw_masters16}.

Firstly, it would be reasonable to consider how modelling of whole speech segments (as is done in many of the systems in Section~\ref{sec:background_utd}) could be used for full-coverage speech segmentation.
In particular, fixed-dimensional acoustic word embeddings have been successfully used in unsupervised query-by-example search systems (Section~\ref{sec:background_segmental}), but have not been considered for full-coverage speech segmentation, where intermediate subword modelling is almost always used (Section~\ref{sec:background_fullcoverage}).
The use of such fixed-dimensional acoustic embeddings is especially attractive for unsupervised discovery since clustering can be performed directly in the embedding space.


Secondly, frame-level unsupervised representation learning incorporating both top-down and bottom-up learning methodologies should ideally be used.
Most full-coverage segmentation systems operate directly on traditional frame-level acoustic features (e.g.\ PLPs or MFCCs).
Unsupervised representation learning methods (Section~\ref{sec:background_phondisc}) aim to find a frame-level mapping from the original acoustic features to a new representation where it is easier to discriminate between subword or word units.
In particular, methods that combine top-down and bottom-up modelling have been shown to intrinsically outperform traditional acoustic features, especially in dealing with data from multiple speakers (Section~\ref{sec:background_topdown}).
Despite this, such representations have not been used in full-coverage speech segmentation systems.
Section~\ref{sec:background_symbolic_wordseg} drew the conclusion that representation learning (or phonetic discovery) should ideally be performed jointly with clustering and word segmentation.
The use of unsupervised representation learning methods that incorporate top-down knowledge from discovered words together with bottom-up information from lower-level features would be a move in this direction.

Thirdly, full-coverage speech segmentation could benefit from modelling context (i.e.\ language modelling), although previous studies suggest that there are challenges involved in doing so.
Word segmentation models that take symbolic input have been shown to benefit from the explicit modelling of word-word dependencies (Section~\ref{sec:background_symbolic_wordseg}).
However, the study conducted at the 2012 JHU CSLP workshop~\citep{jansen+etal_icassp13}, described at the end of Section~\ref{sec:background_symbolic_wordseg}, indicates that word category assignment (clustering) needs to be accurate enough in order to benefit from language modelling over the inferred categories.

\graphicspath{{cae/fig/}}

\chapter{Unsupervised representation learning using autoencoders}
\label{chap:cae}


This chapter introduces the \textit{correspondence autoencoder} (cAE), a novel unsupervised autoencoder-like deep neural network (DNN) for learning feature representations directly from unlabelled speech data.
The weak top-down supervision used for this network is obtained from a first-pass unsupervised term discovery system which finds pairs of isolated word examples of the same unknown type.
For each pair, dynamic programming is used to align the feature frames of the two words, and these are presented as input-output pairs to the cAE.
In an isolated word discrimination task that intrinsically evaluates the quality of speech representations, the cAE achieves large improvements over previous state-of-the art zero-resource representation learning methods. 
The results show that DNN-based feature extraction, which has proven so advantageous in supervised speech recognition, can also result in major improvements for unsupervised representation learning in the extreme zero-resource case.
This chapter is based on~\citep{kamper+etal_icassp15}, a publication resulting from a collaboration with Micha Elsner and my supervisors.

\section{Related work and intuition behind correspondence model}
\label{sec:cae_background}

Section~\ref{sec:background_phondisc} reviewed unsupervised representation learning methods.
The summary in Section~\ref{sec:background_topdown} of methods using weak top-down supervision is particularly relevant to the work presented in this chapter.
In the following we briefly mention those studies that directly inspired the cAE, and then outline the core idea behind the model.

\cite{zeiler+etal_icassp13} and \cite{badino+etal_icassp14} used stacked autoencoders (AEs), while \cite{zhang+etal_icassp12} used deep Boltzmann machines for unsupervised representation learning.
\edit{Even in the very early work of~\cite{elman+zipser_jasa88}---probably the first to apply AEs to speech---the potential of using AEs for unsupervised representation learning on unlabelled speech was noted.}
However, these are all purely bottom-up neural network (NN) approaches, which ignore longer-spanning patterns in the data.
In~\citep{jansen+church_interspeech11,jansen+etal_icassp13b}, it was shown that top-down knowledge of such patterns, potentially obtained from an unsupervised term discovery (UTD) algorithm, could be used in hidden Markov models (HMMs) and Gaussian mixture models (GMMs) to obtain better frame-level representations compared to purely bottom-up training.
The approach of~\cite{jansen+etal_icassp13b} is particularly attractive since it uses unsupervised bottom-up training on all the data to obtain a GMM-based universal background model (UBM), and then partitions the UBM based on mixture components which are active in aligned frames from discovered word pairs; this approach therefore combines bottom-up learning on a large corpus of raw speech with top-down learning using a smaller set of automatically discovered words.
See Figure~\ref{fig:jansen+etal_icassp13b} and the accompanying discussion for more details on this approach.

As in~\citep{jansen+etal_icassp13b}, which also took inspiration from some much earlier work~\citep{hunt+etal_icassp91}, the central idea of the new NN-based algorithm introduced in this chapter, is that aligned frames from different instances of the same word should contain information useful for finding a better feature representation.
Using layer-wise pretraining of a stacked AE, this approach uses a large corpus of untranscribed speech to find a suitable initialization; this is analogous to the bottom-up UBM trained in the first step of~\citep{jansen+etal_icassp13b}.
As in that work, word pairs discovered using UTD are then DTW-aligned to obtain frame-level constraints, which are presented as input-output pairs to the AE-like model.
We refer to this DNN, trained using weak top-down constraints, as a \textit{correspondence autoencoder} (cAE).
The cAE is therefore trained as a mapping between aligned frames, with intermediate layers (ideally) capturing common aspects of the underlying subword unit from which the frames originate, while normalizing out aspects that are not common (e.g.\ speaker).
We use the cAE as an unsupervised feature extractor by taking the encoding from a middle layer.
Using the same-different word discrimination task, we compare the new cAE feature representations to the original input features, to features obtained from posteriorgrams over the partitioned UBM of~\cite{jansen+etal_icassp13b}, and to features from a standard stacked AE.
One shortcoming of~\citep{jansen+church_interspeech11,jansen+etal_icassp13b} is that the UTD-step was simulated by using gold standard word pairs extracted from transcriptions; here a real UTD system~\citep{jansen+vandurme_asru11} is used.

\section{Unsupervised training algorithm}

Below we first give a concise overview of standard AEs, how these can be stacked to form deep networks, and how stacked AEs  can be used to initialize supervised DNNs.
We then present the training algorithm of the cAE, a neural network using weak top-down supervision in the form of word pairs obtained from a UTD system.

\subsection{Autoencoders, pretraining and supervised DNNs}
\label{sec:cae_ae}

An AE is a feedforward NN where the target output of the network is equal to its input~\citep{rumelhart+etal_pdp86};~\citep[$\S 4.6$]{bengio_ftml09}.
A single-layer AE encodes its input $\vec{y} \in \mathbb{R}^D$ to a hidden representation $\vec{a} \in \mathbb{R}^{D'}$ using
\begin{equation}
    \vec{a} = s(\vec{W}^{(0)} \vec{y} + \vec{b}^{(0)}) \label{eq:ae1}
\end{equation}
where $\vec{W}^{(0)}$ is a weight matrix, $\vec{b}^{(0)}$ is a bias vector, and $s$ is a non-linear vector function ($\tanh$ in our case).
The output $\vec{z} \in \mathbb{R}^D$ of the AE is obtained by decoding the hidden representation using
\begin{equation}
    \vec{z} = \vec{W}^{(1)}\vec{a} + \vec{b}^{(1)} \label{eq:ae2}
\end{equation}
The network is trained using the backpropagation algorithm to achieve a minimum reconstruction error, typically using the loss function $||\vec{y} - \vec{z}||^2$ when dealing with real-valued data.

A deep network can be obtained by stacking several AEs, each AE-layer taking as input the encoding from the layer below it:
\begin{align}
    \vec{z}^{(l)} &= \vec{W}^{(l - 1)} \vec{a}^{(l - 1)} + \vec{b}^{(l - 1)} \label{eq:stacked1} \\
    \vec{a}^{(l)} &= s(\vec{z}^{(l)}) \label{eq:stacked2}
\end{align}
with $\vec{a}^{(0)} = \vec{y}$. 
This stacked AE is trained one layer at a time, each layer minimizing the loss of its output with respect to its input, i.e.\ for layer $l$ the loss $||\vec{a}^{(l - 1)} - \vec{z}^{(l + 1)}||^2$ is minimized.\footnote{There is a subtlety here. When training layer $l$, the encoding parameters $\vec{W}^{(l - 1)}$ and $\vec{b}^{(l - 1)}$ are updated together with the decoding parameters $\vec{W}^{(l)}$ and $\vec{b}^{(l)}$ from layer $l$'s output: $\vec{z}^{(l + 1)} = \vec{W}^{(l)} \vec{a}^{(l)} + \vec{b}^{(l)}$. When then going on to train the next layer $l + 1$, these \textit{decoding} parameters $\vec{W}^{(l)}$ and $\vec{b}^{(l)}$ from layer $l$ are discarded; instead, these parameters are now trained as the \textit{encoding} parameters of layer $l + 1$.  So in layer-wise pretraining of a stacked AE, the weights $\vec{W}^{(0)}$ and $\vec{b}^{(0)}$ are obtained when training layer $l = 1$, the weights $\vec{W}^{(1)}$ and $\vec{b}^{(1)}$ are obtained when training layer $l = 2$, and so on.}

AEs are often used for non-linear dimensionality reduction by having a hidden layer that is narrower than its input dimensionality~\citep{hinton+salakhutdinov_science06}.
Although AEs with more hidden units than the input are in principle able to learn the identity function to achieve zero reconstruction error, \cite{bengio+etal_nips07} found that in practice such networks often still learn a useful representation since early stopping provides a form of regularization.
In the experiments presented in this chapter, we see that such bottom-up stacked AEs provide a crucial initialization for the new AE-like network; the aim here is not dimensionality reduction, but to find a better feature~\mbox{representation.}

In a supervised setting, the above procedure is one form of {unsupervised bottom-up pretraining} of a DNN.
This is followed by {supervised fine-tuning} where an additional output layer is added to perform some supervised prediction task, resulting in a supervised feedforward DNN~\citep{bengio+etal_nips07}.
Unsupervised pretraining was one of the main contributing factors in the resurgence of supervised DNNs~\citep{hinton+salakhutdinov_science06,hinton+etal_neurocomp06,bengio+etal_nips07}, but the common consensus now is that, given enough training data, supervised DNNs can be trained from scratch using randomly initialized weights. 
Nevertheless, we show that in this unsupervised setting, bottom-up pretraining is crucial.

\subsection{The correspondence autoencoder (cAE)}
\label{sec:cae_cae}

\begin{figure}[tbp]
    \centering
    \centerline{\includegraphics[width=\linewidth]{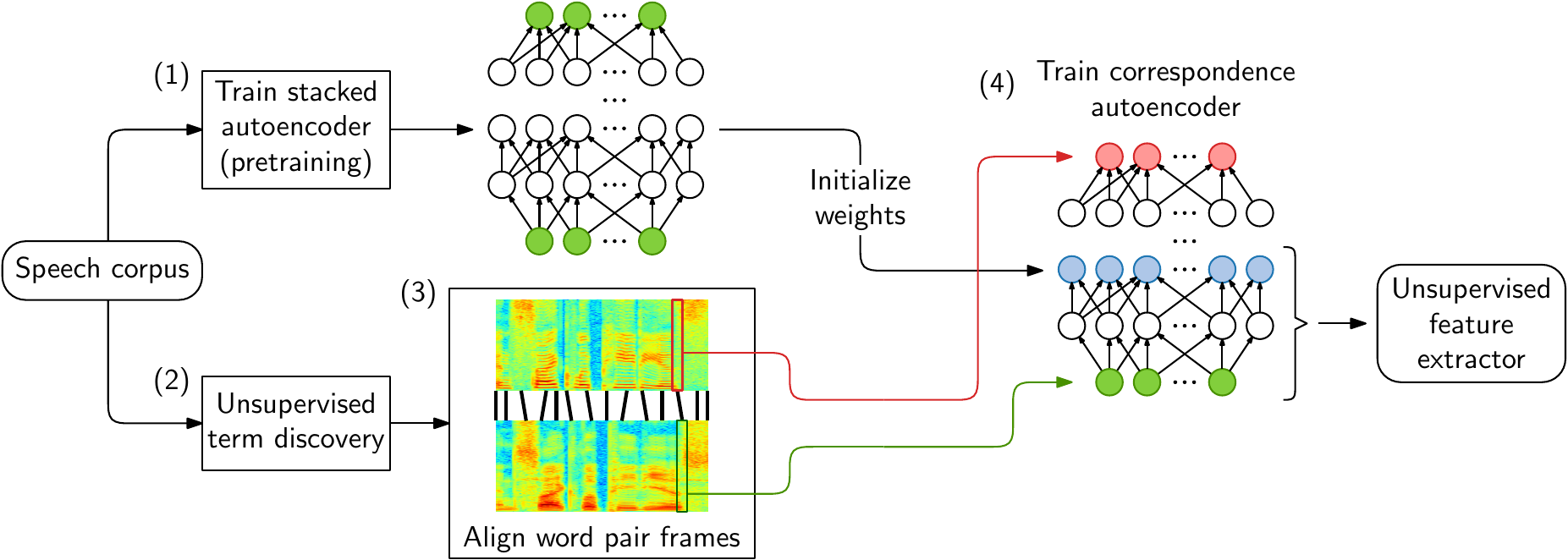}}
    \caption{Algorithm schematic for training the cAE for unsupervised feature extraction.
    The shaded green and red units indicate input and output features, while blue-shaded units indicate the new feature representation obtained from the final network.
    }
    \label{fig:training_algorithm}
\end{figure}

This section describes the novel training algorithm of the cAE.
While standard stacked AEs trained on speech, such as those of~\citep{zeiler+etal_icassp13,badino+etal_icassp14}, use the same feature frame(s) as input and output, the cAE uses weak top-down constraints in the form of (discovered) word pairs to have input and output frames from different instances of the same word.
The algorithm follows four steps, which are illustrated in Figure~\ref{fig:training_algorithm}.

\myemph{Step 1: Train a stacked AE.}
A corpus of speech is parametrized into the set $\mathcal{Y} = \{ \vec{y}_1, \vec{y}_2, \ldots, \vec{y}_T \}$, where each $\vec{y}_t \in \mathbb{R}^D$ is the frame-level acoustic feature representation of the signal (e.g.\ MFCCs).
Given $\mathcal{Y}$, a stacked AE is trained unsupervised directly on the acoustic features using backpropagation; the  network is constructed as in equations~\eqref{eq:stacked1} and~\eqref{eq:stacked2}. 
Using this network as initialization for the cAE, the model takes advantage of a large amount of untranscribed speech data to start at a point in weight-space where the network provides a representation close to the speech itself.

\myemph{Step 2: Unsupervised term discovery.}
A UTD system is run on the speech corpus.
This produces a collection of $N$ word segment pairs, which the model uses as weak top-down constraints.
In~\citep{jansen+etal_icassp13b}, this step was simulated by using gold standard word segment pairs extracted from forced alignments of the data. 
We present experiments both when using gold standard word pairs and when using pairs obtained from the UTD system of~\citep{jansen+vandurme_asru11}.

\myemph{Step 3: Align word pair frames.}
In the third step of the algorithm, the $N$ word-level constraints from UTD are converted to frame-level constraints.
For each word pair, a DTW alignment~\citep{sakoe+chiba_assp78} is performed using cosine distance as frame-wise similarity metric to find a minimum-cost frame alignment between the two words.
This is done for all $N$ word pairs, and taken together provides a set $\mathcal{F} = \{ (\vec{y}_{i,a}, \vec{y}_{i, b}) \}_{i = 1}^F$ of $F$ frame-level constraints.
Note that although each frame pair is unique, the time warping allowed in the alignment can result in the same frame occurring in multiple pairs.

\myemph{Step 4: Train the cAE.}
Using the stacked AE from step~1 to provide the initial weights and biases, the cAE is trained on the frame-level pairs $\mathcal{F}$.
For every pair $(\vec{y}_a, \vec{y}_b)$, $\vec{y}_a$ is presented as input to the network while $\vec{y}_b$ is taken as output: given input $\vec{y}_a$, the complete $K$-layer network is trained using backpropagation to minimize the loss $\left|\left|\vec{y}_b - \vec{z}^{(K + 1)}\right|\right|^2$, where $\vec{z}^{(K + 1)}$ is the final output of the network.\footnote{To obtain $\vec{z}^{(K + 1)}$, a final linear transformation as in~\eqref{eq:stacked1} is applied to the non-linearity $\vec{a}^{(K)}$ of the $K^{\text{th}}$ layer, so that the final network output $\vec{z}^{(K + 1)} \in \mathbb{R}^D$ matches the dimensionality of the target $\vec{y}_b$.}

Although we refer to the resulting network as an autoencoder to emphasize the relationship between its input and output, it can also be described differently.
Firstly, it can be seen as a type of denoising autoencoder~\citep{vincent+etal_icml08}, an AE were the input is corrupted  by adding Gaussian noise or setting some inputs to zero; this allows more robust features to be learnt.
In the cAE, the input $\vec{y}_a$ can be seen as a corrupted version of output $\vec{y}_b$, but instead of artificially corrupting the input, the noise is obtained automatically from the data.
Secondly, the cAE can also be described as a  standard DNN with a linear regression output layer, initialized using layer-wise pretraining.
Sometimes the term DNN is associated specifically with a supervised prediction task, and our network can be seen as predicting $\vec{y}_b$ when presented with input $\vec{y}_a$.

The aim here is to use the cAE as an unsupervised feature extractor that provides better word-discrimination properties than the original input features.
To use it as such, the encoding obtained from one of its middle layers is finally taken as the feature representation of new input speech, as illustrated in the right-most block of Figure~\ref{fig:training_algorithm}.

The core idea behind the cAE is that it learns useful features by performing a mapping between aligned frames, capturing the common underlying characteristics shared in the frames while normalizing out aspects which are not common (Section~\ref{sec:cae_background}).
As an example, some of the pairs presented to the cAE will come from different speakers, meaning that the model will need to reconstruct a frame from one speaker when presented with a frame from another.
The hope is that the cAE would learn a representation that is then abstracted away from aspects that are not common (such as the particular speaker).
The evaluation below (Section~\ref{sec:cae_samedifferent}) mainly considers how discriminative the resulting representations are across different speakers, so this particular aspect is evaluated directly in the word discrimination task.

\section{Experiments}

\subsection{Experimental setup}
\label{sec:cae_experimental_setup}

We use data from the Switchboard corpus of English conversational telephone speech.
Although English is not a low- or zero-resource language, we use this data in order to compare to previous studies.
Using HTK~\citep{htk_book}, data is parameterized as Mel-frequency cepstral coefficients (MFCCs) with first  and second order derivatives, yielding 39-dimensional feature vectors.
Cepstral mean and variance normalization (CMVN) is applied per conversation side.

For training the stacked AE (step 1), 180 conversations are used, which corresponds to about 23~hours of speech.
This same set was used for UBM training in~\citep{jansen+etal_icassp13b}.
For experiments using gold standard word pairs, we use the set used in~\citep{jansen+etal_icassp13b} for partitioning the UBM; it consists of word segments of at least 5 orthographic characters and 0.5 seconds in duration extracted from a forced alignment of the transcriptions of the 23~hour training set.
The full gold standard set consists of nearly $N=100\text{k}$ word segment pairs, comprised of about 105 minutes of speech.
About 3\% of these pairs are same-speaker word pairs.
DTW alignment of the 100k pairs (step 3) provides a frame-level constraint set of about $F = 7\text{M}$ frames, on which the cAE is trained (step 4).
In the truly unsupervised setup, we use word pairs discovered using the UTD system of~\citep{jansen+vandurme_asru11}.
We consider two sets.
The first consists of about $N=25\text{k}$ word pairs obtained by searching the above 23 hour training set.  About 17\% of these pairs are produced by the same speaker.
The second set consists of about 80k pairs obtained by including an additional 180 conversations in the~search.
About 11\% of these are same-speaker~pairs.


All DNNs are trained with minibatch stochastic gradient descent using Pylearn2~\citep{goodfellow+etal_arxiv13}.
A batch size of 256 is used, with 30 epochs of pretraining (step 1) at a learning rate of $250\cdot10^{-6}$ and 120 epochs of cAE training (step~4) at a learning rate of $2 \cdot 10^{-3}$.
Initially these parameters were set to the values given in~\citep{weng+etal_icassp14}, and were then adjusted based on training set loss function curves and development tests.
Although it is common to use nine or eleven sliding frames as input to DNN speech recognition systems, we use single-frame input here.
This was also done in~\citep{badino+etal_icassp14}, and allows for fair comparison with previous work.

\subsection{Same-different evaluation}
\label{sec:cae_samedifferent}

Our goal is to show the suitability of features from the cAE in downstream zero-resource search and recognition tasks.
We therefore use a multi-speaker word discrimination task developed specifically for this purpose~\citep{carlin+etal_icassp11}.
The \textit{same-different task} (which we have already summarized in Section~\ref{sec:background_phondisc_eval} but recap here) quantifies the ability of a speech representation to associate words of the same type and to discriminate between words of different types.
For every word pair in a test set of pre-segmented words, the DTW distance is calculated using the feature representation under evaluation. Two words can then be classified as
being of the same or different type based on some threshold,
and a precision-recall curve is obtained by varying the threshold.
To evaluate representations
across different operating points, the area under the precision-recall curve is calculated
to yield the final evaluation metric, referred to as the average precision~(AP). 

We use the same test set for the same-different task as that used in~\citep{jansen+etal_icassp13b}.
It consists of about $11\text{k}$ word tokens drawn from a portion of Switchboard distinct from any of the above sets.
The set results in $60.7\text{M}$ word pairs of which 96k are from the same word type.
Of these 96k pairs, only about $3\%$ were produced by the same speaker.
Additionally, we also extracted a comparable 11k-token development set, again from a disjoint portion of Switchboard.
Since tuning the hyperparameters of a NN is often an art, we present performance on the development set when varying hyperparameters.

Since we share a common test setup, we can compare the cAE feature representation directly to previous work~\citep{carlin+etal_icassp11,jansen+etal_icassp13b}. 
As a first baseline we use MFCCs directly to perform the same-different task.
We then compare our model to the partitioned UBM of~\cite{jansen+etal_icassp13b}, as described in Section~\ref{sec:cae_background}, and to
the supervised NN systems of~\cite{carlin+etal_icassp11}.
These {latter single-layer multistream NNs} were trained to estimate phone class posterior probabilities on transcribed speech data from the Switchboard and CallHome corpora, and have a comparable number of parameters to our networks.
We consider systems trained on 10 and 100 hours of speech.
For the partitioned UBM and NNs, test words are parameterized by generating posteriorgrams over components/phone classes, and symmetrized Kullback-Leibler divergence is used as frame-level metric for the same-different task.
For MFCCs and the AE- and cAE-based features, cosine distance is used.

\subsection{Choosing the network architecture}

Choosing the hyperparameters of NNs is challenging.  We therefore describe the optimization process followed on the development data.

\begin{figure}[!t]
    \centering
    \centerline{\includegraphics[scale=0.825]{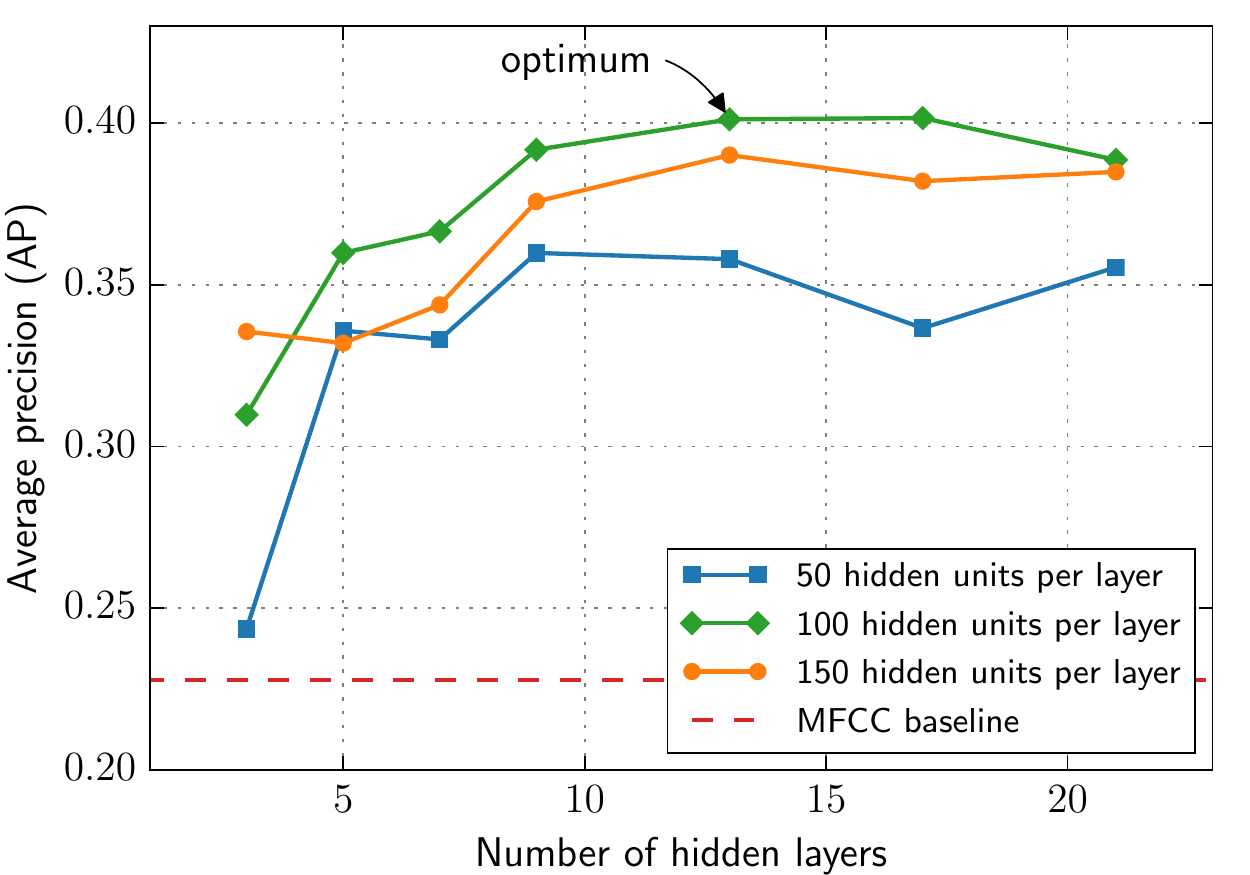}}
    \caption{Average precision (AP) on the development set for cAEs with varying numbers of hidden layers and units.
    In each case the best hidden layer on the development set was used.
    }
    \label{fig:ae100k_units_layers_dev}
\end{figure}

To use the cAE as feature extractor, the encoding from one of its middle layers is taken.
We found that using features from between the fourth-last to second-last encoding layers gave robust performance.
It is common practice to use a narrow bottle-neck layer to force the network to learn a lower-dimensional encoding at a particular layer.
We experimented with this, but found that performance was similar or slightly worse in most cases and therefore decided to only vary the number of hidden layers and units.
We do use a bottleneck layer for the cAE used in Chapter~\ref{chap:bucktsong}, but there it is because a specific encoding dimensionality is required.

We experimented with cAEs ranging from 3 to 21 hidden layers with 50, 100 and 150 hidden units per layer trained on the $100\text{k}$ gold standard word-pair set.
AP performance on the development set is presented in Figure~\ref{fig:ae100k_units_layers_dev}.
On this set, all networks achieve performance greater than that of the input MFCCs.
For all three hidden unit settings, performance is within $12\%$ relative to the respective optimal settings for networks with 7 to 21~layers.

\subsection{Gold standard weak top-down constraints}
\label{sec:cae_gold}

Table~\ref{tbl:100k_results} shows the AP performance on the test set using the baseline MFCCs, the UBM models from~\citep{jansen+etal_icassp13b}, the \mbox{(c)AE} networks, and the supervised NNs from~\citep{carlin+etal_icassp11}.
The partitioned UBM and the cAE were both trained on the gold standard $100\text{k}$ word-pair set.
The optimal cAE network on the development set (Figure~\ref{fig:ae100k_units_layers_dev}) is used here, taking its representation from the second-to-last encoding layer.
The second-to-last encoding layer was also used for obtaining representations from the stacked AE.

As reported before, although the UBM alone does not yield significant gains, the 100-component partitioned UBM results in a 34\% relative improvement over the baseline MFCCs.
Analogous to the UBM, the stacked AE alone also produces no improvement over the MFCCs.
This contrasts with the results reported in~\citep{badino+etal_icassp14}, where small improvements were obtained.
However,~\cite{badino+etal_icassp14} used much smaller training and test sets, had a different training setup, and had the explicit aim of tokenizing speech into categorical subword-like units rather than unsupervised feature extraction.

Without initializing the weights from the stacked AE, very poor performance is achieved by the cAE (0.024 AP).
However, when bottom-up pretraining is used, the resulting cAE outperforms the partitioned UBM by 64\% relative, and more than doubles the performance of the original MFCC features.
The cAE and partitioned UBM use exactly the same data for bottom-up initialization and weak top-down supervision.
The improvement of the cAE (0.469 AP) over the partitioned UBM (0.286) therefore indicates that the neural network method is much better able to exploit the information gained from the top-down constraints and bottom-up initialization than the GMM-based~model.

\begin{table}[tbp]
    \mytable
    \caption{Average precision (AP) on the test set using MFCCs, the UBM and partitioned UBM, the stacked and correspondence AEs trained on the 100k gold standard word pairs, and supervised NNs.
    Best performance when using weak top-down constraints is highlighted.}
    \begin{tabularx}{0.918\linewidth}{@{}Lc}
        \toprule
        Features  & AP \\
        \midrule
        MFCC with CMVN          & 0.214 \\
        \addlinespace
        UBM with 1024 components~\citep{jansen+etal_icassp13b} & 0.222 \\
        1024-UBM partitioned into 100 components~\citep{jansen+etal_icassp13b} & 0.286 \\
        \addlinespace
        100-unit, 13-layer stacked AE                           & 0.215 \\
        100-unit, 13-layer correspondence AE, no pretraining   & 0.024 \\
        100-unit, 13-layer correspondence AE, pretraining      & \textbf{0.469} \\
        \addlinespace
        English NN, 10 hours~\citep{carlin+etal_icassp11}        & 0.439 \\
        English NN, 100 hours~\citep{carlin+etal_icassp11}       & 0.516 \\
        \bottomrule
    \end{tabularx}
    \label{tbl:100k_results}
\end{table}

The cAE also outperforms the 10-hour supervised NN on this task, and comes close to the level of the 100-hour system.
Since the cAE use gold standard word pairs here comprised of only 105 minutes of speech, these results are potentially significant from a low-resource perspective.
Although these improvements are surprising, the form of explicit pair-wise supervision provided to the cAE is closely related to the word discrimination task.
Furthermore, these supervised baselines are single-layer NNs and could probably be improved using some of the more recent supervised DNN-based speech recognition approaches.

As in~\citep{jansen+etal_icassp13b}, to investigate dependence on the amount of supervision, we varied the number of gold standard word-pair constraints $N=100\text{k}, 10\text{k}, 1\text{k}\text{ and } 100$ by taking random subsets of the full 100k set; consequently, the number of frame-level constraints $F$ is varied. Results are shown in Table~\ref{tbl:less_pairs_results}.
For every set, the cAE was optimized on the development data.
In all cases the cAE outperforms the partitioned UBM and the baseline MFCCs.
With as few as $1\text{k}$ word pairs, the cAE gives the same performance as the partitioned UBM trained with all pairs.

\begin{table}[tbp]
    \mytable
    \caption{Average precision (AP) on the test set using the partitioned UBM~\citep{jansen+etal_icassp13b} and correspondence AEs when varying the number of gold standard word pairs $N$, with $F$ the resulting number of frame pairs.}
    \begin{tabularx}{0.7\linewidth}{ccCC}
        \toprule
        $N$ & $F$ & Partitioned UBM \par AP
        & Correspondence AE AP \\
        \midrule
        $10^5$ & $7\cdot 10^6$ & 0.286 & 0.469 \\
        $10^4$ & $7\cdot 10^5$ & 0.284 & 0.385 \\
        $10^3$ & $7\cdot 10^4$ & 0.266 & 0.286 \\
        $10^2$ & $7\cdot 10^3$ & 0.206 & 0.259 \\
        \bottomrule
    \end{tabularx}
    \label{tbl:less_pairs_results}
\end{table}

\subsection{Weak top-down constraints from unsupervised term discovery}
\label{sec:cae_utd}

\begin{table}[bp]
    \mytable
    \caption{Average precision (AP) on the test set when using weak top-down constraints from unsupervised term discovery (UTD). The number of word pairs $N$ and pair-wise accuracy of the UTD system is also shown. Best unsupervised performance is highlighted.}
    \begin{tabularx}{0.918\linewidth}{@{}lcCc}
        \toprule
        Features & $N$ & UTD Acc.\ (\%) & AP \\
        \midrule
        MFCC with CMVN & - & - & 0.214 \\
        100-unit, 13-layer stacked AE & - & - & 0.215 \\
        \addlinespace
        100-unit, 9-layer correspondence AE                       & 25k   & 46  & 0.339 \\
        100-unit, 13-layer correspondence AE                      & 80k   & 36  & \textbf{0.341} \\
        \addlinespace
        English NN, 10 hours~\citep{carlin+etal_icassp11}  & - & - & 0.439 \\
        English NN, 100 hours~\citep{carlin+etal_icassp11} & - & - & 0.516 \\
        \bottomrule
    \end{tabularx}
    \label{tbl:std_results}
\end{table}

Finally, we present truly unsupervised results where the UTD system of~\cite{jansen+vandurme_asru11} is used to provide the word pairs for weak supervision.
In this case, the cAE is truly used as an unsupervised frame-level representation learning method.
Results are shown in Table~\ref{tbl:std_results}, with some baselines repeated from Table~\ref{tbl:100k_results}.
Two UTD runs are used (Section~\ref{sec:cae_experimental_setup}), and Table~\ref{tbl:100k_results} includes their pair-wise accuracies; by mapping each discovered segment to the ground truth word label with which it overlapped most in the forced alignments, this accuracy is calculated as the proportion of correctly matching pairs.
The first UTD run produced 25k word pairs at an accuracy of 46\%, while the second (applied to additional unlabelled speech data) produced 80k pairs at 36\%.
Correspondence AEs were trained separately on the two sets of weak top-down constraints, with each optimized on the development data.
In both cases, representations from the second-to-last encoding layer were again used.

Both cAEs significantly outperform the MFCCs and stacked AE baselines by more than 57\% relative in AP, coming to within 23\% of the 10-hour supervised NN baseline.
Compared to the partitioned UBM trained on 100k gold standard word pairs (0.286 AP, Table~\ref{tbl:100k_results}), the completely unsupervised cAEs still perform better by almost 19\%, despite the much noisier form of weak supervision.
Performance of the best cAE from the gold standard word-pair case (0.469 AP, Table~\ref{tbl:100k_results}) relative to the best unsupervised cAE (0.341 AP, Table~\ref{tbl:std_results}), indicates that the noise introduced by the true UTD-step results in a penalty of 34\%; the unsupervised cAE nevertheless provides a better representation than the other unsupervised baselines.
It is unclear if the same will hold for the previous models~\citep{jansen+church_interspeech11,jansen+etal_icassp13b} where the truly zero-resource case was not considered.

A comparison of the two cAEs in Table~\ref{tbl:std_results} shows that, despite using significantly more pairs and allowing a deeper network to be trained, the 80k set does not provide a major improvement over the 25k set.
This is attributed to the lower word-pair accuracy of the former, and shows that there is a trade-off between UTD accuracy and the number of pairs produced.
Compared to the analysis in Table~\ref{tbl:less_pairs_results}, the 25k unsupervised-obtained pairs still provide more useful supervision than 1k gold standard word pairs.
A finer-grained investigation of the trade-off between the number of word pairs and accuracy, which can be varied by searching more data or by adjusting the search threshold, could be the focus of future work.

%

\section{Follow-on and related work by others}
\label{sec:cae_followup}

Before drawing final conclusions regarding the cAE, we briefly highlight follow-up work by others, as well as other recent zero-resource studies related to the cAE.
Although input was given on some of these studies, the research described here is that of others and should not be considered a contribution of this thesis.

Follow-on work has extended and applied the cAE in different evaluation settings and on data from different languages.
\cite{yuan+etal_interspeech16}, for instance, showed that bottleneck features obtained from a supervised DNN trained on data from high-resource languages (Spanish and Mandarin Chinese) can be used to improve the performance of the cAE on a target zero-resource language (they used Switchboard English, in order to compare to our work) by using the bottleneck features as input to the cAE.

\begin{figure}[!b]
    \centering
    \includegraphics[width=0.918\linewidth]{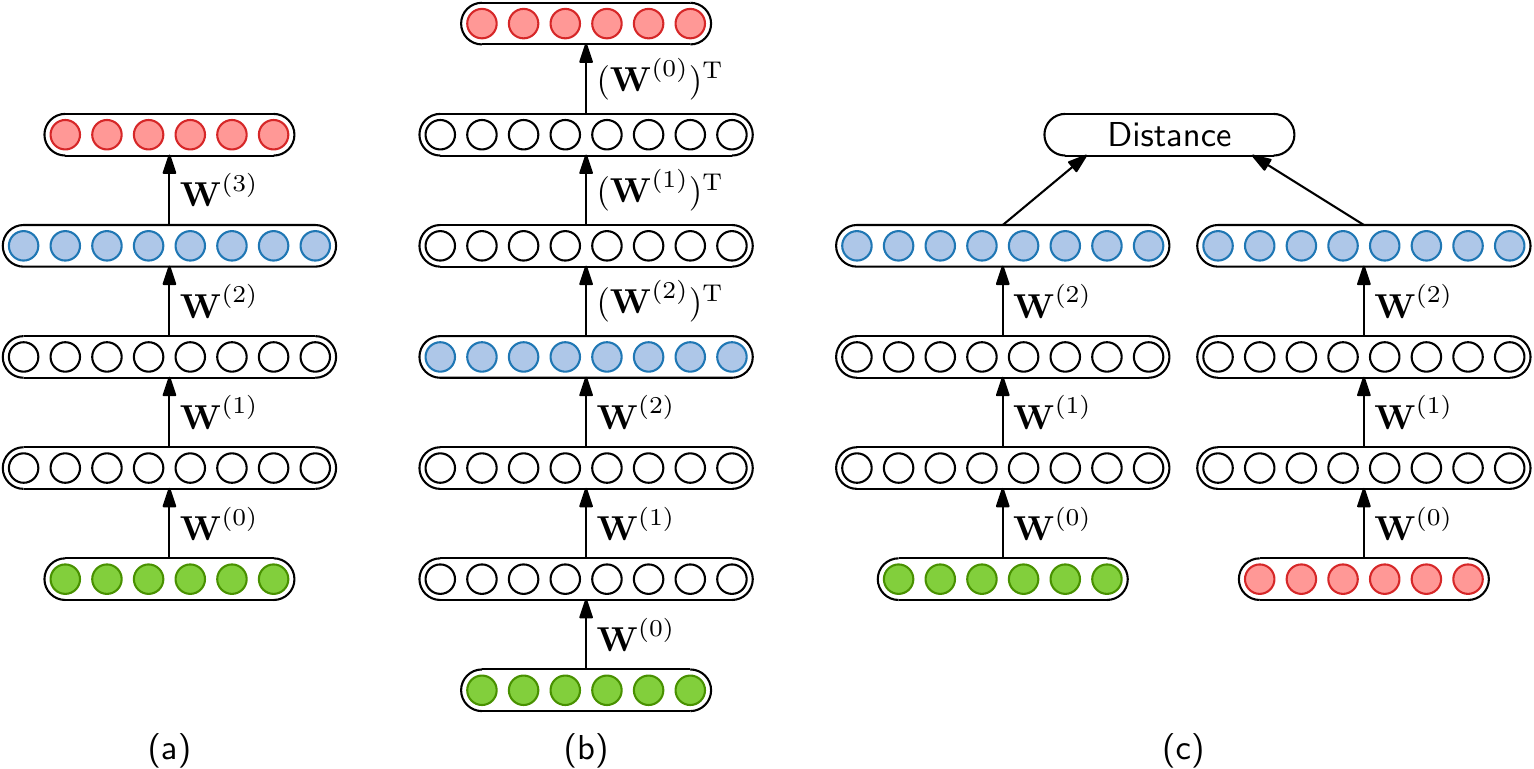}
    \caption{
    (a) The cAE as used in this chapter. The encoding layer (blue) needs to be chosen based on performance on a development set.
    (b) The cAE with symmetrical tied weights proposed by \cite{renshaw+etal_interspeech15}. The encoding from the middle layer (blue) is always used.
    (c) The siamese DNN proposed by \cite{synnaeve+etal_slt14}. The cosine distance between aligned frames (green and red) is either minimized or maximized depending on whether the frames belong to the same (discovered) word or not.
    }
    \label{fig:cae_siamese}
\end{figure}

As part of his MPhil thesis, \cite{renshaw_masters16} investigated the structure and training setup of the cAE, and also applied it to other datasets using a different evaluation framework.
This resulted in the publication~\citep{renshaw+etal_interspeech15}. 
In terms of structure, he proposed to tie weights from lower layers to those of higher layers in a symmetrical fashion, as shown in Figure~\ref{fig:cae_siamese}(b).
Although this does not improve performance, it removes the need for carefully choosing the hidden layer from which to take the output representations, as we did in this chapter using development data: see Figure~\ref{fig:cae_siamese}(a).
In~\citep{renshaw+etal_interspeech15}, the aligned frame pairs $(\vec{y}_a, \vec{y}_b)$ were also presented in both directions as input and output to the cAE, i.e.\ in this chapter, $\vec{y}_a$ is always used as input and $\vec{y}_b$ as output, while in~\citep{renshaw+etal_interspeech15}, these were also presented in the opposite direction.
This resulted in small improvements, and we do the same in cAE training in Chapter~\ref{chap:bucktsong}.
\cite{renshaw+etal_interspeech15} also found that layer-wise pretraining of the stacked AE can be done for far fewer epochs; in Chapter~\ref{chap:bucktsong} we use 5 epochs of pretraining instead of the 30 epochs used here.

More importantly, Renshaw et al.\ compared the cAE to more models using ABX evaluation on data from two languages: English and Xitsonga.
The ABX task (described in Section~\ref{sec:background_phondisc_eval}) is related to the same-different task used in this chapter, but considers discrimination of minimal-pair phone trigram segments instead of words.
As shown in Table~\ref{tbl:cae_abx}, Renshaw et al.\ found that, as in this chapter, cAE representations are more discriminative than MFCCs as well as representations from stacked AEs on both languages.
Importantly, they also showed that the cAE outperforms a denoising AE (dAE).
In Section~\ref{sec:cae_cae}, we described the cAE as a special kind of dAE: the cAE uses `realistic' noise obtained from data, instead of using artificial noise as in the standard dAE.
The results of~\citep{renshaw+etal_interspeech15} indicate that the data-driven noise used in the cAE yields more useful features than artificial noise.
The ABX evaluation in~\citep{renshaw+etal_interspeech15} was also performed separately on individual speakers and across speakers; Table~\ref{tbl:cae_abx} shows that the gains of the cAE across speakers are generally more pronounced than within speaker (where the baseline MFCC features already do quite well).
A further important aspect of \citep{renshaw+etal_interspeech15} is that all hyperparameter tuning for the cAE was carried out only on English data; these same hyperparameters were used directly in the Xitsonga cAE model.
The Xitsonga results in Table~\ref{tbl:cae_abx} therefore show that the cAE can be applied directly to a new language without requiring any resources for tuning, and still gives large improvements over the baseline approaches.

\begin{table}[tbp]
    \mytable
    \caption{ABX error rates (lower is better) of models by others~\citep{renshaw+etal_interspeech15,thiolliere+etal_interspeech15} on English and Xitsonga data. Both within-speaker and across-speaker discriminability are~shown.}
    \begin{tabularx}{\linewidth}{@{}lCCCC@{}}
        \toprule
        & \multicolumn{2}{c}{English ABX (\%)} & \multicolumn{2}{c}{Xitsonga ABX (\%)} \\
        \cmidrule(l){2-3} \cmidrule(l){4-5}
        Model & Within & Across & Within & Across \\
        \midrule
        MFCCs & 15.6 & 28.1 & 19.1 & 33.8 \\
        Stacked AE & 16.9 & 28.6 & 17.4 & 29.5 \\
        Denoising AE~\citep{renshaw+etal_interspeech15} & 15.8 & 25.3 & 15.8 & 25.9 \\
        Correspondence AE~\citep{renshaw+etal_interspeech15} & 13.5 & 21.1 & 11.9 & 19.3 \\
        Siamese DNN~\citep{thiolliere+etal_interspeech15} & 12.0 & 17.9 & 11.7 & 16.6 \\        
        \bottomrule
    \end{tabularx}
    \label{tbl:cae_abx}
\end{table}

As mentioned in Section~\ref{sec:background_topdown}, at the same time we were carrying out the work presented in this chapter, \cite{synnaeve+etal_slt14} were developing another NN approach using the same kind of top-down supervision.
Instead of using a reconstruction criterion as in the cAE, they used Siamese networks: tied networks that take pairs of speech frames and minimize or maximize a distance depending on whether they come from same or different word classes (as predicted by UTD).
This approach is illustrated in Figure~\ref{fig:cae_siamese}(c).
They used gold word pairs in~\citep{synnaeve+etal_slt14}, but follow-up work used UTD pairs~\citep{thiolliere+etal_interspeech15,zeghidour+etal_icassp16}.
In~\citep{thiolliere+etal_interspeech15}, ABX evaluation was performed on the same data as used by \cite{renshaw+etal_interspeech15}.
The last row of Table~\ref{tbl:cae_abx} indicates that in most cases the Siamese DNN performs better than the cAE.
The use of negative frame pairs from words that are predicted to be of different types might be giving the Siamese DNN extra information, resulting in improvements~\citep{versteegh+etal_sltu16}.
On the other hand, this can also be seen as an advantage of the cAE since the negative pairs for the Siamese model need to be chosen carefully~\citep{thiolliere+etal_interspeech15}.

\section{Summary and conclusions}

This chapter introduced a novel unsupervised frame-level representation learning method using an autoencoder-like deep neural network.
This model, referred to as the {correspondence autoencoder} (cAE), uses weak top-down supervision from word pairs obtained using an unsupervised term discovery (UTD) system, together with bottom-up initialization.
We evaluated the cAE using the same-different word discrimination task---an intrinsic evaluation of the quality of speech feature representations.
In order to compare to previous studies, we used English data from Switchboard.


In experiments where gold standard word pairs from transcriptions were used for weak supervision, we showed that the proposed cAE gives a 64\% relative improvement over previously reported state-of-the-art results using the same test setup.
To achieve this result, it was crucial to use unsupervised bottom-up pretraining together with the top-down supervision.
The cAE also outperformed a (weak) supervised baseline, despite using much less labelled data.
This could have major implications for low-resource speech recognition, where limited amounts of labelled data are available.
In particular, future work could consider low-resource supervised speech recognition using cAE features instead of standard acoustic features (see Section~\ref{sec:conclusion_future}).

In an unsupervised setup, UTD was used to provide the weak top-down constraints for the cAE.
In this setting, the cAE is used as a truly unsupervised frame-level representation learning method since it learns directly from unlabelled speech.
Here, the model outperformed both baseline MFCCs and a standard stacked autoencoder by more than 57\%, coming to within 23\% of the supervised system trained on 10 hours of transcribed speech.
This is a significant result for any downstream zero-resource speech processing task where transcriptions and pronunciation dictionaries are not available for system development.

Since publication of~\citep{kamper+etal_icassp15}, which first introduced the cAE and served as foundation for this chapter, others have extended and applied the cAE in different settings; the cAE proved effective in other intrinsic zero-resource evaluations~\citep{renshaw+etal_interspeech15,renshaw_masters16} and has been applied to different languages~\citep{yuan+etal_interspeech16}.
The analysis of \cite{renshaw+etal_interspeech15} suggests that the cAE is particularly effective in improving discriminability across speakers, and that the cAE can be applied to a new language without the need to tune hyper-parameters on within-language data.

All the results presented in this chapter, as well as those by others, have considered the intrinsic quality of the frame-level representations produced by the cAE.
In Chapter~\ref{chap:bucktsong} we use the cAE to provide input features for a complete zero-resource segmentation and lexicon discovery system.
There we see that the cAE also results in extrinsically better performance, in particular by improving speaker- and gender-independence when applied to unlabelled speech data from multiple speakers.

\graphicspath{{tidigits/fig/}}

\chapter{A segmental Bayesian model for small-vocabulary word segmentation and clustering}
\label{chap:tidigits}

Zero-resource speech processing is not only concerned with representation learning at the phone or frame level, but also with the discovery of longer-spanning word or phrase-like patterns in raw speech.
The tasks of segmenting and clustering unlabelled speech into meaningful units is essential in zero-resource systems that aim to analyze, search and summarize unlabelled speech data.
Similarly, realistic models of infant language acquisition would be required to break up speech into word-like segments, and to cluster these segments into hypothesized word (lexical) groupings.
Early zero-resource approaches focused on identifying isolated reoccurring terms in a corpus.
More recent full-coverage systems have attempted to completely segment and cluster speech audio into word-like units---effectively performing a type of unsupervised speech recognition.

This chapter introduces a novel unsupervised segmental Bayesian model for full-coverage word segmentation and clustering of unlabelled speech.
In this approach, a potential word segment (of arbitrary length) is embedded in a fixed-dimensional acoustic vector space. The model, implemented as a Gibbs sampler, then builds a whole-word acoustic model in this embedding space while jointly performing segmentation.
The result is a complete unsupervised tokenization of the input speech in terms of discovered word types.
This approach is distinct from any presented before.

In this chapter, we compare our approach to a more traditional HMM-based system of a previous study on a small-vocabulary English dataset of connected digit sequences.
By mapping the unsupervised decoded output to ground truth transcriptions, we report unsupervised word error rates (WERs).
The segmental Bayesian model achieves around 20\% WER in this small-vocabulary multi-speaker evaluation, outperforming the HMM-based system by about 10\% absolute.
Moreover, in contrast to the baseline, the segmental Bayesian model does not require a pre-specified vocabulary size.
The system presented here is applied to a small-vocabulary task since it allows for a thorough analysis of the discovered structures and pose less computational issues.
There are several challenges in directly applying this system to large-vocabulary speech; these challenges are addressed in the next chapter which presents our large-vocabulary~system.

Preliminary work for the research presented in this chapter is described in the two conference publications~\citep{kamper+etal_slt14} and~\citep{kamper+etal_interspeech15}.
These were extended and refined into the expanded journal publication~\citep{kamper+etal_taslp16}, on which this chapter is mainly based.

\section{Related work and comparison to proposed model}
\label{sec:tidigits_related_work}

Below we briefly refer back to the most relevant studies described in Chapter~\ref{chap:background}, and in particular discuss how previous work relates to the model which we introduce here.

\subsection{Discovery of words in speech}

Section~\ref{sec:background_sdtw} described studies on unsupervised term discovery (UTD), where the task is to find meaningful repeated word- or phrase-like patterns in raw speech audio.
Like our own system, many of these UTD systems operate on whole-word representations, with no subword level of
representation. However, each word is represented as a vector
time series with variable dimensionality (number of frames),
requiring dynamic time warping (DTW) for comparisons.
Section~\ref{sec:background_segmental} described recent studies in which variable-duration segments are mapped to fixed-dimensional acoustic word representations.
We follow such an acoustic word embedding approach here; we can then define an
acoustic model over these embeddings and make comparisons without requiring any alignment.
In addition, UTD systems aim to find and cluster repeated, isolated acoustic segments, leaving much of the input data as background. In contrast, we aim for full-coverage segmentation of the entire speech input into hypothesized words.

\subsection{Word segmentation of symbolic input}

Section~\ref{sec:background_symbolic_wordseg} described studies from the cognitive modelling community which performed full-coverage segmentation of data into a sequence of words.
However, these models generally take phonemic or phonetic strings as input, rather than continuous speech.

The model presented in this chapter is based (to a large extent) on the non-parametric Bayesian approach of \cite{goldwater+etal_cognition09}, which operates on transcribed phonemic sequences.
When first presented, their model was shown to yield more accurate segmentations than previous work.
Their approach learns a language model over the tokens in
its inferred segmentation, incorporating priors that favour predictable
word sequences and a small vocabulary.
The original method uses a Gibbs sampler to sample individual boundary
positions; the sampler we use for the model in this chapter is based on the later work of \cite{mochihashi+etal_acl09}, who presented a blocked sampler that uses dynamic programming
to resample the segmentation of a full utterance at once.

Goldwater et al.'s original model assumed that every instance of a
word is represented by the same sequence of phonemes; later
studies~\citep{neubig+etal_interspeech10,elsner+etal_emnlp13,heymann+etal_asru13} proposed noisy-channel extensions in order to deal
with variation in word pronunciation.  Our
model can also be viewed as a noisy-channel extension to the original
model, but with a different type of channel model. \edit{In~\citep{neubig+etal_interspeech10,elsner+etal_emnlp13,heymann+etal_asru13},}
variability is modeled symbolically as the conditional probability of
an output phone given the true phoneme (so the input to the models is
a sequence or lattice of phones), whereas the channel model used here is a true acoustic
model (the input is the speech signal).  

\subsection{Full-coverage segmentation of speech}
\label{sec:tidigits_related_studies_full_coverage}

Section~\ref{sec:background_fullcoverage} described several recent studies that share the goal of full-coverage word segmentation of speech.
Below we note two of these studies which have inspired our work in particular, and then compare our proposed approach to previous full-coverage systems in general.

\cite{lee+etal_tacl15} developed a non-parametric hierarchical Bayesian model for full-coverage speech segmentation.
As in their model, we also follow a Bayesian approach, which is useful for incorporating prior knowledge and for finding sparser solutions by using appropriate priors~\citep{goldwater+griffiths_acl07}.
However, our proposed model operates directly at the whole-word level instead of having a hierarchy of layers of words, syllables and subwords. 
In addition, in this chapter we evaluate on a small-vocabulary multi-speaker corpus rather than large-vocabulary single-speaker data.


The work that is most directly comparable to this chapter is that of \cite{walter+etal_asru13}, who developed a fully unsupervised HMM-based system for connected digit recognition using the TIDigits corpus.
Eleven  whole-word HMMs were trained, one for each of the digits in the corpus.
For evaluation, the output from their unsupervised system was mapped to the ground truth labels. 
Using a random initialization, their system achieved an unsupervised word error rate (WER) of 32.1\%; using UTD~\citep{park+glass_taslp08} to provide initial word identities and boundaries, 18.1\% WER was achieved.
In a final improvement, the decoded output was used to train from scratch standard 
whole-word HMMs. 
This led to further improvements by leveraging the well-developed HMM tools used for supervised speech recognition.
This study provides a
useful baseline on a standard dataset, and gives a reproducible evaluation method in terms of the standard WER.
Our model is comparable to Walter et al.'s word discovery system before the refinement using a traditional continuous-density HMM recognizer.
We therefore use the results they obtained before refinement as baselines in the experiments here.

In contrast to the work by \cite{lee+etal_tacl15}, \cite{walter+etal_asru13}, and most of the other full-coverage studies described in Section~\ref{sec:background_fullcoverage}~\citep{chung+etal_icassp13,rasanen+etal_interspeech15}, our model operates directly at the whole-word level instead of having both word and subword layers.
By taking this different perspective, this segmental whole-word approach 
is a complementary contribution to the field of zero-resource speech processing.
The approach is further motivated by the observation that it 
is often easier to identify cross-speaker similarities between words than between subwords~\citep{jansen+etal_icassp13b}, which is why most UTD systems focus on longer-spanning patterns (see Figure~\ref{fig:spectrogram}). There is also evidence that infants are able to segment whole words from continuous speech while still learning phonetic contrasts in their native language~\citep{bortfeld+etal_psychol05,feldman+etal_ccss09}.
A benefit of the segmental embedding approach we use is that segments can be compared directly in a fixed-dimensional embedding space, meaning that word discovery can be performed using standard clustering methods.
We use a Bayesian Gaussian mixture acoustic model for this purpose, which we compared to several other clustering methods in~\citep{kamper+etal_slt14} for the case where the true word segmentation was known, and found to be most accurate for clustering the acoustic word embeddings used here.

On the other hand, direct whole-word modelling makes it more difficult to explicitly include intermediate modelling layers (phones, syllables, morphemes) as Lee et al.\ did.
Furthermore, the proposed whole-word approach is completely reliant on the
quality of the embeddings; in Section~\ref{sec:tidigits_scaling} we see that
the embedding function used here deals poorly with short segments.
Improved embedding techniques are the subject of current research by several groups~\citep{kamper+etal_icassp16,chung+etal_arxiv16}, but this is a challenging problem in itself.
Nevertheless, the framework introduced here is not reliant on any particular embedding method: it would be straightforward to replace the current approach with any other (including one that incorporates subword modelling).

\section{The segmental Bayesian model}
\label{sec:tidigits_segmental_bayesian_model}

Below we first give an intuitive overview of our proposed segmental Bayesian model.
The different components are then described in more detail.

In our approach, any potential word segment (of arbitrary length) is mapped to a vector in a fixed-dimensional space $\mathbb{R}^D$.
The goal of this \textit{acoustic word embedding} procedure is that word instances of the same type should lie close together in this space.
The different hypothesized word types are then modelled in this $D$-dimensional space using a Gaussian mixture model (GMM) with Bayesian priors.
Every mixture component of the GMM corresponds to a discovered type; the component mean can be seen as an average embedding for that word.
However, since the model is unsupervised, we do not know the identities of the true word types to which the components correspond.

\begin{figure}[tbp]
    \centering
    \includegraphics[width=\linewidth]{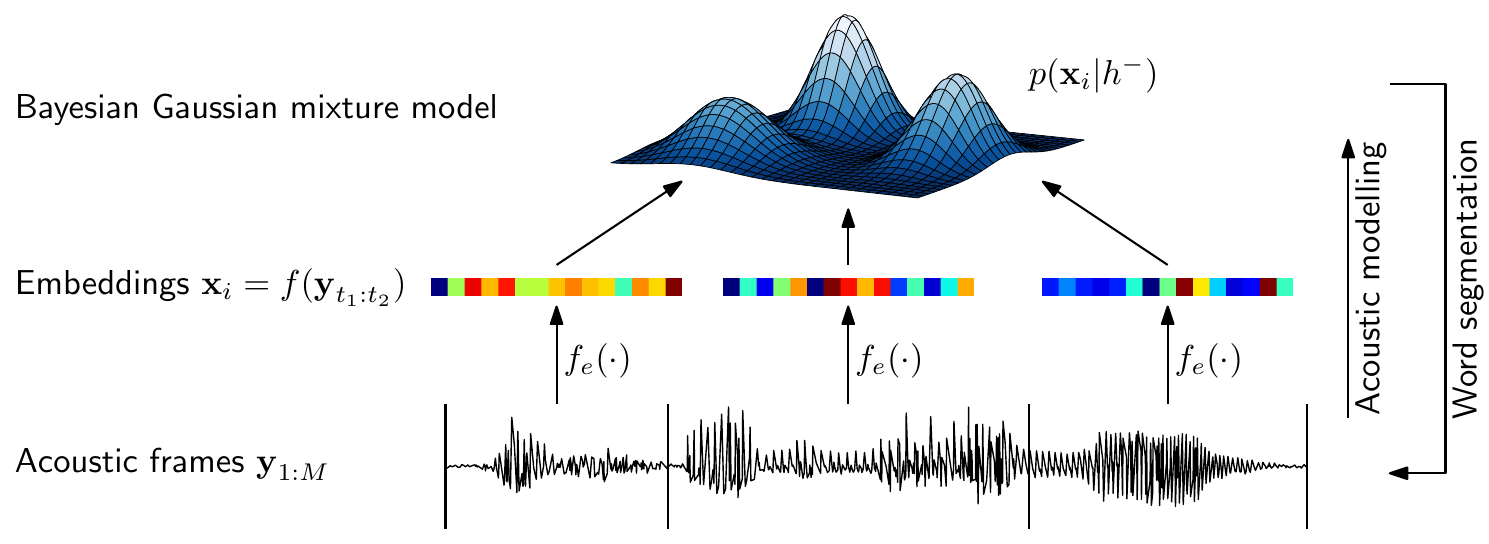}
    \caption{Overview of the segmental Bayesian model for unsupervised segmentation and clustering of speech.}
    \label{fig:unsup_wordseg}
\end{figure}

Assume for the moment such an ideal GMM exists.
This Bayesian GMM is the core component in our overall approach, which is illustrated in Figure~\ref{fig:unsup_wordseg}.
Given a new unsegmented unlabelled utterance of acoustic feature frames $\vec{y}_{1:M} = \vec{y}_1, \vec{y}_2, \ldots, \vec{y}_M$, the aim is to hypothesize where words start and end in the stream of features, and to which word type (GMM mixture component) every word segment belongs.
Given a proposed segmentation hypothesis (Figure~\ref{fig:unsup_wordseg} bottom), we can calculate the acoustic embedding vector for every proposed word segment (Figure~\ref{fig:unsup_wordseg} middle), calculate a likelihood score for each embedding under the current GMM (Figure~\ref{fig:unsup_wordseg} top), and obtain an overall
score for the current segmentation hypothesis.
The aim then is to find the optimal segmentation under the current GMM, which can be done using dynamic programming.
In our model, we sample a likely segmentation with a dynamic programming Gibbs sampling algorithm using
the probabilities we obtain from the Bayesian GMM. The result is a complete segmentation of the input
utterance and a prediction of the component to which every word segment belongs. 

In our actual model, the Bayesian GMM is built up jointly while performing segmentation: the GMM provides the likelihood terms required for segmentation, while the segmentation hypothesizes the boundaries for the word segments which are then clustered using the GMM.
The GMM (details in Section~\ref{sec:tidigits_gmm}) can thus be seen as an acoustic model which discovers the underlying word types of a language, while the segmentation component (Section~\ref{sec:tidigits_word_segmentation}) discovers where words start and end.
Below we provide complete details of the model.

\subsection{Fixed-dimensional representation of speech segments}
\label{sec:tidigits_embeddings}

The proposed model requires that any acoustic speech segment in an utterance be embedded in a fixed-dimensional space.
In principle, any approach that is able to map an arbitrary-length vector time series to a fixed-dimensional vector can be used.
We follow the embedding approach developed by~\cite{levin+etal_asru13}, which was outlined in Section~\ref{sec:background_segmental} and describe in more detail below.

The notation $Y = \vec{y}_{1:T}$ is used to denote a vector time series, where each $\vec{y}_t$ is the frame-level acoustic features (e.g.\ MFCCs).
We need a mapping function $f_e(Y)$ that maps time series $Y$ into a space $\mathbb{R}^D$ in which proximity between mappings indicates similar linguistic content, so 
embeddings of word tokens of the same type will be close together.
In~\citep{levin+etal_asru13}, the mapping $f_e$ is performed as follows.
For a target speech segment, a reference vector is constructed by calculating the DTW alignment cost to every exemplar in a reference set $\mathcal{Y}_{\text{ref}} = \{ Y_i \}_{i = 1}^{N_\text{ref}}$.
Applying dimensionality reduction to the reference vector yields the embedding in $\mathbb{R}^D$.
Dimensionality reduction is performed using Laplacian eigenmaps~\citep{belkin+niyogi_neurocomp03}.

Intuitively, Laplacian eigenmaps tries to find an optimal non-linear mapping such that the $k$-nearest neighbouring speech segments in the reference set $\mathcal{Y}_{\text{ref}}$ are mapped to similar regions in the target space $\mathbb{R}^D$.
To embed an arbitrary segment $Y$ which is not an element of $\mathcal{Y}_{\text{ref}}$, a kernel-based out-of-sample extension is used~\citep{belkin+etal_jmlr06}.
This performs a type of interpolation using the exemplars in $\mathcal{Y}_{\text{ref}}$ that are similar to target segment~$Y$.

In all experiments we use a radial basis function kernel:
\begin{equation}
    K(Y_i, Y_j) = \exp \left\{ - \frac{ \left[\text{DTW} (Y_i, Y_j) \right]^2 }{2 \sigma_K^2} \right\}
\end{equation}
where $\text{DTW} (Y_i, Y_j)$ denotes the DTW alignment cost between segments $Y_i$ and $Y_j$, and $\sigma_K$ is the kernel width parameter.
In~\citep{belkin+etal_jmlr06}, it was shown that the optimal projection to the $j^{\text{th}}$ dimension in the target space is given by
\begin{equation}
    h_j(Y) = \sum_{i = 1}^{N_\text{ref}} \alpha_i^{(j)} K(Y_i, Y)
    \label{eq:embed1}
\end{equation}
The $\alpha_i^{(j)}$ terms are the solutions to the generalized eigenvector problem $(\vec{L}\vec{K} + \xi \vec{I}) \vec{\alpha} = \lambda \vec{K} \vec{\alpha}$, with $\vec{L}$ the normalized graph Laplacian, $\vec{K}$ the Gram matrix with elements $K_{ij} = K(Y_i, Y_j)$ for $Y_i, Y_j \in \mathcal{Y}_{\text{ref}}$, and $\xi$ a regularization parameter. An arbitrary speech segment $Y$ is then mapped to the embedding $\vec{x} \in \mathbb{R}^D$ given by $\vec{x} = f_e(Y) = \left[ h_1(Y), h_2(Y), \ldots, h_d(Y) \right]\T$.

We have given only a brief outline of the embedding method here; complete details can be found in~\citep{belkin+niyogi_neurocomp03,belkin+etal_jmlr06,levin+etal_asru13}.

\subsection{Acoustic modelling: Discovering word types}
\label{sec:tidigits_gmm}

Given a segmentation hypothesis of a corpus (indicating where words start and end), the acoustic model needs to cluster the hypothesized word segments (represented as fixed-dimensional vectors) into groups of hypothesized word types.
Note again that acoustic modelling is performed jointly with word segmentation (next section), 
but here we describe the acoustic model under the current segmentation hypothesis.
Formally, given the embedded word vectors $\mathcal{X} = \{ \vec{x}_i\}_{i = 1}^{N}$ from the current segmentation hypothesis, the acoustic model needs to assign each vector $\vec{x}_i$ to one of $K$ clusters. 

We choose for the acoustic model a 
Bayesian GMM with fixed spherical covariance.
This model treats its mixture weights and component means as random variables rather than point estimates as is done in a regular GMM.
In~\citep{kamper+etal_slt14} we showed that the Bayesian GMM performs significantly better in clustering word embeddings than a regular GMM trained with expectation-maximization.
The former also fits naturally within the sampling framework of our complete model.

\begin{figure}[tbp]
    \centering
    \includegraphics[scale=0.95]{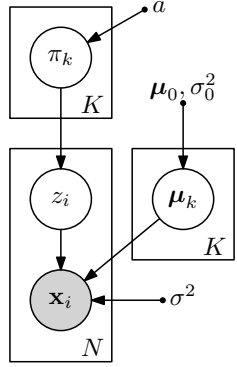}
    \caption{The graphical model of the Bayesian Gaussian mixture model with fixed spherical covariance used as acoustic model.}
    \label{fig:unsup_wordseg_fbgmm}
\end{figure}

The Bayesian GMM is illustrated in Figure~\ref{fig:unsup_wordseg_fbgmm}.
For each observed embedding $\vec{x}_i$, latent variable $z_i$ indicates the component to which $\vec{x}_i$ belongs.
The prior probability that $\vec{x}_i$ belongs to component $k$ is $\pi_k = P(z_i = k)$.
Given $z_i = k$, $\vec{x}_i$ is generated by the $k^\text{th}$ Gaussian mixture component with mean vector $\vec{\mu}_k$.
All components share the same fixed covariance matrix $\sigma^2 \vec{I}$; preliminary experiments, based on~\citep{kamper+etal_slt14}, indicated that it is sufficient to only model component means while keeping covariances fixed.
Formally, the model is then defined~as:

\noindent
\begin{minipage}{.45\linewidth}
    \centering
    \begin{alignat}{2}
        &\vec{\pi}  &&\sim \textrm{Dir}\left( a/{K} \vec{1} \right) \label{eq:fbgmm1} \\
        &z_i &&\sim \vec{\pi}  \label{eq:fbgmm2}
    \end{alignat}
    ~
\end{minipage}
\hfill
\begin{minipage}{.45\linewidth}
    \centering
    \begin{alignat}{2}
        &\vec{\mu}_k  &&\sim \mathcal{N} (\vec{\mu}_0, \sigma_0^2 \vec{I})  \label{eq:fbgmm3} \\
        &\vec{x}_i &&\sim \mathcal{N} (\vec{\mu}_{z_i}, \sigma^2 \vec{I})  \label{eq:fbgmm4}
    \end{alignat}
    ~
\end{minipage}

\noindent We use a symmetric Dirichlet prior in~\eqref{eq:fbgmm1} since it is conjugate to the categorical distribution in~\eqref{eq:fbgmm2}~\citep[p.~171]{barber}, and a spherical-covariance Gaussian prior in~\eqref{eq:fbgmm3} since it is conjugate to the Gaussian distribution in~\eqref{eq:fbgmm4}~\citep{murphy_bayesgauss07}.
We use $\vec{\beta} = (\vec{\mu}_0, \sigma_0^2, \sigma^2)$ to denote all the hyperparameters of the mixture components.

Given $\mathcal{X}$, we infer the component assignments $\vec{z} =
(z_1, z_2, \ldots, z_N)$ using a collapsed Gibbs sampler~\citep{resnik+hardisty_gibbs_tutorial10}.
Since we chose conjugate priors, we can marginalize over $\vec{\pi}$ and $\left\{ \vec{\mu}_k \right\}_{k = 1}^K$ and only need to sample $\vec{z}$.
This is done in turn for each $z_i$ conditioned on all the other current component assignments:
\begin{align}
    &P(z_i = k|\vec{z}_{\backslash i}, \mathcal{X} ; a, \vec{\beta} ) 
    \propto P(z_i = k|\vec{z}_{\backslash i}; a)  p(\vec{x}_i|\mathcal{X}_{\backslash i}, z_i = k, \vec{z}_{\backslash i}; \vec{\beta})
    \label{eq:collapsed1}
\end{align}
where $\vec{z}_{\backslash i}$ is all latent component assignments excluding $z_i$ and $\mathcal{X}_{\backslash i}$ is all embedding vectors apart from $\vec{x}_i$.

By marginalizing over $\vec{\pi}$, the first term on the right hand side of \eqref{eq:collapsed1} can be calculated as:
\begin{equation}
    P(z_i = k|\vec{z}_{\backslash i}; {a}) = \frac{N_{k\backslash i} + a/K}{N + a - 1}
    \label{eq:first_term7}
\end{equation}
where $N_{k \backslash i}$ is the number of embedding vectors from mixture component $k$ without taking $\vec{x}_i$ into account~\citep[p.~843]{murphy}.
This term can be interpreted as a discounted unigram language modelling probability.
Similarly, it can be shown that by marginalizing over $\vec{\mu}_k$, the second term 
\begin{equation}
    p(\vec{x}_i|\mathcal{X}_{\backslash i}, z_i = k, \vec{z}_{\backslash i}; \vec{\beta}) = p(\vec{x}_i | \mathcal{X}_{k \backslash i}; \vec{\beta})
    \label{eq:second_term1}
\end{equation}
is the posterior predictive of $\vec{x}_i$ for a Gaussian distribution with known spherical covariance and a conjugate prior over its means, which is itself a spherical covariance Gaussian distribution~\citep{murphy_bayesgauss07}.
Here, $\mathcal{X}_{k \backslash i}$ is the set of embedding vectors assigned to component $k$ without taking $\vec{x}_i$ into account.
Since the multivariate distributions in~\eqref{eq:fbgmm3} and~\eqref{eq:fbgmm4} have known spherical covariances, the probability density function (PDF) of the multivariate posterior predictive simply decomposes into the product of univariate PDFs; for a single dimension $x_i$ of vector $\vec{x}_i$, this PDF is given by
\begin{equation}
    p(x_i|\mathcal{X}_{k \backslash i})
    = \mathcal{N} (x_i|\mu_{N_{k \backslash i}}, \sigma_{N_{k \backslash i}}^2 + \sigma^2) \label{eq:univariate_post_predict}
\end{equation}
where
\begin{equation}
    \sigma_{N_{k \backslash i}}^2 = \frac{\sigma^2\sigma_0^2}{N_{k \backslash i}\sigma_0^2 + \sigma^2} \text{\ \ ,\ \ }
    \mu_{N_{k \backslash i}} = \sigma_{N_{k \backslash i}}^2 \left( \frac{\mu_0}{\sigma_0^2} + \frac{N_{k \backslash i}\overline{x}_{k \backslash i}}{\sigma^2} \right)
\end{equation}
and $\overline{x}_{k \backslash i}$ is component $k$'s sample mean for this dimension~\citep{murphy_bayesgauss07}.

Although we use a model with a fixed number of components $K$,
Bayesian models that marginalize over their parameters have been shown
to prefer sparser solutions than maximum-likelihood models with the
same structure~\citep{goldwater+griffiths_acl07}.  Thus, our Bayesian GMM tends towards
solutions where most of the data are clustered into just a few
components, and we can find good minimally constrained solutions by
setting $K$ to be much larger than the expected true number of
types and letting the model decide how many of those components
to use.

\subsection{Joint segmentation and clustering}
\label{sec:tidigits_word_segmentation}

The acoustic model of the previous section can be used to cluster
existing segments. Our joint segmentation and clustering system works
by first sampling a segmentation of the current utterance based on the
current acoustic model (marginalizing over cluster assignments for
each potential segment), and then resampling the clusters of the newly
created segments. The inference algorithm is a blocked Gibbs sampler
using dynamic programming, based on the work of \cite{mochihashi+etal_acl09}.

\begin{algorithm}[tbp]
\begin{algorithmic}[1]
\small
\State Choose an initial segmentation (e.g.\ random).
\For{$j = 1$ to $J$}\Comment{Gibbs sampling iterations}

    \For{$i = $ randperm$(1$ to $S)$} \Comment{Select utterance $\vec{s}_i$}
    
        \State Remove embeddings $\mathcal{X}(\vec{s}_i)$ from acoustic model.\label{alg_line:remove_embeds}
        
        \State Calculate $\alpha$'s using~\eqref{eq:forward}.
        \label{alg_line:forward}
        
        \State Draw $\mathcal{X}(\vec{s}_i)$ by sampling word boundaries using~\eqref{eq:backward}. \label{alg_line:sample_bounds} 

        \For{embedding $\vec{x}_i$ in newly sampled $\mathcal{X}(\vec{s}_i)$} \label{alg_line:component_assignment_start}
        
            \State Sample $z_i$ for embedding $\vec{x}_i$ using~\eqref{eq:collapsed1}.
            \label{alg_line:fbgmm_inside_loop}
        
        \EndFor \label{alg_line:component_assignment_end}

    \EndFor
\EndFor
\end{algorithmic}
\caption{Gibbs sampler for word segmentation and clustering of speech.}\label{alg:gibbs_wordseg}
\end{algorithm}

More formally, given acoustic data $\{ \vec{s}_i \}_{i = 1}^S$, where every utterance $\vec{s}_i$ consists of acoustic frames $\vec{y}_{1:M_i}$, we need to hypothesize word boundary locations and a word type (mixture component) for each hypothesized segment.
$\mathcal{X}(\vec{s}_i)$ denotes the embedding vectors under the current segmentation for utterance $\vec{s}_i$.
Pseudo-code for the blocked Gibbs sampler, which samples a segmentation utterance-wide, is given in Algorithm~\ref{alg:gibbs_wordseg}.
An utterance $\vec{s}_i$ is randomly selected; the embeddings from the current segmentation $\mathcal{X}(\vec{s}_i)$ are removed from the Bayesian GMM; a new segmentation is sampled; and finally the embeddings from this new segmentation are added back into the~Bayesian~GMM.

For each utterance $\vec{s}_i$ a new set of embeddings $\mathcal{X}(\vec{s}_i)$ is sampled in line~\ref{alg_line:sample_bounds} of Algorithm~\ref{alg:gibbs_wordseg}.  This is done using the forward filtering backward sampling dynamic programming algorithm~\citep{scott_jasa02}.
Forward variable $\alpha[t]$ is defined as the density of the frame sequence $\vec{y}_{1:t}$, with the last frame the end of a word: $\alpha[t] \defeq p(\vec{y}_{1:t} | {h^-})$.
{The embeddings and component assignments for all words not in $\vec{s}_i$, and the hyperparameters of the GMM, are denoted together as $h^- = (\mathcal{X}_{\backslash s}, \vec{z}_{\backslash s}; a, \vec{\beta})$.}
To derive recursive equations for $\alpha[t]$, we use a variable $q_t$ to indicate the number of
{acoustic observation frames in the hypothesized word that ends at}
frame $t$: if $q_t = j$, then $\vec{y}_{t - j + 1:t}$ is a word.
The forward variables can then be recursively calculated as:
\begin{align}
    \alpha[t]
    &= p(\vec{y}_{1:t} | h^-) \nonumber \\
    &= \sum_{j = 1}^t p(\vec{y}_{1:t}, q_t = j | h^-) \nonumber \\
    &= \sum_{j = 1}^t p(\vec{y}_{1:{t-j}}, \vec{y}_{{t - j + 1}:t}, q_t = j | h^-) \nonumber \\
    &= \sum_{j = 1}^t p(\vec{y}_{{t - j + 1}:t} | \vec{y}_{1:{t-j}}, q_t = j, h^-) p(\vec{y}_{1:{t-j}}, q_t = j | h^-) \nonumber \\
    &= \sum_{j = 1}^t p(\vec{y}_{{t - j + 1}:t} | h^-) p(\vec{y}_{1:{t-j}}, q_t = j | h^-) \nonumber \\
    &= \sum_{j = 1}^t p(\vec{y}_{{t - j + 1}:t} | h^-) \alpha[t - j] \label{eq:forward}
\end{align}
{starting with $\alpha[0] = 1$ and calculating~\eqref{eq:forward}
for $1 \leq t \leq M - 1$.}

The $p(\vec{y}_{{t - j + 1}:t} | h^-)$ term in~\eqref{eq:forward} is the value of a joint PDF over acoustic frames $\vec{y}_{{t - j + 1}:t}$.
In a frame-based supervised setting, this term would typically be calculated as the product of the PDF values of a GMM (or prior-scaled posteriors of a deep neural network) for the frames involved.
However, we work at a whole-word segment level, and our acoustic model is defined over a whole segment, which means we need to define this term explicitly. 
Let $\vec{x}' = f_e(\vec{y}_{{t - j + 1}:t})$ be the word embedding calculated on the acoustic frames $\vec{y}_{{t - j + 1}:t}$ (the hypothesized word).
We then treat the term as:
\begin{equation}
    p(\vec{y}_{{t - j + 1}:t} | h^-) \defeq \left[p \left(\vec{x}' | h^- \right) \right]^j \label{eq:prob_segment}
\end{equation}
Thus, as in the frame-based supervised case, each frame is assigned a PDF score.
But instead of having a different PDF value for each frame, all $j$ frames in the segment $\vec{y}_{{t - j + 1}:t}$ are assigned the PDF value of the whole segment under the current acoustic model.
In initial experiments we found that without this factor, severe over-segmentation occurred.
The marginal term in~\eqref{eq:prob_segment} can be calculated as:
\begin{align}
    p(\vec{x}' | h^-)
    &= \sum_{k = 1}^{K} p(\vec{x}', z_h = k | \mathcal{X}_{\backslash i}, \vec{z}_{\backslash i}; a, \vec{\beta}) \nonumber \\
    &= \sum_{k = 1}^{K} P(z_h = k | \vec{z}_{\backslash i}; a)  p(\vec{x}'| \mathcal{X}_{k\backslash i}; \vec{\beta})
    \label{eq:likelihood_fbgmm}
\end{align}
The two terms in~\eqref{eq:likelihood_fbgmm} are provided by the Bayesian GMM acoustic model, as given in equations~\eqref{eq:first_term7} and~\eqref{eq:second_term1}, respectively.

Once all forward variables have been calculated, a segmentation can be sampled backwards~\citep{mochihashi+etal_acl09}.
Starting from the final position $t = M$, we sample the preceding word boundary using
\begin{align}
    P(q_M  = j | \vec{y}_{1:M}, h^-)
    &\propto p(\vec{y}_{1:M}, q_M = j | h^-) \nonumber \\
    &= p(\vec{y}_{1:{M - j}}, \vec{y}_{{M - j + 1}:M}, q_M = j | h^-) \nonumber \\
    &= p(\vec{y}_{{M - j + 1}:M} | \vec{y}_{1:{M - j}}, q_M = j, h^-) p(\vec{y}_{1:{M - j}}, q_M = j | h^-) \nonumber \\
    &= p(\vec{y}_{{M - j + 1}:M} | h^-) p(\vec{y}_{1:{M - j}}, q_M = j | h^-) \nonumber \\
    &= p(\vec{y}_{{M - j + 1}:M} | h^-) \alpha[M - j]
\end{align}
In general, we can sample the preceding boundary from position $t$ using
\begin{equation}
    P(q_t = j | \vec{y}_{1:t}, h^-) \propto p(\vec{y}_{{t - j + 1}:t} | h^-) \alpha[t - j]
    \label{eq:backward}
\end{equation}
We calculate~\eqref{eq:backward} for $1 \leq j \leq t$ and sample while $t - j \geq 1$.

Algorithm~\ref{alg:gibbs_wordseg} gives the complete sampler for our model, showing how segmentation and clustering of speech is performed jointly.
The inner part of Algorithm~\ref{alg:gibbs_wordseg} is also illustrated in Figure~\ref{fig:unsup_wordseg}: lines~\ref{alg_line:remove_embeds} to~\ref{alg_line:sample_bounds} perform word segmentation which proceeds from top to bottom in Figure~\ref{fig:unsup_wordseg}, while lines~\ref{alg_line:component_assignment_start} to~\ref{alg_line:component_assignment_end} perform acoustic modelling which proceeds from bottom to top in the figure.

\newpage
\subsection{Iterating the model}
\label{sec:tidigits_iterating}

\begin{figure}[htbp]
    \centering
    \includegraphics[width=\linewidth]{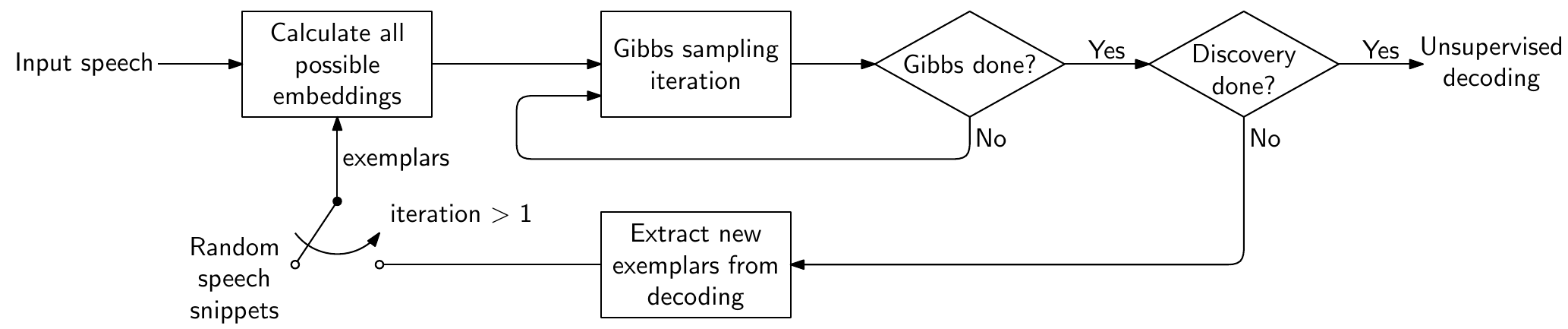}
    \caption{Flow diagram showing our complete approach. In the outer loop exemplars are iteratively refined, while Gibbs sampling is performed in the inner loop.}
    \label{fig:flow_diagram}
\end{figure}

As explained in Section~\ref{sec:tidigits_embeddings}, the fixed-dimensional embedding extraction relies on a reference set $\mathcal{Y}_{\text{ref}}$.
In~\citep{levin+etal_asru13}, this set was composed of true word segments.
In this unsupervised setting, we do not have such a set.
We therefore start with exemplars extracted randomly from the data.
Using this set, we extract embeddings and then run our sampler in an unconstrained setup where it is free to discover an order of magnitude more clusters than the true number of word types.
From the biggest clusters discovered in this first iteration (those that cover 90\% of the data), we extract a new exemplar set, which is used to recalculate embeddings.
We repeat this procedure for a number of iterations, resulting in a refined exemplar set~$\mathcal{Y}_{\text{ref}}$.
The complete process is illustrated in Figure~\ref{fig:flow_diagram}.

This iterative refinement procedure can be seen as a way in which discovered top-down knowledge are used to improve the representations.
At first, when the reference set consists of random noisy exemplars, the acoustic word embedding method can be seen as a purely bottom-up approach since it does not use any knowledge of longer-spanning patterns in the data and relies solely on the lowest-level acoustic features.
By then using terms automatically discovered using this bottom-up segmental representations, we start to incorporate automatically obtained top-down knowledge to improve the representations.
In the experiments presented next, we show that this top-down refinement of the acoustic word embedding method is essential in obtaining good full-coverage segmentation and clustering performance.

\section{Experiments}
\label{sec:tidigits_experiments}

\subsection{Evaluation setup}
\label{sec:tidigits_evaluation}

We evaluate using the TIDigits connected digit corpus~\citep{leonard_icassp84}, which has a vocabulary of eleven English digits: `oh' and `zero' through~`nine'.
Using this simple small-vocabulary task, we are able to thoroughly analyze 
the discovered units and report results on the same corpus as 
several previous unsupervised studies~\citep{tenbosch+cranen_interspeech07,sun+vanhamme_csl13,walter+etal_asru13,vanhainen+salvi_icassp14}.
In particular, we use the recent results of \cite{walter+etal_asru13} as baselines in our own experiments.

TIDigits consists of an official training set with 112~speakers (male and female) and 77 digit sequences per speaker, and a comparable test set.
Each set contains about 3~hours of speech.
Our model is unsupervised, which means that the concepts of training and test data become blurred.
We run our model on both sets separately---in each case, unsupervised modelling and evaluation is performed on the same set.
To avoid confusion with supervised regimes, we relabel the official TIDigits training set as `{TIDigits1}' and the test set as `{TIDigits2}'.
TIDigits1 was used during development for tuning hyperparameters (see Section~\ref{sec:tidigits_model_implementation}), while TIDigits2 was treated exclusively as unseen final test set.

For evaluation, the unsupervised decoded output of a system is compared to the ground truth transcriptions. From this comparison a mapping matrix $\vec{G}$ is constructed: 
$G_{ij}$ is the number of acoustic frames that are labelled as digit $i$ in the ground truth transcript and labelled as discovered word type $j$ by the model.
We then use the following three quantitative evaluation metrics.


\myemph{Average cluster purity:} Every discovered word type (cluster) is mapped to the most common ground truth digit in that cluster, given by $i' = \argmax_{i} G_{ij}$ for cluster $j$. Average purity is then defined as the total proportion of the correctly mapped frames: ${\sum_j \max_{i} G_{ij}} / {\sum_{i,j} G_{ij}}$.
If the number of discovered types is more than the true number of types, more than one cluster may be mapped to a single ground truth type, i.e.\ a many-to-one mapping, as in~\citep{sun+vanhamme_csl13}.

\myemph{Unsupervised WER:} Discovered types are again mapped, but here at most one cluster is mapped to a ground truth digit~\citep{walter+etal_asru13}. 
By then aligning the mapped decoded output from a system to the ground truth transcripts, we calculate $\textrm{WER} = \frac{S + D + I}{N}$, with $S$ the number of substitutions, $D$ deletions, $I$ insertions, and $N$ the tokens in the ground truth.
In cases where the number of discovered types is greater than the true number, {some clusters
will be left unassigned and counted as errors.}

\myemph{Word boundary $F$-score:} By comparing the
word boundary positions proposed by a system to those from
forced alignments of the data (falling within 40 ms), we 
calculate word boundary precision and recall, and report the $F$-scores.

We consider two system initialization strategies, which were also used in~\citep{walter+etal_asru13}: (i)~random initialization; and (ii)~initialization from a separate UTD system.
In the UTD condition, the boundary positions and cluster assignments for the words discovered by a UTD system can be used.
Walter et al.\ used both the boundaries and assignments, while we use only the boundaries for initialization 
(we did not find any gain by using the cluster identities as well).
We use the UTD system of~\cite{jansen+vandurme_asru11}.

As mentioned in Section~\ref{sec:tidigits_related_studies_full_coverage}, Walter et al.\ constrained their system to only discover eleven clusters (the true number).
For our model we consider two scenarios: (i) in the \textit{constrained} setting, we fix the number of components of the model to $K = 15$; (ii)~in the \textit{unconstrained} setting, we allow the model to discover up to $K = 100$ clusters.
For the first, we use $K = 15$ instead of $11$ since we found that more consistent performance on TIDigits1 is achieved when allowing some variation in cluster discovery.
In the second setting, $K = 100$ allows the model to discover many more clusters than the true number of types.
Since the Bayesian GMM is able to (and does) empty out some of its components (not all 100 clusters need to be used) this represents the case where we do not know vocabulary size upfront and the model itself is required to find a suitable number of clusters.

\subsection{Model implementation and hyperparameters}
\label{sec:tidigits_model_implementation}

The hyperparameters of our model are set mainly based on previous work on other tasks~\citep{kamper+etal_slt14}.
However, some parameters were changed by hand during development. 
These changes were made exclusively based on performance on {TIDigits1}.
Below, we also note the changes we made from our preliminary work on TIDigits, presented in the conference publication~\citep{kamper+etal_interspeech15}.
The hyperparameters used in~\citep{kamper+etal_interspeech15} led to far less consistent performance over multiple sampling runs: WER standard deviations were in the order of 9\% absolute, compared to the deviations of less than 1\% that we obtain here in Section~\ref{sec:tidigits_results}.

For the acoustic model (Section~\ref{sec:tidigits_gmm}), we use the following hyperparameters, based on~\citep{murphy_bayesgauss07,wood+black_jnm08,kamper+etal_slt14}: all-zero vector for $\vec{\mu}_0$, $a = 1$, $\sigma^2 = 0.005$, $\sigma_0^2 = \sigma^2/\kappa_0$ and $\kappa_0 = 0.05$.
Based on~\citep{levin+etal_asru13,kamper+etal_slt14} we use the following parameters for the fixed-dimensional embedding extraction (Section~\ref{sec:tidigits_embeddings}): dimensionality $D = 11$, $k = 30$, $\sigma_K = 0.04$, $\xi = 2.0$ and $N_\text{ref}=8000$.
The embedding dimensionality for this small-vocabulary task is less than that typically used for other larger-vocabulary unsupervised tasks, e.g.\ $D = 50$ in \citep{kamper+etal_slt14}.
In our preliminary work on TIDigits~\citep{kamper+etal_interspeech15}, we used $D = 15$ with $N_\text{ref}=5000$, but here we found that using $D = 11$ with a bigger reference set $N_\text{ref}=8000$ gave more consistent performance on TIDigits1.
For embedding extraction, speech is parameterized as 15-dimensional frequency-domain linear prediction features~\citep{athineos+ellis_asru03} at a frame rate of 10~ms, and cosine distance is used as similarity metric in DTW alignments.

As in~\citep{kamper+etal_slt14}, embeddings are normalized to the unit sphere.
We found that some embeddings were close to zero, causing issues in the sampler.
We therefore add low-variance zero-mean Gaussian noise before normalizing: the standard deviation of the noise is set to $0.05 \cdot \sigma_E$, where $\sigma_E$ is the sample standard deviation of all possible embeddings. 
Changing the $0.05$ factor within the range $[0.01, 0.1]$ made little difference.

Section~\ref{sec:tidigits_iterating} explained that to find the reference set $\mathcal{Y}_\textrm{ref}$ for embedding extraction, we start with exemplars extracted randomly from the data, and then iteratively refine the set by using the decoded output from our model.
In the first iteration we use $N_\text{ref}=8000$ random exemplars.
In subsequent iterations, we use terms from the biggest discovered clusters that cover at least 90\% of the data: 
we use the word tokens with the highest marginal densities as given by~\eqref{eq:likelihood_fbgmm} in each of these clusters to yield $4000$ discovered exemplars which we use in addition to $4000$ exemplars again extracted randomly from the data, to give a total set of size $N_\text{ref}=8000$.
We found that performance was more consistent when still using some random exemplars in $\mathcal{Y}_\textrm{ref}$ after the first iteration.

To make the search problem in Algorithm~\ref{alg:gibbs_wordseg} tractable, we require potential words to be between 200~ms and 1~s in duration, and we only consider possible word boundaries at 20~ms intervals.
By doing this, the number of possible embeddings is greatly reduced.
{Although embedding comparisons are fast, the calculation of the embeddings is not, and this is the main bottleneck of our approach.}
In our implementation, all allowed embeddings are pre-computed.
The sampler can then look up a particular embedding without the need to compute it on the~fly.
\edit{The calculation of all possible embeddings given a particular exemplar set (i.e.\ one iteration of the outer loop in Figure~\ref{fig:flow_diagram}) takes about one day when parallelized over $20$ CPUs, each with a speed of $2.8~\textrm{GHz}$.}

To improve sampler convergence, we use simulated annealing~\citep{goldwater+etal_cognition09}, by raising the boundary probability in~\eqref{eq:backward} to the power $\frac{1}{\xi}$ before sampling, where $\xi$ is a temperature parameter.
We also found that convergence is improved by first running the sampler in Algorithm~\ref{alg:gibbs_wordseg} without sampling boundaries.
In all experiments we do this for 25 iterations.
Subsequently, the complete sampler is run for $J = 25$ Gibbs sampling iterations with 5 annealing steps in which $\frac{1}{\xi}$ is increased linearly from $0.01$ to $1$.
In all cases we run 5 sampling chains in parallel~\citep{resnik+hardisty_gibbs_tutorial10}, and report average performance and standard deviations.
\edit{Each chain takes about $15$ hours on a single $2.8~\textrm{GHz}$ CPU.}

\subsection{Results and analysis}
\label{sec:tidigits_results}

\subsubsection{Unconstrained model evaluation}

As explained, we use our model to iteratively rediscover the embedding reference set $\mathcal{Y}_{\text{ref}}$.
Table~\ref{tbl:exemplars} shows the performance of the unconstrained segmental Bayesian model on TIDigits1 as the reference set is refined.
Random word boundary initialization is used throughout.
Unconstrained modelling represents the most realistic setting where vocabulary size is not known upfront.
Standard deviations are less than $0.3\%$ absolute for all~metrics.

\begin{table}[!tb]
    \mytable
    \caption{Performance of the unconstrained segmental Bayesian model on TIDigits1 over iterations in which the reference set is refined.}
    \begin{tabularx}{0.918\linewidth}{@{}lCCCCC@{}}
        \toprule
        Metric     & 1 & 2 & 3 & 4 & 5 \\
        \midrule
        WER (\%)                        & $35.4$ & $23.5$ & $21.5$ & $21.2$ & $22.9$ \\
        Average cluster purity (\%)       & $86.5$ & $89.7$ & $89.2$ & $88.5$ & $86.6$ \\
        Word boundary $F$-score (\%)         & $70.6$ & $72.2$ & $71.8$ & $70.9$ & $69.4$ \\
        Clusters covering 90\% of data   & 20             & 13 & 13 & 13 & 13 \\
        \bottomrule
    \end{tabularx}
    \label{tbl:exemplars}
\end{table}

Despite being allowed to discover many more clusters (up to 100) than the true number of word types (11), the model achieves a WER of 35.4\% in the first iteration, which improves to around 21\% in iterations 3 and 4.
This shows that to achieve good performance, it is essential to use top-down refinement for the acoustic word embeddings used here.
Error rate increases slightly in iteration 5.
Cluster purity over all iterations is above $86.5\%$, which is higher than the scores of around 85\% reported by \cite{sun+vanhamme_csl13}, as described in Section~\ref{sec:background_fullcoverage}. 
Word boundary $F$-scores are around 70\% over all iterations.
As mentioned, the Bayesian GMM is biased not to use all of its 100 components.
Despite this, none of the models empty out any of their components.
However, most of the data is covered by only a few components: the last row in Table~\ref{tbl:exemplars} shows that in the first iteration, 90\% of the data is covered by the 20 biggest mixture components, while this number drops to 13 clusters in subsequent iterations.

\begin{figure*}[!p]
    \centering
    \includegraphics[scale=0.825]{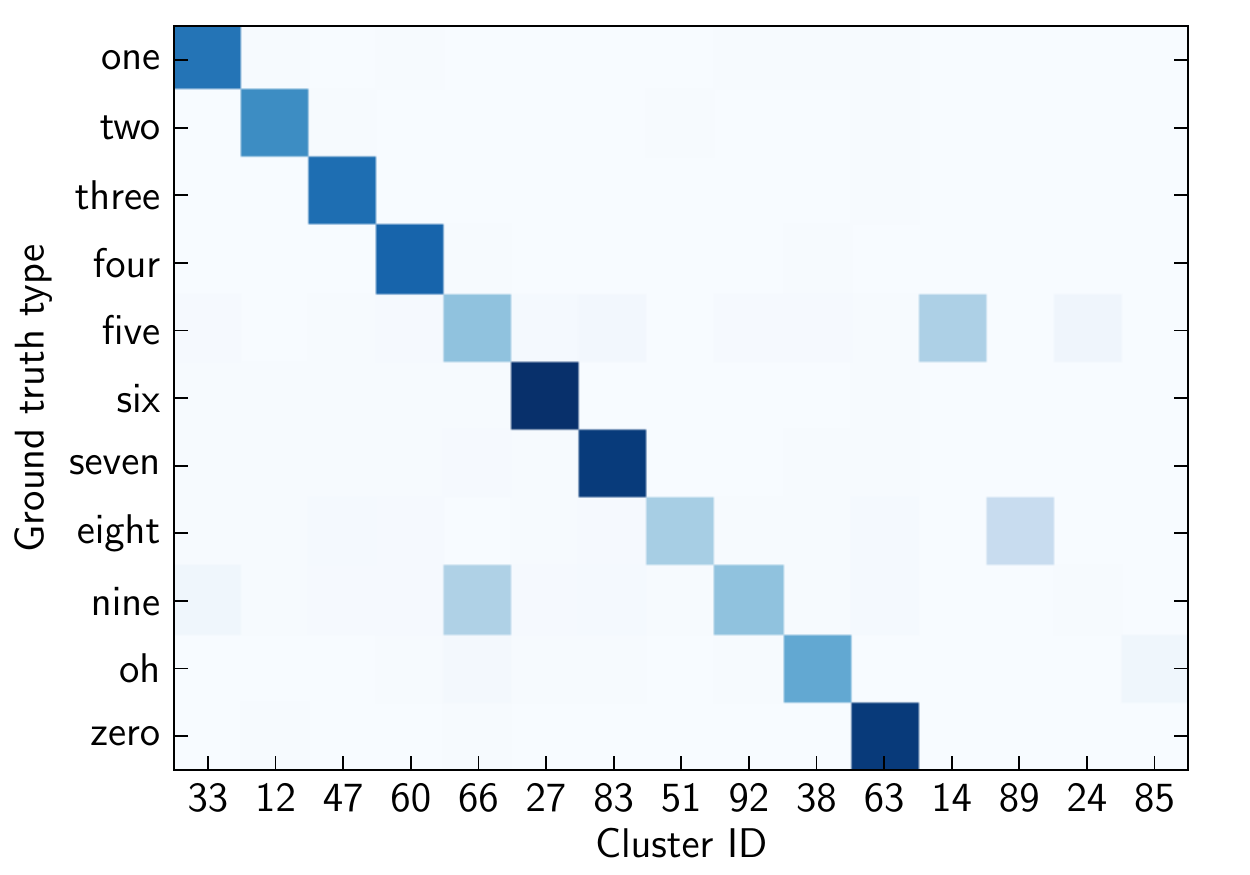}
    \caption{Mapping matrices between ground truth digits and discovered word types for the third iteration unconstrained model in Table~\ref{tbl:exemplars}.
    }
    \label{fig:unconstrained_mappings3}
\end{figure*}

\begin{figure*}[!p]
    \centering
    \includegraphics[scale=0.825]{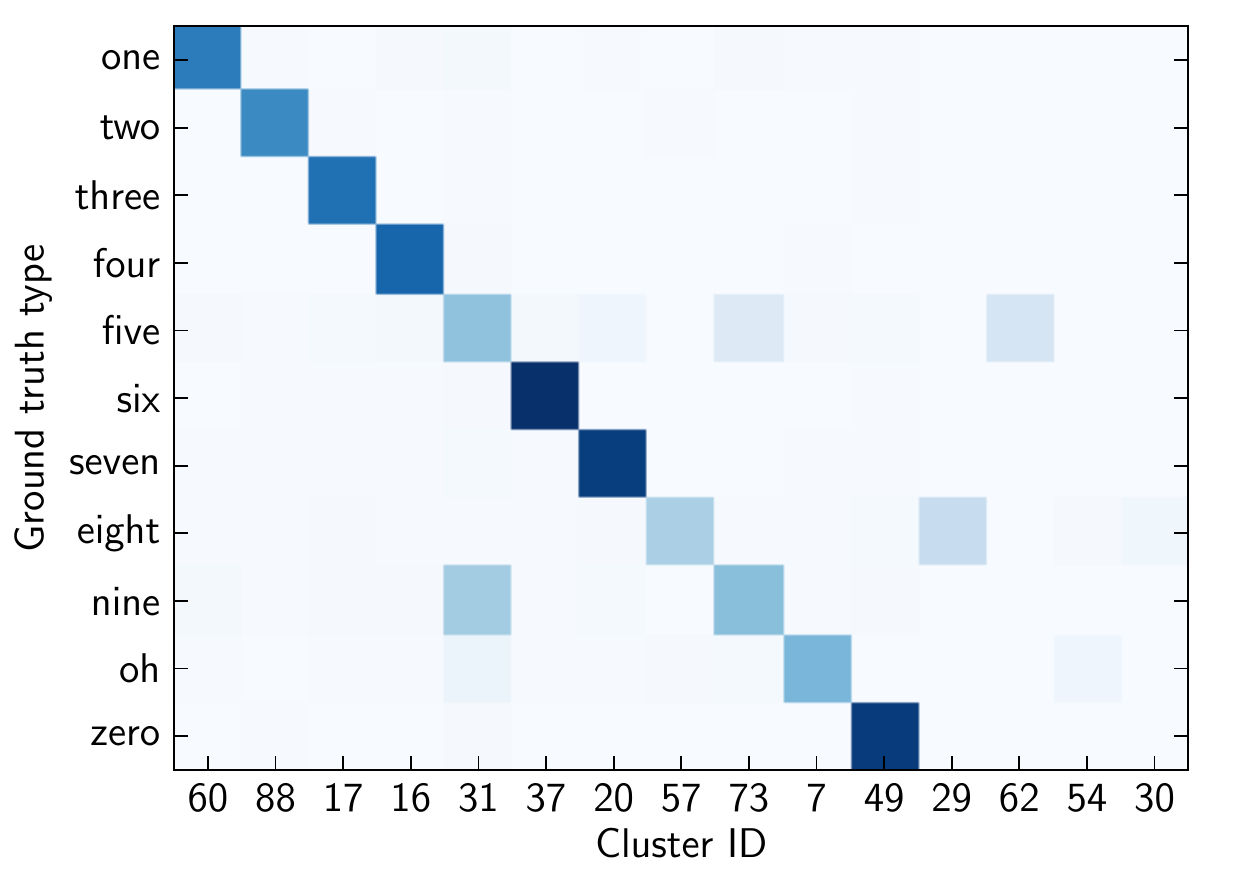}
    \caption{Mapping matrices between ground truth digits and discovered word types for the fifth iteration unconstrained model in Table~\ref{tbl:exemplars}.
    }
    \label{fig:unconstrained_mappings5}
\end{figure*}

In order to analyze the type of errors that are made, we visualize the mapping matrix $\vec{G}$, which gives the number of frames of overlap between the ground truth digits and the discovered word types (Section~\ref{sec:tidigits_evaluation}).
Figures~\ref{fig:unconstrained_mappings3} and~\ref{fig:unconstrained_mappings5} show the mappings for the 15 biggest clusters 
of the unconstrained models of iterations 3 and 5 of Table~\ref{tbl:exemplars}, respectively.

\begin{figure}[!tb]
    \centering
    \includegraphics[width=0.8\linewidth]{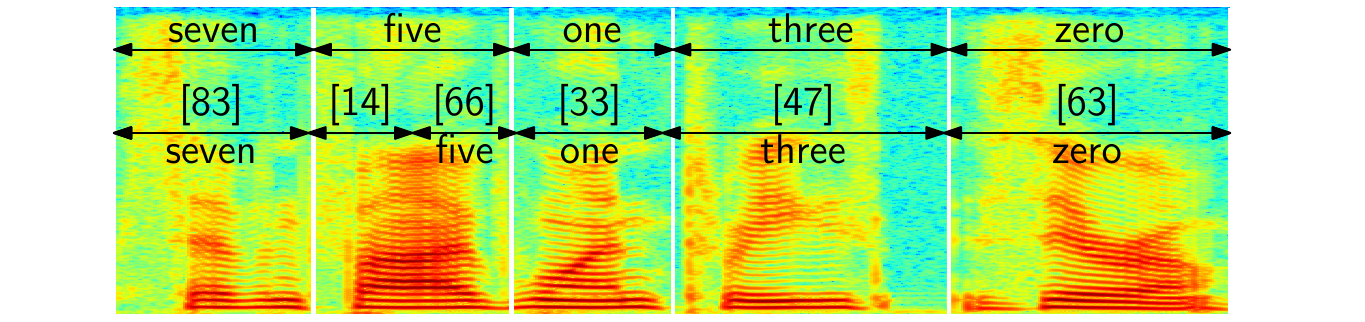}
    \caption{\edit{Example of ground truth alignments (top) with model output (cluster IDs in brackets with mappings below) for the utterance `seven five one three zero' using the iteration 3 unconstrained model in Table~\ref{tbl:exemplars}.}}
    \label{fig:tidigits_segmentation_eg}
\end{figure}

Consider the mapping in Figure~\ref{fig:unconstrained_mappings3} for iteration 3.
Qualitatively we observe a clear correspondence between the ground truth and discovered word types, which coincides with the high average purity of 89.2\%.
Apart from cluster 66, all other clusters overlap mainly with a single digit.
Listening to cluster 66 reveals that most tokens correspond to [ay v] from the end of the digit `five' and tokens of [ay n] from the end of `nine', both dominated by the diphthong. 
Correspondingly, most of the tokens in cluster 14 are the beginning [f ay] of `five', while cluster 92 is mainly the beginning [n~ay] of `nine'.
The digit `eight' is split across two clusters: cluster 51 mainly contains `eight' tokens where the final [t] is not pronounced, 
while in cluster 89 the final [t] is explicitly produced.
\edit{To qualitatively illustrate the accuracy of the predicted segmentation, Figure~\ref{fig:tidigits_segmentation_eg} compares ground truth alignments to the segmentation predicted by the model for a particular utterance; note how the digit `five' is split across two clusters, as observed above.}

Table~\ref{tbl:exemplars} shows that performance deteriorates slightly in iteration 5. 
{By comparing Figures~\ref{fig:unconstrained_mappings3} and~\ref{fig:unconstrained_mappings5}, the source of the extra errors can be observed:}
overall the mapping in the fifth iteration (Figure~\ref{fig:unconstrained_mappings5}) looks similar to that of the third (Figure~\ref{fig:unconstrained_mappings3}), except the digit `five' is now also partly covered  by a third cluster (73). 
This cluster mainly contains beginning portions of `five' and `nine', again dominated by the diphthong [ay].
Cluster 62 in this case mainly contains tokens of the fricative [f] from `five'.
Note that both WER and boundary $F$-score penalize the splitting of digits, although the discovered clusters correspond to consistent partial words.
Below, this issue is discussed~further.

\begin{figure}[!t]
    \centering
    \includegraphics[scale=0.85]{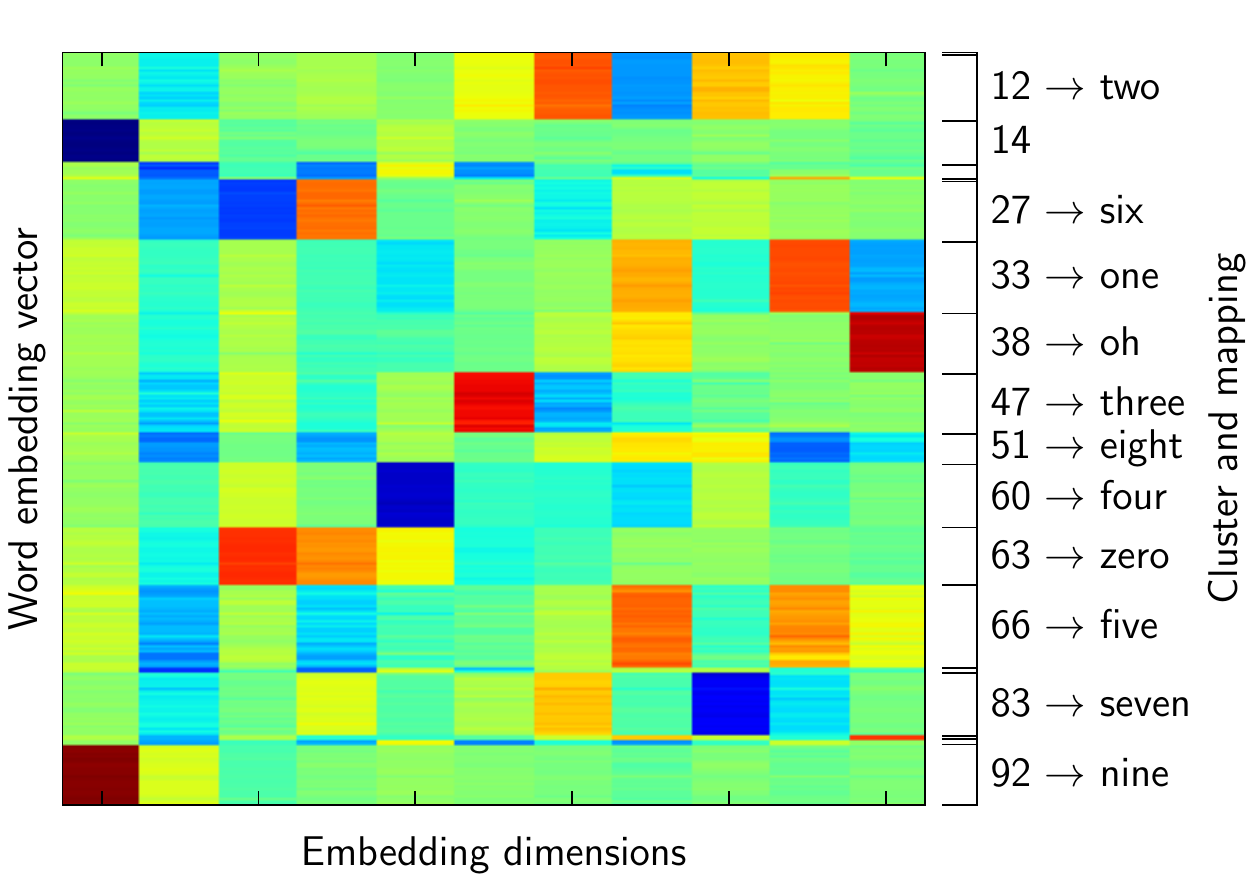}
    \caption{Embedding vectors for the discovered word types from a single speaker for the iteration 3 unconstrained model in Table~\ref{tbl:exemplars}. The greedy mapping from discovered to true word type (for calculating WER) is given on the right; since there are more clusters than digits, some clusters are left unmapped.}
    \label{fig:discovered_embeddings}
\end{figure}

One might suspect from the analysis in Figures~\ref{fig:unconstrained_mappings3} and~\ref{fig:unconstrained_mappings5} that some of the
discovered word types are bi-modal, i.e.\ that when a single component
of the Bayesian GMM contains two different true types  (e.g. cluster
66 in Figure~\ref{fig:unconstrained_mappings3}, with tokens of both `five' and `nine'), there might
be two relatively distinct sub-clusters of embeddings within that
component. However, this is not the case. Figure~\ref{fig:discovered_embeddings} shows the embeddings
of the discovered word types for a single speaker from the model in
iteration 3 of Table~\ref{tbl:exemplars}; embeddings are ordered and stacked by
discovered type along the $y$-axis, with the embedding values coloured
along the $x$-axis.  The embeddings for cluster 66 appear uni-modal,
despite containing both [ay v] and [ay n] tokens; yet they are
distinct from the embeddings in cluster 92 ([n ay] tokens) and cluster
14 ([f ay]). This analysis suggests that the model is finding sensible
clusters given the embedding representation it has, and to
consistently improve results we would need to focus on developing more
discriminative embeddings, an interesting area for future work (Chapter~\ref{chap:conclusion}).

\subsubsection{Constrained model evaluation and comparison}

To compare with the discrete HMM-based system of \cite{walter+etal_asru13}, we use the exemplar set discovered in  
iteration 3 of Table~\ref{tbl:exemplars} (using an unconstrained setup up to this point) and then constrain the Bayesian segmental model to 15 components.
Table~\ref{tbl:train_results} shows WERs achieved on TIDigits1. 
Under random initialization, the constrained segmental Bayesian model performs $12.7\%$ absolute better than the discrete HMM.
When using UTD for initialization, the discrete HMM does better by $1.3\%$ absolute.
The WER of the third-iteration unconstrained model in Table~\ref{tbl:exemplars} is repeated in the last row of Table~\ref{tbl:train_results}.
Despite only mapping eleven out of 100 clusters to true labels, this unconstrained model still yields 10.6\% absolute lower WER  than the randomly-initialized discrete HMM with the correct number of clusters.
By comparing rows two and three, we observe that there is only a $2.1\%$ absolute gain in WER by constraining the Bayesian model to a stricter number of types.

\begin{table}[!tbp]
    \mytable
    \caption{WER (\%) on TIDigits1 of the unsupervised discrete HMM system of \cite{walter+etal_asru13} and the segmental Bayesian model.}
    \begin{tabularx}{0.918\linewidth}{@{}lCcC@{}}
        \toprule
        Model & Constrained  & Random init. & UTD init. \\ 
        \midrule
        Discrete HMM~\citep{walter+etal_asru13} & yes & 32.1 & 18.1 \\ 
        Segmental Bayesian & yes & $19.4 \pm 0.3$ & $19.4 \pm 0.1$ \\ 
        Segmental Bayesian & no & $21.5 \pm 0.1$ & -  \\ 
        \bottomrule
    \end{tabularx}
    \label{tbl:train_results}
\end{table}

\subsubsection{Generalization and hyperparameters}

As noted in Section~\ref{sec:tidigits_model_implementation}, some development decisions were made based on performance on TIDigits1.
TIDigits2 was kept as unseen data up to this point.
Using the setup developed on TIDigits1, we repeated exemplar extraction and segmentation separately on TIDigits2.
Three iterations of exemplar refinement were used.
Table~\ref{tbl:train_test_results} shows the performance of randomly-initialized systems on both TIDigits1 and TIDigits2, with the performance on TIDigits1 repeated from Table~\ref{tbl:train_results}.

\begin{table*}[!bp]
    \mytable
    \caption{Performance of the Bayesian segmental model on TIDigits1 and TIDigits2, using random initialization.
    Results from both the constrained ($K = 15$) and unconstrained ($K = 100$) segmental Bayesian models are shown.}
    \begin{tabularx}{\linewidth}{@{}lcCCcCC@{}}
        \toprule
         & \multicolumn{3}{c}{TIDigits1 (\%)} & \multicolumn{3}{c}{TIDigits2 (\%)} \\
        \cmidrule(l){2-4} \cmidrule(l){5-7}
        Model & WER & Cluster purity & Boundary $F$-score & WER & Cluster purity & Boundary $F$-score \\
        \midrule
        Constrained & $19.4 \pm 0.3$ & $88.4 \pm 0.06$ & $70.6 \pm 0.2$ & $13.2 \pm 1.0$ & $91.2 \pm 0.2$ & $76.7 \pm 0.7$ \\
        Unconstrained & $21.5 \pm 0.1$ & $89.2 \pm 0.1$ & $71.8 \pm 0.2$ & $17.6 \pm 0.2$ & $92.5 \pm 0.1$ & $77.6 \pm 0.3$\\
        \bottomrule
    \end{tabularx}
    \label{tbl:train_test_results}
\end{table*}

\begin{figure}[tbp]
    \centering
    \includegraphics[scale=0.825]{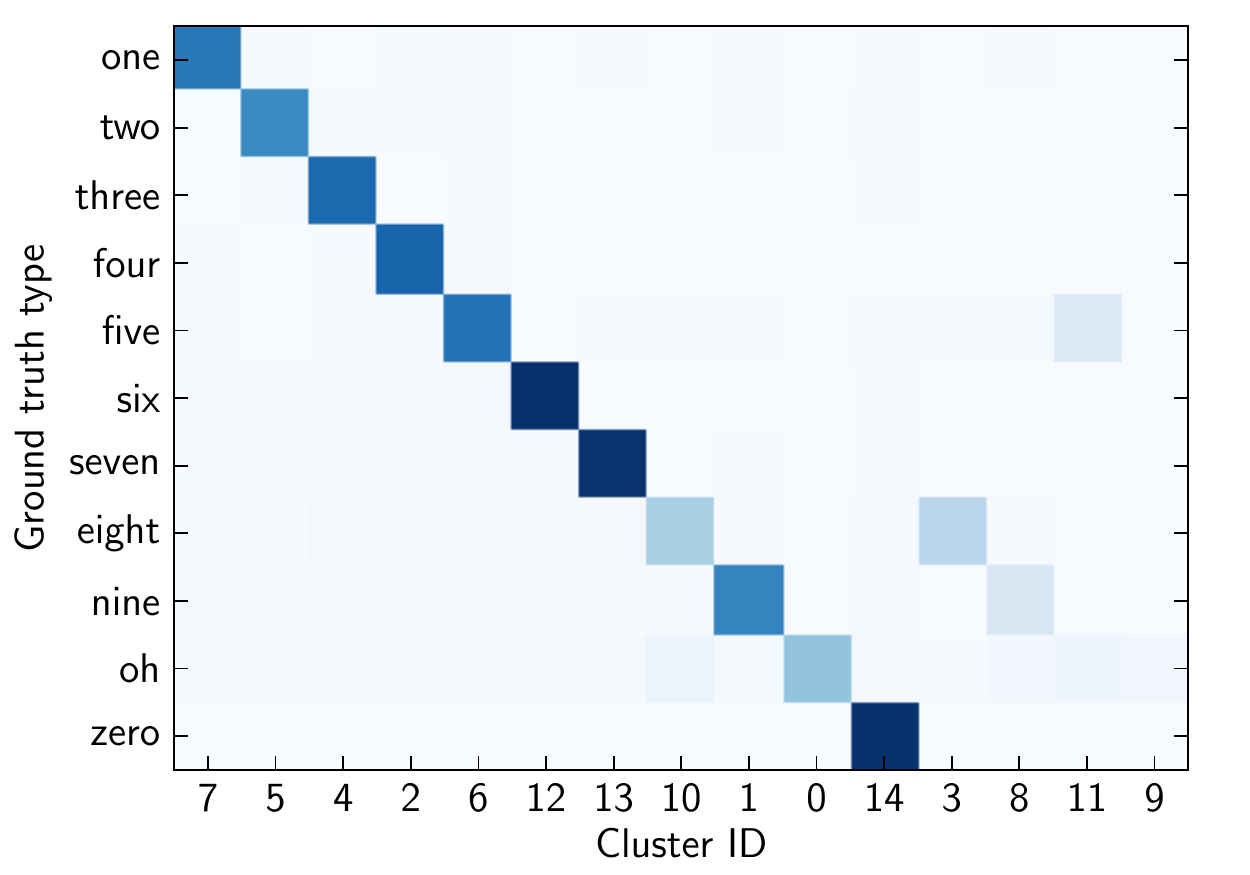}
    \caption{Mapping matrix between ground truth digits and discovered word types for the constrained segmental Bayesian model in Table~\ref{tbl:train_test_results} on TIDigits2.}
    \label{fig:mapping_test_v8_4_pass_3_am_K_15}
\end{figure}

Across all metrics, performance is better on TIDigits2 than on TIDigits1: WERs drop by 6.2\% and 3.9\% absolute for the constrained and unconstrained models, respectively; cluster purity improves by around 3\% absolute; and boundary $F$-score is higher by 6\% absolute.
To understand this discrepancy, consider the mapping matrix in Figure~\ref{fig:mapping_test_v8_4_pass_3_am_K_15} for the constrained segmental Bayesian model on TIDigits2 (13.2\% WER, Table~\ref{tbl:train_test_results}).
The figure shows that every cluster is dominated by data from a single ground truth digit. Furthermore, all digits apart from `eight' are found in a single cluster.
Now consider the mapping in Figure~\ref{fig:unconstrained_mappings3} for the unconstrained segmental Bayesian model on TIDigits1 (giving the higher WER of 21.5\%, Table~\ref{tbl:train_test_results}).
This mapping is similar to that of Figure~\ref{fig:mapping_test_v8_4_pass_3_am_K_15}, apart from two digits: both `five' and `nine' are split into two clusters, corresponding to beginning and end partial words.
Although these digits are consistently decoded as the same sequence of clusters, WER counts the extra clusters as insertion errors.
These small differences in the discovered word types results in a non-negligible difference in WER between TIDigits1 and TIDigits2.

This analysis and the previous discussion of Figures~\ref{fig:unconstrained_mappings3} and~\ref{fig:unconstrained_mappings5} indicate that 
unsupervised WER is a particularly harsh measure of unsupervised word segmentation performance: the model may discover consistent units, but if these units do not coincide with whole words, the system will be penalized.
This is also the case for word boundary $F$-score.
Average cluster purity is less affected since a many-to-one mapping is performed; Table~\ref{tbl:train_test_results} shows that purity changes the least of the three metrics when moving from TIDigits1 to TIDigits2.

\begin{figure}[tbp]
    \centering
    \includegraphics[scale=0.825]{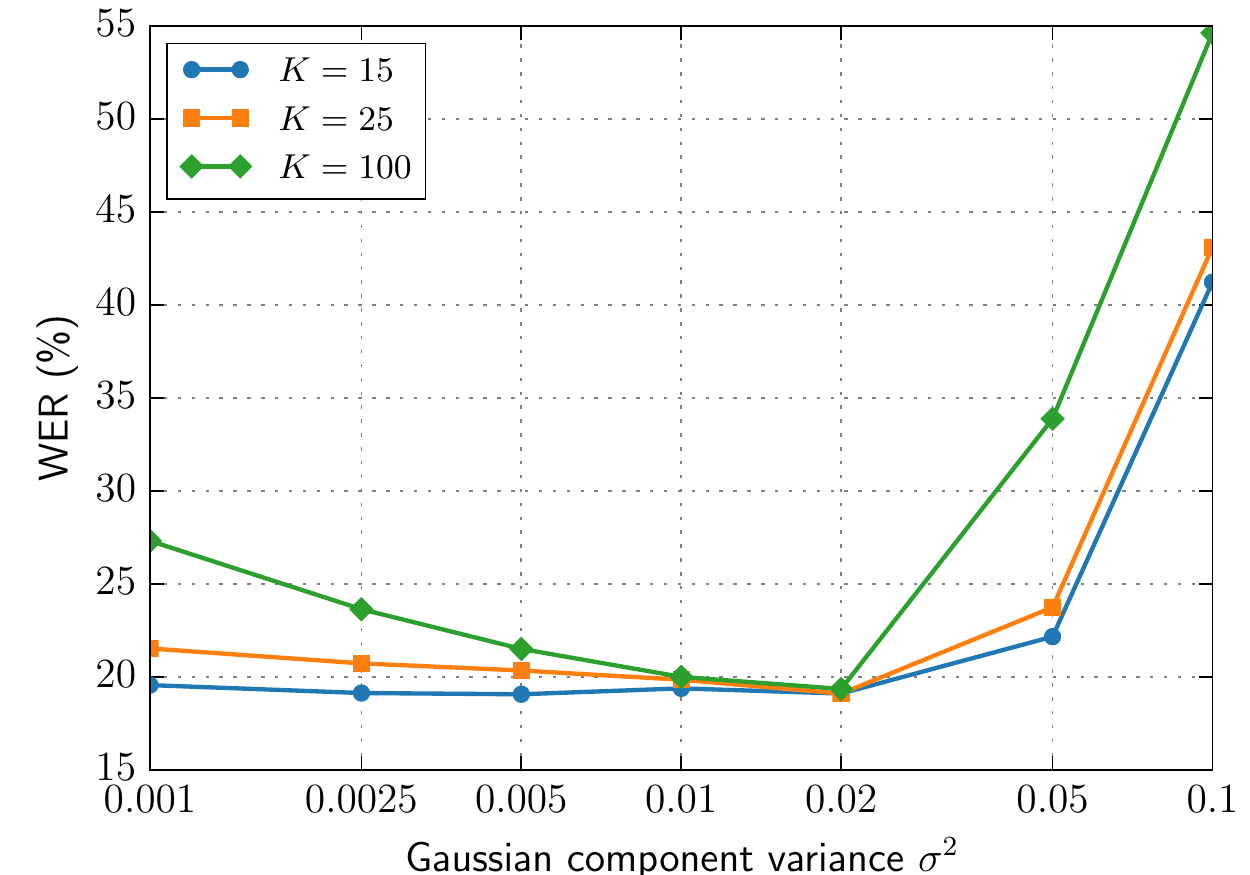}
    \caption{WERs of the segmental Bayesian model on TIDigits1 as the number of Gaussian components $K$ and variance $\sigma^2$ is varied (log-scale on $x$-axis).}
    \label{fig:train_optimize_wer}
\end{figure}

In a final set of experiments, we considered the effect of model hyperparameters.
We found that performance is most sensitive to changes in the maximum number of allowed Gaussian components $K$ and the component variance $\sigma^2$.
Figure~\ref{fig:train_optimize_wer} shows the effect on WER when changing these hyperparameters.
Results are reasonably stable for $\sigma^2$ in the range $[0.0025, 0.02]$, with WERs below 25\%.
When allowing many components ($K = 100$) and using a small variance, as on the left of the figure, fragmentation takes place with digits being separated into several clusters.
On the right side of the figure, where large variances are used, a few garbage clusters start to capture the majority of the data, leading to poor performance.
The figure also shows that lower WER could be achieved by using a $\sigma^2 = 0.02$ instead of $0.005$, which we used in the experiments above, based on~\citep{kamper+etal_slt14}.
The reason for the three curves meeting at this $\sigma^2$ setting is that, for all three settings of $K$, more than 90\% of the data are captured by only the eleven biggest clusters.

\begin{figure}[tbp]
    \centering
    \includegraphics[scale=0.825]{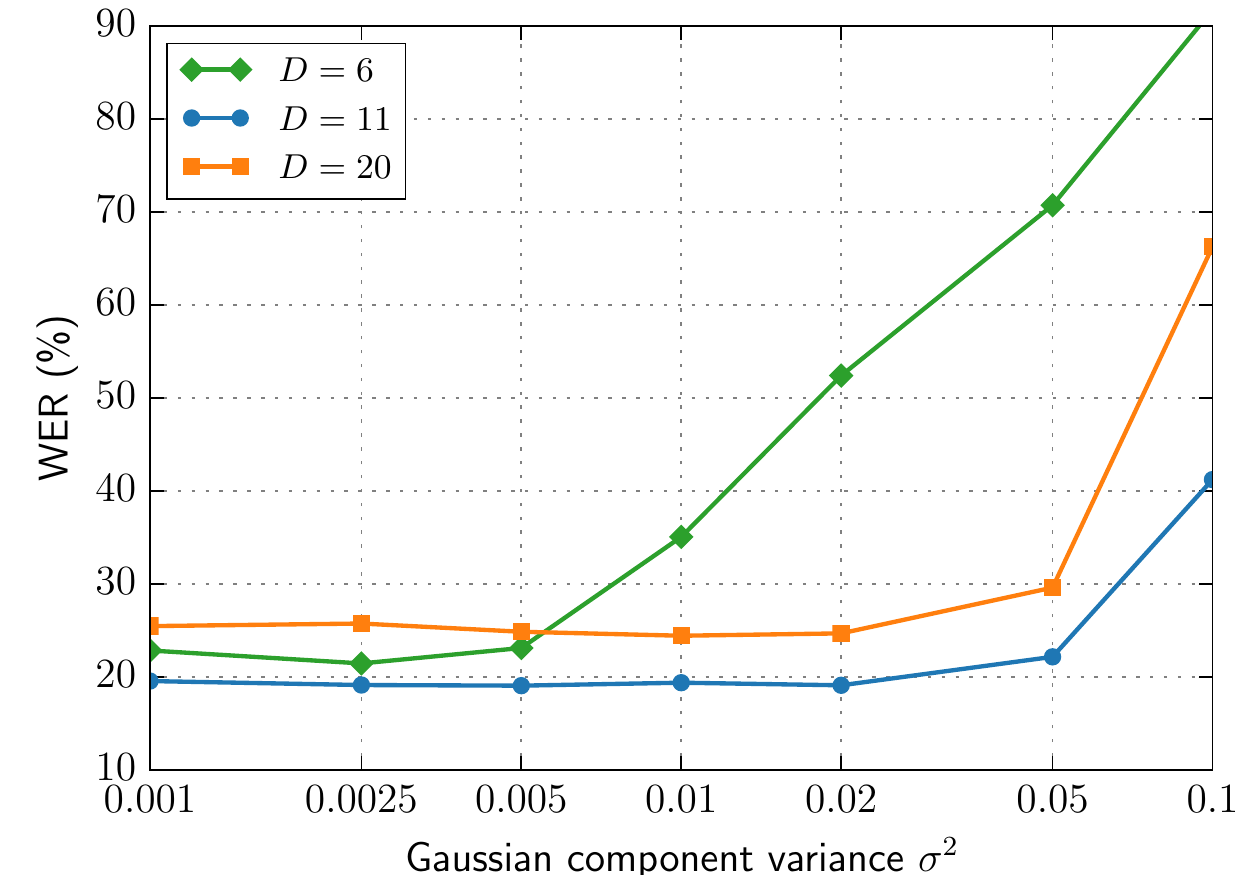}
    \caption{WERs of the segmental Bayesian model on TIDigits1 as the embedding dimensionality $D$ and variance $\sigma^2$ is varied (log-scale on $x$-axis; $K = 15$).}
    \label{fig:train_optimize_dims_wer}
\end{figure}

{We similarly varied the target embedding dimensionality $D$ using a constrained setup ($K = 15$), as shown in Figure~\ref{fig:train_optimize_dims_wer}. 
For $D = 6$, garbage clusters start to capture the majority of the tokens at lower settings of $\sigma^2$ than for $D = 11$ and $D = 20$.
Much more stable performance is achieved in the latter two cases.
The slightly worse performance of the $D = 20$ setting compared to the others is mainly due to a cluster containing the diphthong [ay], which is present in both `five' and `nine'.
}

\section{Challenges in scaling to larger vocabularies}
\label{sec:tidigits_scaling}

In this chapter we evaluated our system on a small-vocabulary multi-speaker dataset in order to compare to previous work and to allow us to thoroughly analyze the discovered structures.
However, the ultimate aim for zero-resource speech processing methods 
are to be useful on realistic multi-speaker corpora with larger vocabularies.
The system as presented here is not directly able to scale to such settings.
Below we discuss why this is the case, and propose possible solutions, some of which are explored in the next chapter.

\begin{figure}[tbp]
    \centering
    \includegraphics[scale=0.825]{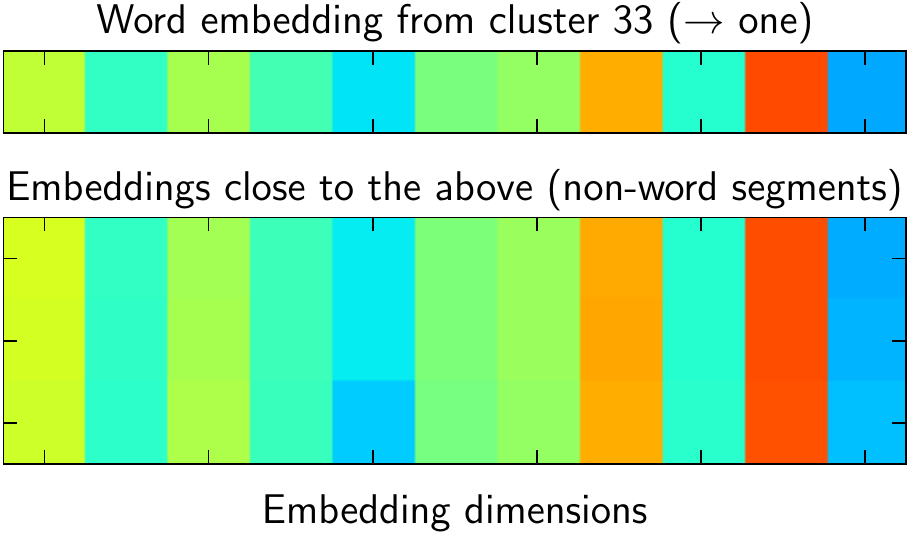}
    \caption{Four embeddings from the same speaker as in Figure~\ref{fig:discovered_embeddings}: the top one is from cluster 33, the bottom three are from short non-word speech segments.}
    \label{fig:nearest_neighbour_one}
\end{figure}

Although acoustic word embeddings have several benefits (Section~\ref{sec:tidigits_related_studies_full_coverage}) and are computationally much more efficient than performing exhaustive DTW comparisons between potential segments, 
the
fixed-dimensional embedding calculations are
still
the main bottleneck in our overall approach, since embeddings must be computed for each of the very large number of
potential word segments. 
The embeddings also limit accuracy; one case in particular where the embedding function produces poor embeddings is for  very short speech segments.
An example is given in Figure~\ref{fig:nearest_neighbour_one}.
The first embedding is from cluster 33 in Figure~\ref{fig:discovered_embeddings}, which is reliably mapped to the digit `one'.
The bottom three embeddings are from short segments not overlapping with the true digit `one', with
respective durations 20~ms, 40~ms and 80~ms.
Although these three speech segments have little similarity to the segments in cluster 33, Figure~\ref{fig:nearest_neighbour_one} shows that their embeddings are a good fit to this cluster.
This is possibly due to the aggressive warping in the DTW alignment of these short segments, together with artefacts from normalizing the embeddings to the unit sphere. 
This failure-mode is easily dealt with by setting a minimum duration constraint (Section~\ref{sec:tidigits_model_implementation}), but again shows our model's reliance on accurate embeddings.

To scale to larger corpora, both the efficiency (and ideally the accuracy) of the embeddings must therefore be improved. 
More importantly, the above discussion highlights a shortcoming of our approach: the sampler considers potential word boundaries at any position, 
without regard to the original acoustics or any notion of a minimal unit.
Many of the previous studies~\citep{walter+etal_asru13,lee_phd14,lee+etal_tacl15,rasanen+etal_interspeech15} use a first-pass method to find positions of high acoustic change and then only allow word boundaries at these positions.
This implicitly defines a minimal unit: the pseudo-phones or pseudo-syllables segmented in the first pass.
By using such a first-pass method in our system, 
the number of embedding calculations would greatly be reduced and it would provide a more principled way to deal with artefacts from short segments. 

Another challenge when dealing with larger vocabularies is the choice of the number of clusters $K$.
An upper-bound of $K = 100$, as we use for our unconstrained model here, would not be sufficient for realistic vocabularies.
There are two possible solutions. First, 
the Bayesian framework would allow us to make our model non-parametric: the Bayesian GMM could be replaced by an infinite GMM~\citep{rasmussen_nips99,kamper_bayesgmm15} which infers the number of clusters automatically.
The second possibility is to use some heuristic to set $K$: in~\citep{rasanen+etal_interspeech15}, for example, an unsupervised syllabification method is used for presegmentation, and $K$ is then set as a proportion of the pseudo-syllables discovered in the first pass.

Finally, in this chapter we made a unigram word predictability assumption (Section~\ref{sec:tidigits_word_segmentation}) since the digit sequences do not have
any word-word dependencies. However, in a realistic corpus, such
dependencies will exist and could prove useful 
for
segmentation and lexicon discovery. In particular,~\cite{elsner+etal_emnlp13} showed
that for joint segmentation and clustering of noisy phone sequences, 
a bigram model was needed to improve
clustering accuracy.
Exact
computation of the extended model will be slow (e.g.\ the bigram
extension of equation~\eqref{eq:likelihood_fbgmm} requires marginalizing over the
cluster assignment of both the current and preceding embeddings) but reasonable
approximations might be possible. 
The development of these extensions and
approximations are considered in the next chapter.

\section{Summary and conclusions}

This chapter introduced a novel Bayesian model, operating on fixed-dimensional embeddings of speech, which segments and clusters unlabelled continuous speech into hypothesized word units---an approach which is very different from any presented before.
We applied the model to a small-vocabulary digit recognition task and compared performance to a more traditional HMM-based approach of a previous study.
Our model outperformed the baseline by more than 10\% absolute in unsupervised word error rate, without being constrained to a small number of word types (as the HMM was).
To obtain this performance, the acoustic embedding function (used to map variable duration segments to fixed-dimensional embeddings) were iteratively refined in a top-down manner using terms discovered by our system in an outer loop of segmentation.

Analysis showed that our model is strongly reliant on the acoustic word embeddings: when partial words are consistently mapped to a similar region in embedding space, the model proposes these as separate word types.
Most of the errors of the model were therefore due to consistent splitting of particular digits into partial-word clusters, or separate clusters for the same digit based on pronunciation variation.

For scaling this small-vocabulary system to larger vocabularies, we concluded that preprocessing is necessary to reduce the number of possible embeddings that the system needs to consider---embedding calculation being the main bottleneck of the approach.
We also concluded that the efficiency of the embedding method itself should be improved.
Fortunately, our model is not restricted to a particular embedding approach.
These challenges are addressed in the next chapter.
The system presented here makes use of standard acoustic features as input to the embedding function; in the next chapter we also consider using features obtained from the unsupervised representation learning method proposed in Chapter~\ref{chap:cae} (which combines top-down and bottom-up methodologies) as frame-level input to the segmentation model.

\graphicspath{{bucktsong/fig/}}

\chapter{Segmentation and clustering of large-vocabulary speech}
\label{chap:bucktsong}

Although the system of the previous chapter was able to accurately segment and cluster the small number of word types in the data, it was not possible to apply the same approach directly to realistic multi-speaker data with larger vocabularies, as discussed in Section~\ref{sec:tidigits_scaling}.
In this chapter we address these issues, and present an unsupervised large-vocabulary segmental Bayesian system.
To our knowledge, this is the first work that applies a full-coverage segmentation system to large-vocabulary multi-speaker data.

To improve efficiency, we use a bottom-up unsupervised
syllable boundary detection method~\citep{rasanen+etal_interspeech15}
to eliminate unlikely word
boundaries, reducing the number of potential word segments that need
to be considered. We also use a computationally much simpler embedding
approach based on downsampling~\citep{levin+etal_asru13}.
In this chapter, we investigate both MFCCs and frame-level features obtained from the correspondence autoencoder (Chapter~\ref{chap:cae}) as input to the segmental system.
This is the first time that unsupervised representation learning is incorporated into a full-coverage segmentation system.


We report results in both speaker-dependent and
speaker-independent settings on conversational speech datasets from
two languages: English and Xitsonga.
Xitsonga is a southern African Bantu language and is considered severely under-resourced because of the limited availability of transcribed speech data.
These datasets were also used as part of the
Zero Resource Speech Challenge (ZRS) at Interspeech 2015~\citep{versteegh+etal_interspeech15} and we
show that our system outperforms competing systems~\citep{lyzinski+etal_interspeech15,rasanen+etal_interspeech15} on
several of the ZRS metrics.
In particular, we find that by proposing a
consistent segmentation and clustering over a whole utterance, our
approach makes better use of the bottom-up syllabic constraints than
the purely bottom-up syllable-based system of~\cite{rasanen+etal_interspeech15}.
Moreoever, we
achieve similar $F$-scores for word tokens, types, and boundaries
whether training in a speaker-dependent or speaker-independent mode.
We also show that the discovered clusters can be made less speaker- and gender-specific by using features from the correspondence autoencoder.

The model of Chapter~\ref{chap:tidigits} makes a unigram word predictability assumption.
In a final set of experiments in this chapter, we consider a bigram model for improving clustering accuracy.
We show, however, that because of the peaked nature of the acoustic component of the model, the bigram model is unable to accurately learn word-word dependencies in the data.
This chapter is mainly based on the arXiv publication~\citep{kamper+etal_arxiv16}. 

\section{Related work and comparison to proposed model}
\label{sec:bucktsong_related_work}

For the general background most relevant to this chapter, see
Section~\ref{sec:background_fullcoverage}, which describes studies performing full-coverage segmentation of speech.
Also see the overview in Section~\ref{sec:tidigits_related_work} of how the segmental Bayesian model compares to previous word discovery and segmentation models, and the discussion in Section~\ref{sec:cae_background} of how the correspondence autoencoder relates to other studies on unsupervised representation learning.
This section summarizes work that relates specifically to the model and evaluation presented in the rest of the chapter.

\subsection{Unsupervised term discovery and syllable-based word segmentation}
\label{sec:bucktsong_utd}

As part of evaluation, we compare our approach to submissions to the lexical discovery track of the Zero Resource Speech Challenge (ZRS) at Interspeech 2015~\citep{versteegh+etal_interspeech15}.
The metrics used in the ZRS evaluate different aspects of unsupervised discovery in speech, not only for full-coverage segmentation systems, but also for unsupervised term discovery (UTD) systems.

The baseline provided as part of the ZRS is a UTD system~\citep{versteegh+etal_interspeech15} based on the earlier work of~\cite{jansen+vandurme_asru11}.
The other UTD submission to the ZRS by \cite{lyzinski+etal_interspeech15} extended the baseline system  using improved graph clustering algorithms.
We compare to both these systems in this chapter.
As was the case for the model in Chapter~\ref{chap:tidigits}, our system here
shares the property of UTD systems that it has no subword level of representation and operates directly on whole-word representations.
However, instead of representing each segment as a vector time series with variable duration as in UTD, we map each potential word segment to a fixed-dimensional acoustic word embedding; we can then define an acoustic model in the embedding space and use it to compare segments without performing dynamic time warping (DTW) alignment.
Our system also performs full-coverage segmentation and clustering, in contrast to UTD which segments and clusters only isolated acoustic patterns.

The third system we compare to is the ZRS submission of \cite{rasanen+etal_interspeech15}; we also draw extensively on their work to help us scale our approach to larger vocabularies.
Their full-coverage word segmentation system (also described in Section~\ref{sec:background_fullcoverage}) relies on an unsupervised method that predicts boundaries for syllable-like units, and then clusters these units on a per-speaker basis.
Using a bottom-up greedy mapping, reoccurring syllable clusters are then predicted as words.
From here onward we use \textit{syllable} to refer to the syllable-like units detected in the first step of their approach.

In our model, we incorporate the syllable boundary detection method of~\cite{rasanen+etal_interspeech15}, i.e.\ the first component of their system, as a presegmentation method to eliminate unlikely word boundaries.
Both human infants~\citep{emias_jasm99} and adults~\citep{mcqueen_memory98} use syllabic cues for word segmentation, and using such a bottom-up unsupervised syllabifier can therefore be seen as one way to incorporate prior knowledge of the speech signal into a zero-resource system~\citep{versteegh+etal_sltu16}.
Since we use the syllabification component of their approach in our own system, the two systems have much in common.
Apart from also only being able to predict word boundaries coinciding with syllable boundaries, R{\"a}s{\"a}nen et al.\ uses an averaging method to obtain fixed-dimensional acoustic embeddings of the syllable-like units, which is very similar to the downsampling scheme we use (Section~\ref{sec:downsampling}).
However, instead of $K$-means, we use a Bayesian GMM for clustering---earlier work~\citep{kamper+etal_slt14} suggests that there are benefits in the latter.
Furthermore, word discovery and segmentation 
are fundamentally different.
In~\citep{rasanen+etal_interspeech15}, a greedy bottom-up search is performed where reoccurring cluster sequences (ranging from three to one syllable) are treated as word candidates.
In contrast, our model samples word boundaries and cluster assignments consistently over complete utterances: although a particular boundary hypothesis might be less optimal in isolation, it could result in a better overall segmentation for that utterance.
Our model therefore imposes a top-down constraint in that utterances should be segmented consistently, while adhering to the bottom-up constraints from syllabification.
It is easy to also incorporate additional bottom-up constraints in our model (such as a minimum word duration), which would require additional heuristics in the pure bottom-up approach of R{\"a}s{\"a}nen et al.
Despite these benefits of our approach, we acknowledge that the algorithm in~\citep{rasanen+etal_interspeech15} is much simpler in terms of computational complexity and implementation than our~model.

\subsection{Combining unsupervised representation learning and word discovery}
\label{sec:buckstong_lit_unsuprepr}

Unsupervised representation learning, reviewed in Section~\ref{sec:background_phondisc}, refers to the task of 
learning a mapping from traditional frame-level acoustic features to a new frame-level representation which improves discrimination between linguistic units.
Presumably, UTD and full-coverage speech segmentation systems 
could benefit from such improved features, but few zero-resource studies have considered this.


A notable exception is the work of Zhang and Glass (described in Section~\ref{sec:background_bottomup}) who used bottom-up-trained GMM-UBM and unsupervised neural network features in segmental DTW systems for query-by-example search~\citep{zhang+glass_asru09,zhang+glass_icassp10,zhang+etal_icassp12}.
However, these bottom-up approaches do not take advantage of any weak top-down supervision, which have proven advantageous in several other zero-resource studies (as discussed in Section~\ref{sec:background_topdown}).

In this chapter we use a UTD system to discover word pairs and then use these word pairs as weak top-down supervision for the correspondence autoencoder (cAE), introduced in Chapter~\ref{chap:cae}.
We showed in Chapter~\ref{chap:cae} that features from the cAE intrinsically outperform traditional acoustic features as well as representations from other state-of-the-art bottom-up and top-down methods. 
Here we use cAE features as input to the large-vocabulary segmental Bayesian system.
To our knowledge, we are the first to use unsupervised representation learning within a full-coverage zero-resource segmentation system.
We argued in Section~\ref{sec:background_conclusion} that ideally unsupervised representation learning should be performed jointly with segmentation and clustering; 
here, these tasks are still not performed completely jointly since we use a separate UTD system, but the approach here is a first step in this direction.

\begin{figure}[tbp]
    \centering
    \includegraphics[width=\linewidth]{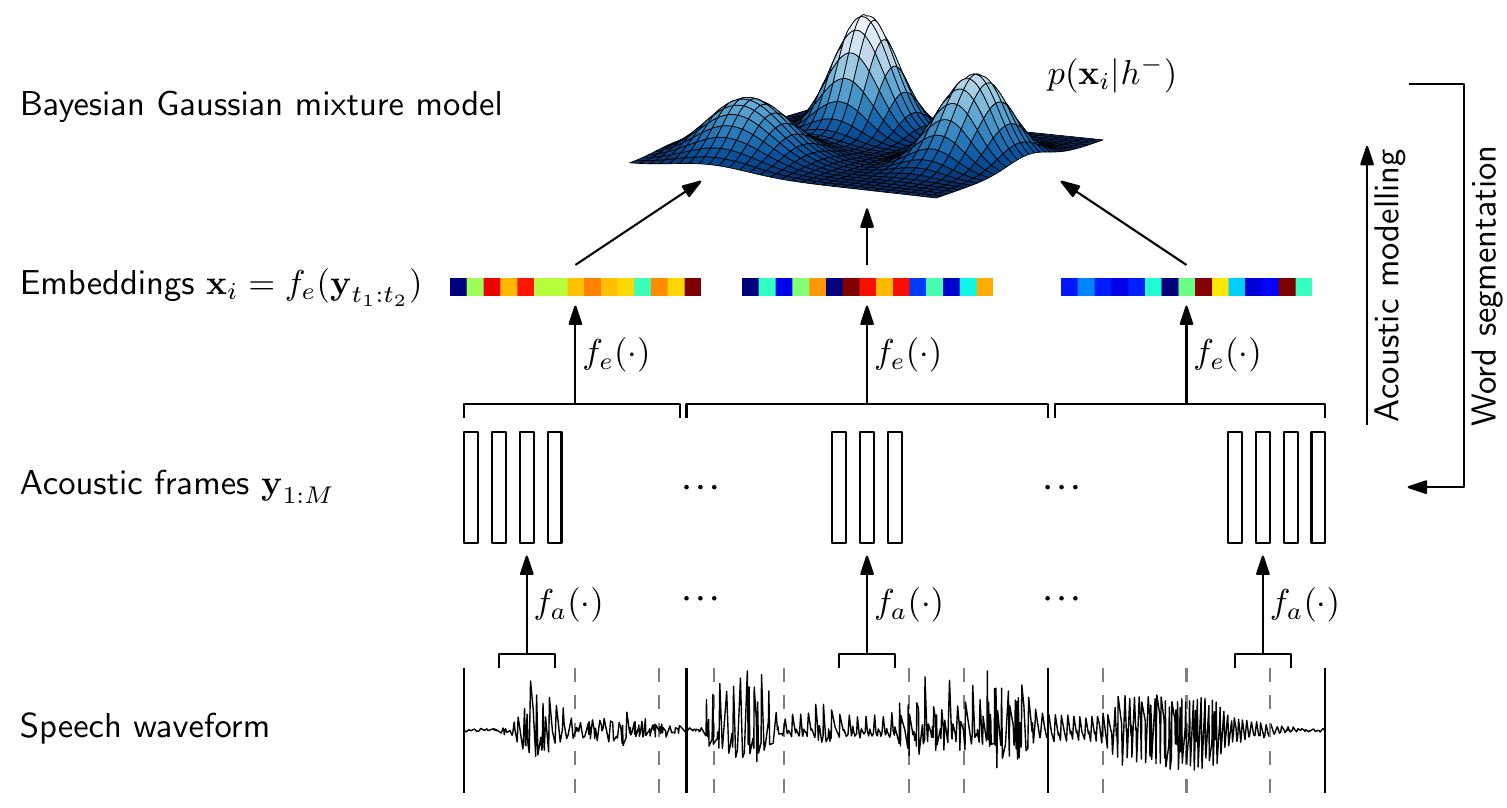}
    \caption{The large-vocabulary segmental Bayesian model. The dashed lines indicate where word boundaries are allowed according to an unsupervised syllable presegmentation algorithm. The function $f_a$ takes a window of raw signal and outputs a frame-level representation, while $f_e$ is an acoustic word embedding function which takes a variable number of frame-level features and outputs a single embedding vector.}
    \label{fig:bucktsong_unsup_repr_wordseg}
\end{figure}

\section{Large-vocabulary segmental Bayesian modelling}

The large-vocabulary segmental Bayesian model is illustrated in Figure~\ref{fig:bucktsong_unsup_repr_wordseg}.
The underlying model is the same as the small-vocabulary segmental Bayesian model introduced in Section~\ref{sec:tidigits_segmental_bayesian_model} of Chapter~\ref{chap:tidigits}.
While in the illustration of the small-vocabulary system in Figure~\ref{fig:unsup_wordseg} it was implicitly assumed that the speech waveform was converted to standard acoustic features, we denote the feature extraction function $f_a$ explicitly in Figure~\ref{fig:bucktsong_unsup_repr_wordseg} since here we also consider cAE features.
Furthermore, while the core model and blocked Gibbs sampler is the same (complete details in Section~\ref{sec:tidigits_segmental_bayesian_model}), the system here is different from the small-vocabulary case in that it has more components to cover the larger vocabulary, uses a different acoustic embedding function, and incorporates syllable-based bottom-level constraints.
These components are described in detail below.

\subsection{Unsupervised syllable boundary detection}
\label{sec:bucktsong_syllables}

Without any constraints, the input at the bottom of Figure~\ref{fig:bucktsong_unsup_repr_wordseg} could be segmented into any number of possible words using a huge number of possible segmentations.
In Chapter~\ref{chap:tidigits}, potential word segments were therefore required to be between 200~ms and 1~s in duration, and word boundaries were only considered at 20~ms intervals.
This still results in a very large number of possible segments.
Here we instead use a syllable boundary detection method to eliminate unlikely word boundaries, with word candidates spanning a maximum of six syllables.
On the waveform in Figure~\ref{fig:bucktsong_unsup_repr_wordseg}, solid and dashed lines are used to indicate the only positions where boundaries are considered during sampling, as determined by the syllabification method.
The syllabifier therefore implicitly defines the minimal unit in our system.

\begin{figure}[tbp]
    \centering
    \includegraphics[trim=1.9cm 22.8cm 10.9cm 2.5cm,clip=true,scale=1.5]{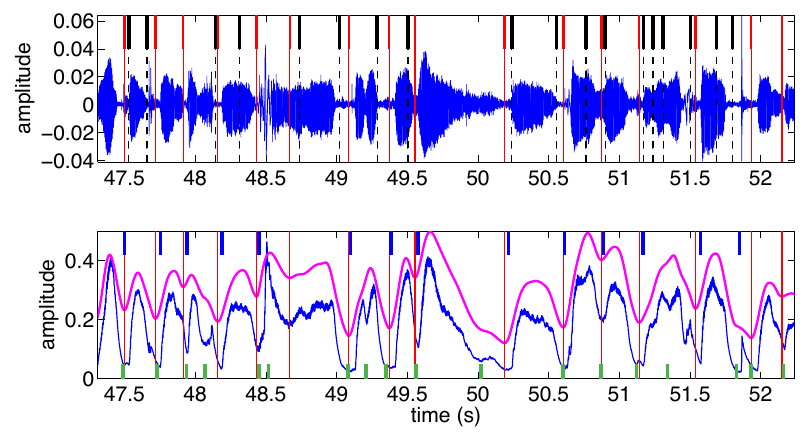}
    \caption{Example output from the unsupervised syllable boundary detection algorithm of \cite{rasanen+etal_interspeech15}.
    The top panel shows the original raw waveform. The bottom panel shows the signal's amplitude envelope in blue and the oscillator's amplitude in magenta.
    Dashed black lines indicate ground truth word boundaries, while red lines indicate the boundaries predicted according to the minima of the oscillator's amplitude. The small blue and green lines in the bottom panel shows the boundaries detected by two other syllabification methods considered in~\citep{rasanen+etal_interspeech15}.}
    \label{fig:rasanen_page3}
\end{figure}

\cite{rasanen+etal_interspeech15} evaluated several syllable boundary detection algorithms, and we use the best of these.
First the envelope of the raw waveform is calculated by downsampling the rectified signal and applying a low-pass filter.
Inspired by neuropsychological studies which found that neural oscillations in the auditory cortex occur at frequencies similar to that of the syllabic rhythm in speech, the calculated envelope is used to drive a discrete time oscillation system with a centre frequency of typical syllabic rhythm.
Minima in the oscillator's amplitude give the predicted syllable boundaries.
Figure~\ref{fig:rasanen_page3}, taken directly from~\citep{rasanen+etal_interspeech15}, shows example output.
In this work, we use syllabification code kindly provided by the authors of~\citep{rasanen+etal_interspeech15}; this is an updated version of the code used in their original publication, and we use this latest version without any modification and with the default parameter settings.


\subsection{Acoustic word embeddings and unsupervised representation learning}
\label{sec:downsampling}

Given the large number of embeddings that still need to be calculated, the embedding method used in Chapter~\ref{chap:tidigits} is prohibitively slow.
The approach there (described in Section~\ref{sec:tidigits_embeddings}) involved constructing a reference vector by calculating the DTW alignment costs to a large number of exemplars in a reference set, each 
DTW alignment 
having quadratic time complexity.
Finding the required reference set was itself a challenge: an outer loop of segmentation was used to to refine a randomly-initialized exemplar set with automatically discovered segments (Section~\ref{sec:tidigits_iterating}).

A simple and fast approach to obtain acoustic word embeddings is to uniformly {downsample} so that any segment is represented by the same fixed number of vectors~\citep{ossama+etal_interspeech13,levin+etal_asru13}.
A similar approach is to divide a segment into a fixed number of intervals and average the frames in each interval~\citep{lee+lee_taslp13,rasanen+etal_interspeech15}.
The downsampled or averaged frames are then flattened to obtain a single fixed-length vector.
For each segment to embed, this is a single linear operation in the segment length.
Although these very simple approaches are less accurate at word discrimination than the approach used  in~Chapter~\ref{chap:tidigits}, they 
have been effectively used in several studies, including~\citep{rasanen+etal_interspeech15}, and are computationally much more efficient.
Here we use \textit{downsampling} as the acoustic word embedding function $f_e$ in Figure~\ref{fig:bucktsong_unsup_repr_wordseg}: we keep ten equally-spaced vectors from a segment and use a Fourier-based method for smoothing~\citep{levin+etal_asru13}.\footnote{We use the \texttt{signal.resample} function from the SciPy package. See \url{http://docs.scipy.org/doc/scipy-0.17.0/reference/generated/scipy.signal.resample.html} for details.}

Figure~\ref{fig:bucktsong_unsup_repr_wordseg} shows that 
$f_e$ takes as input a sequence of frame-level features from the feature extracting function $f_a$. 
One option for $f_a$ is to simply use MFCCs.
As an alternative,  unsupervised representation learning is incorporated into the approach by using the cAE as a feature extractor.
Complete details of the cAE are given in Chapter~\ref{chap:cae}, but the training procedure is briefly outlined here for the sake of completeness. 
The UTD system of~\cite{jansen+vandurme_asru11} is used to discover word pairs which serve as weak top-down supervision.
The cAE operates at the frame level, so 
the word-level constraints are converted to frame-level constraints
by aligning 
each word pair using DTW. Taken together across all discovered pairs, this results in a set of $F$ frame-level pairs $\left\{ \left(\vec{y}_{i,a}, \vec{y}_{i,b} \right) \right\}_{i=1}^{F}$.
Here, each frame is a single MFCC vector.
For every pair $\left(\vec{y}_{a}, \vec{y}_{b}\right)$, $\vec{y}_{a}$ is presented as input to the cAE while $\vec{y}_{b}$ is taken as output, and vice versa. 
The cAE consists of several non-linear layers which are initialized by pretraining the network as a standard stacked autoencoder. 
The cAE is then tasked with reconstructing $\vec{y}_{b}$ from $\vec{y}_{a}$, using the loss $\left|\left|\vec{y}_{b} - \vec{y}_{a}\right|\right|^2$. 
To use the trained network as a feature extractor $f_a$, the activations in one of its middle layers are taken as the new feature representation. 

\subsection{Towards bigram modelling}
\label{sec:bucktsong_bisampling_assignments}

The segmental Bayesian model we used in Chapter~\ref{chap:tidigits} makes a unigram assumption of word predictability, as explained in Section~\ref{sec:tidigits_gmm}.
Here we consider extending the model by including a bigram model of word predictability.
Compared to the unigram algorithm outlined in Section~\ref{sec:tidigits_segmental_bayesian_model}, 
changes need to be made in 
sampling the component assignments (line~\ref{alg_line:fbgmm_inside_loop} in Algorithm~\ref{alg:gibbs_wordseg}), and in sampling the segmentation (lines~\ref{alg_line:forward} and~\ref{alg_line:sample_bounds}).
For the experiments in this chapter, we only consider the first of these.
If clustering is not improved by the bigram model, there would be little hope in improving segmentation since the latter is based on marginalizing over cluster assignments.
However, all the equations required for bigram segmentation are derived in Appendix~\ref{appen:derivations_bigramseg}.

As in the unigram Gibbs sampler, the bigram model samples the component assignment $z_i$ of acoustic embedding $\vec{x}_i$ conditioned on all the other current component assignments $\vec{z}_{\backslash i}$ according to
\begin{align}
    P(z_i = k|\vec{z}_{\backslash i}, \mathcal{X} ; \vec{\gamma}, \vec{\beta} ) 
    &\propto P(z_i = k|\vec{z}_{\backslash i} ; \vec{\gamma})  p(\vec{x}_i | \mathcal{X}_{k \backslash i}; \vec{\beta})
    \label{eq:bucktsong_bicollapsed1}
\end{align}
This equation is the same as~\eqref{eq:collapsed1}, with the language modelling hyperparameters denoted here as $\vec{\gamma}$.
The term $p(\vec{x}_i | \mathcal{X}_{k \backslash i}; \vec{\beta})$ is given by the acoustic model, and will be the same for the unigram and bigram models.
However, instead of using~\eqref{eq:first_term7}, the term $P(z_i = k|\vec{z}_{\backslash i} ; \vec{\gamma} ) = P(z_i = k|z_{i - 1} = l, \vec{z}_{\backslash i, i - 1}; \vec{\gamma})$ needs to be defined differently for the bigram model.
The simplest option would be to use a smoothed bigram maximum likelihood estimate~\citep{bell+etal_book46}:
\begin{align}
    \hat{P}(z_i = k|z_{i - 1} = l, \vec{z}_{\backslash i, i - 1} ; \vec{\gamma})
    = \frac{N_{k|l\backslash i}  + b/K }{N_{l\backslash i} + b} = \frac{N_{k|l}  + b/K }{N_{l} + b} \label{eq:basic_ml_bigram}
\end{align}
where $b$ is a smoothing parameter and we drop the $\backslash i$ subscript on the right hand side (here we implicitly assume that $z_i$ is not taken into account in any counts).
$N_{k|l}$ is the number of times that a segment is assigned to component $k$ with the preceding segment assigned to component $l$.  $N_{l}$ is the number of segments that was assigned to component $l$.
We could improve on~\eqref{eq:basic_ml_bigram} by interpolating the estimated bigram probabilities with estimated unigram probabilities~\citep{jelinek+mercer_prp80}, giving the interpolated estimator~\citep{mackay+peto_nle95}:
\begin{equation}
    \hat{P}(z_i = k|z_{i - 1} = l, \vec{z}_{\backslash i, i - 1} ; \vec{\gamma}) = \lambda \frac{N_{k} + a/K}{N + a} + (1 - \lambda) \frac{N_{k|l}  + b/K }{N_{l} + b} \label{eq:interpolated_lm}
\end{equation}
where the interpolation weight and smoothing parameters are denoted together as $\vec{\gamma} = (\lambda, a, b)$.
Again, $N$ here does not contain the count for $z_i$, which is why the denominator in the first term is not reduced by $1$, as in \eqref{eq:first_term7}.

Rather than using smoothed maximum likelihood estimation, a more principled way would be to use the Bayesian hierarchical Dirichlet language model, as presented by \cite{mackay+peto_nle95}.
However, the resulting sampling equation is very similar to~\eqref{eq:interpolated_lm}, except that they derive an automatic re-estimation procedure for setting the language modelling hyperparameters $\vec{\gamma}$.
As a first approach, our bigram segmental model is simply based on~\eqref{eq:interpolated_lm} and we set the hyperparameters explicitly.

The scales of the language modelling and acoustic terms in~\eqref{eq:bucktsong_bicollapsed1} can be very different; this is also the case in standard supervised speech recognition systems, and a language model scaling factor is typically used to deal with this difference in scales~\citep[p.~187]{htk_book}.
We therefore introduce another hyperparameter, the language model scaling factor $\eta$, and, rather than using~\eqref{eq:bucktsong_bicollapsed1} directly, sample component assignments according to
\begin{equation}
    P(z_i = k|\vec{z}_{\backslash i}, \mathcal{X} ; \vec{\gamma}, \vec{\beta} ) 
    \propto \left[ P(z_i = k|\vec{z}_{\backslash i} ; \vec{\gamma}) \right]^\eta  p(\vec{x}_i | \mathcal{X}_{k \backslash i}; \vec{\beta})
    \label{eq:bucktsong_bicollapsed2}
\end{equation}
When $\eta = 1$, we recover~\eqref{eq:bucktsong_bicollapsed1}; when $\eta=0$, a uniform language model is effectively used.

In summary, the Gibbs sampling algorithm for this limited bigram model is exactly the same as Algorithm~\ref{alg:gibbs_wordseg}, but instead of using~\eqref{eq:collapsed1} in line~\ref{alg_line:fbgmm_inside_loop}, we sample component assignments from left to right using~\eqref{eq:interpolated_lm} and~\eqref{eq:bucktsong_bicollapsed2}.

\section{Experiments}

\subsection{Experimental setup and evaluation}
\label{sec:bucktsong_exp_setup}

We use three datasets, summarized in Table~\ref{tbl:bucktsong_data}. 
The first two are disjoint subsets extracted from the Buckeye corpus of conversational English~\citep{pitt+etal_speechcom05}, while the third is a portion of the Xitsonga section of the NCHLT corpus of languages spoken in South Africa~\citep{devries+etal_speechcom14}.
Xitsonga is a Bantu language spoken in southern Africa; although it is considered under-resourced, more than five million people use it as their first language.\footnote{\url{http://www.ethnologue.com/language/tso}} 

\begin{table}[tbp]
    \mytable
    \caption{Statistics for the three datasets used in this chapter. All sets have an equal number of female and male speakers. The last column is an average.}
    \begin{tabularx}{0.918\linewidth}{@{}lCCCCC@{}}
        \toprule
        Dataset & Duration (hours) & No. of speakers & Total word tokens & Total word types & Word types per speaker \\
        \midrule
        English1 & 6.0 & 12 & 89\,681 & 5129 & 1104 \\
        English2 & 5.0 & 12 & 69\,543 & 4538 & 966 \\
        Xitsonga & 2.5 & 24 & 19\,848 & 2288 & 333 \\
        \bottomrule
    \end{tabularx}
    \label{tbl:bucktsong_data}
\end{table}

The two sets extracted from Buckeye, referred to as English1 and English2, respectively contain five and six hours of speech, each from twelve speakers (six female and six male).
The Xitsonga dataset consists of 2.5 hours of speech from 24 speakers (twelve female, twelve male).
English2 and the Xitsonga data were used as test sets in the ZRS challenge, so we can compare our system to others using the same data and evaluation framework~\citep{versteegh+etal_interspeech15}.
English1 was extracted for development purposes from a disjoint portion of Buckeye to match the distribution of speakers in English2.
For all three sets, speech activity regions are taken from forced alignments of the data, as was done in the ZRS.
From Table~\ref{tbl:bucktsong_data}, the average duration of a word in an English set is around 250~ms, while for Xitsonga it is about 450~ms.

We run our model on
all sets separately---in each case, unsupervised modelling and evaluation is performed on the same set.
English1 is the only set used for any development (specifically for setting hyperparameters) in any of the experiments; both English2 and Xitsonga are treated as unseen final test sets.
This allows us to see how hyperparameters generalize within language on data of similar size, as well as across language on a corpus with very different characteristics.

\subsection{Evaluation}
\label{sec:metrics}

\begin{figure}[tbp]
    \centering
    \includegraphics[width=0.918\linewidth]{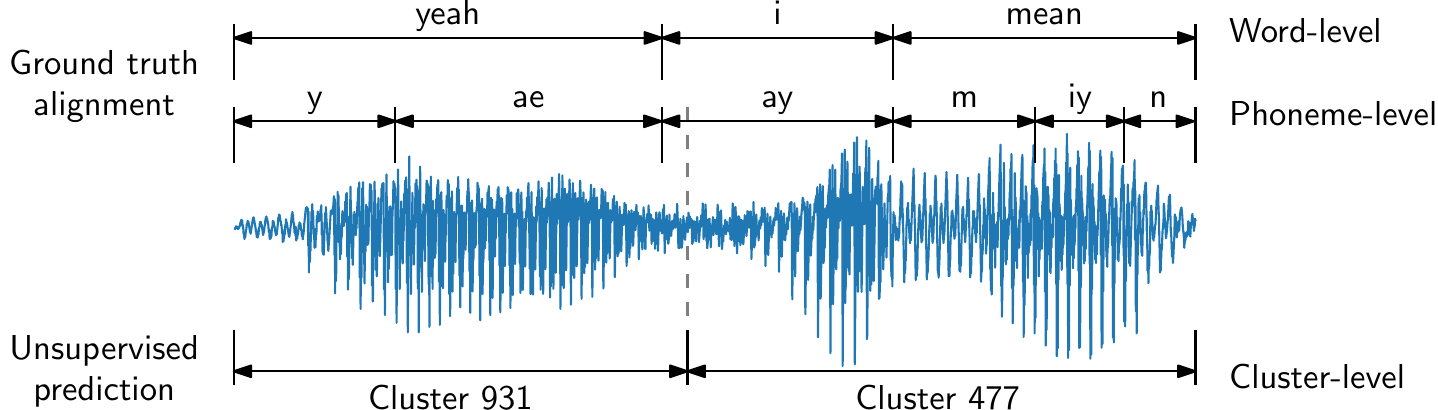}
    \caption{Illustration of the mapping of clusters to true labels for evaluation. Ground truth alignments are shown at the top, with actual output from speaker-dependent BayesSegMinDur-cAE at the bottom.}
    \label{fig:bucktsong_evaluation}
\end{figure}

The evaluation of zero-resource systems that segment and cluster speech is a research problem in itself, as described in Section~\ref{sec:background_eval}.
We use a range of metrics that have been proposed before, all performing some mapping from the discovered structures to ground truth forced alignments of the data, as illustrated in Figure~\ref{fig:bucktsong_evaluation}.
Note that some of these metrics are defined slightly differently from those used in Chapter~\ref{chap:tidigits}; this is in order to be more consistent with the studies that we compare to here.

\myemph{Average cluster purity:} Every discovered {token} is first aligned to the ground truth word token with which it overlaps most. In Figure~\ref{fig:bucktsong_evaluation} the token assigned to cluster 931 would be mapped to the true word `yeah', and the 477-token mapped to `mean'.
Every discovered word {type} (cluster) is then mapped to the most common ground truth word type in that cluster.
For example, if most of the other tokens in cluster 931 are also labelled as `yeah', then cluster 931 would be labelled as `yeah'.
Average purity is then defined as the total proportion of correctly mapped tokens in all clusters.
For this metric, more than one cluster may be mapped to a single ground truth type, i.e.\ many-to-one~\citep{sun+vanhamme_csl13}.

\myemph{Unsupervised word error rate (WER/WER$_\text{m}$):} A similar word-level mapping to the above is used to align the mapped decoded output from
a system to the ground truth transcriptions.
Based on this alignment we calculate $\textrm{WER} = \frac{S + D + I}{N}$, with $S$ the number of substitutions, $D$ deletions, $I$ insertions, and $N$ the tokens in the ground truth~\citep{chung+etal_icassp13,walter+etal_asru13}.
The cluster mapping can be done in one of two ways: many-to-one, where more than one cluster can be assigned the same word label (as in cluster purity), or using a greedy one-to-one mapping, where at most one cluster is mapped to a ground truth word type.
The latter, which we denote simply as WER, might leave some cluster unassigned and these are counted as errors; this was done for the evaluation in Chapter~\ref{chap:tidigits}.
For the former, denoted as WER$_\text{m}$, all clusters are labelled.
Depending on the downstream speech task, it might be acceptable to have multiple clusters that correspond to the same true word; WER penalizes such clusters, while WER$_\text{m}$ does not.
These metrics are useful since they are easily interpretable and well-known in the speech community.

\myemph{Normalized edit distance (NED):} This is the first of the ZRS metrics (the rest follow).
These metrics use a phoneme-level mapping: each discovered token is mapped to the sequence of ground truth phonemes of which at least 50\% or 30~ms are covered by the discovered segment~\citep{ludusan+etal_lrec14,versteegh+etal_interspeech15}.
In Figure~\ref{fig:bucktsong_evaluation}, the 931-token would be mapped to \mbox{/y ae/} and the 477-token to \mbox{/ay m iy n/}.
For a pair of discovered segments, the edit distance between the two phoneme strings is divided by the maximum of the
length of the two strings. This is averaged over all pairs predicted to be of the same type (cluster), to obtain the final NED score. If all segments in each cluster have the same phoneme string, then $\textrm{NED} = 0$, while if all phonemes are different, $\textrm{NED} = 1$.
NED is useful in that it does not assume that the discovered segments need to correspond to true words (as in cluster purity and WER), and it only considers the patterns returned by a system (so it does not require full coverage, as WER does).
As an example, if a cluster contains \mbox{/m iy/} from a realization of the word `meaningful' and a token \mbox{/m iy n/} from the true word `mean', then NED would be $1/3$ for this two-token~cluster.

\myemph{Word boundary precision, recall, $F$-score:} The word boundary positions proposed by a system are compared to those from forced alignments of the data, falling within some tolerance.
A tolerance of 20~ms is often used~\citep{lee+etal_tacl15}, but for the ZRS the tolerance is 30 ms or 50\% of a phoneme (to match the mapping).
In Figure~\ref{fig:bucktsong_evaluation} the detected boundary (dashed line) would be considered correct if it is within
the tolerance from the true word boundary between `yeah' and `i'.

\myemph{Word token precision, recall, $F$-score:} These measure how accurately proposed word tokens match ground truth word tokens in the data.
In contrast to the word boundary scores, both boundaries of a predicted word token need to be correct.
In Figure~\ref{fig:bucktsong_evaluation}, the system would receive credit for the 931-token since it is mapped to \mbox{/y ae/} and therefore match the ground truth word token `yeah'.  However, the system would be penalized for the 477-token (mapped to \mbox{/ay m iy n/}) since it fails to predict word tokens corresponding to \mbox{/ay/} and \mbox{/m iy n/} (the ground truth words `i' and `mean').
Both the word boundary and word token metrics give a measure of how accurately a system is segmenting its input into word-like units.

\myemph{Word type precision, recall, $F$-score:} The set of distinct phoneme mappings from the tokens returned by a system is compared to the set of true word types in the ground truth alignments. If any discovered word token maps to a phoneme sequence that is also found as a word in the ground truth vocabulary, the system is credited for a correct discovery of that word type. 
For example if the type \mbox{/y ae/} (as in `yeah') occurs in the ground truth alignment, the system needs to return at least one token that is mapped to \mbox{/y ae/}.

We evaluate our model in 
both speaker-dependent and speaker-independent settings.
Multiple speakers make it more difficult to discover accurate clusters: non-matching linguistic units might be more similar within-speaker than matching units across speakers.
For the speaker-dependent case, the model is run and scores are computed on each speaker individually, then performance is averaged over speakers.
In the speaker-independent case, the system is run and scores computed over the entire multi-speaker dataset at once.
This typically results in worse purity, NED and WER$_\text{m}$ scores since the task is more difficult and clusters are noisier.
WER is affected even more severely due to the one-to-one mapping that it uses; if there are two perfectly pure clusters that contain tokens from the same true word, but the two clusters are also perfectly speaker-dependent, then only one of these clusters would be mapped to the true word type and the other would be counted as errors.
Despite the adverse effect on these metrics, it is of practical importance to evaluate a zero-resource system in the speaker-independent~setting.

\subsection{Model implementation and hyperparameters}
\label{sec:bucktsong_modeldev}

Most model hyperparameters are set according to previous work.
The changes made here are based exclusively on performance on English1. 

Training parameters for the cAE are based on Chapter~\ref{chap:cae} and~\citep{renshaw+etal_interspeech15}.
The cAE model is pretrained on all data (in a particular set) for 5 epochs using minibatch stochastic gradient descent with a batch size of 2048 and a fixed learning rate of $2\cdot10^{-3}$.
Subsequent correspondence training is performed for 120 epochs using a learning rate of $32\cdot10^{-3}$.
Each pair is presented in both directions as input and output.
Pairs are extracted using the UTD system of~\cite{jansen+vandurme_asru11}: for English1, 14\,494 word pairs are discovered; for English2, 10\,769 pairs; and for Xitsonga, 6979.
The cAE is trained on each of these sets separately.
In all cases, the model consists of nine hidden layers of 100 units each, except for the eighth layer which is a bottleneck layer of 13 units.
We use $\tanh$ as non-linearity.
The position of the bottleneck layer is based on intrinsic evaluation on English1.
Although it is common in DNN speech systems to use nine or eleven sliding frames as input, we use 
single-frame MFCCs with first and second order derivatives (39-dimensional), as also done in Chapter~\ref{chap:cae} and~\citep{renshaw+etal_interspeech15}.
For feature extraction, the cAE is cut at the bottleneck layer, resulting in 13-dimensional output (chosen to match the dimensionality of the static MFCCs). 
\edit{The combination of pretraining and correspondence training takes around $5$ hours on a single GTX TITAN GPU.}
For both the MFCC and cAE acoustic word embeddings, we downsample a segment to ten frames, resulting in 130-dimensional embeddings. 
\edit{Extracting all possible downsampled embeddings for one of the datasets takes less than one minute on a single $2.8~\textrm{GHz}$ CPU.}
As in Chapter~\ref{chap:tidigits} and~\citep{kamper+etal_slt14}, embeddings are normalized to the unit sphere. 

For the acoustic model 
we use the following hyperparameters, as in Chapter~\ref{chap:tidigits}: all-zero vector for $\vec{\mu}_0$, $\sigma_0^2 = \sigma^2/\kappa_0$ and $\kappa_0 = 0.05$. 
For MFCC embeddings we use $\sigma^2 = 1\cdot10^{-3}$ for the fixed shared spherical covariance matrix, while for cAE embeddings we use $\sigma^2 = 1\cdot10^{-4}$. This was based on speaker-dependent English1 performance.
We found that $\sigma^2$ is one of the parameters most sensitive to the input representation and often requires tuning; generally, however, it is robust if it is chosen small enough (in the ranges used here).
For the unigram model we use $a = 1$ without any language model scaling, i.e.\ $\eta = 1$. For the bigram model (Section~\ref{sec:bucktsong_bisampling_assignments}), we use $a = 1$, $b = 1$, $\lambda = 0.1$ and we vary $\eta$.

We use the updated version of the oscillator-based syllabification system of \cite{rasanen+etal_interspeech15} without modification.
Word candidates are limited to span a maximum of six syllables.
As discussed in Section~\ref{sec:tidigits_scaling} in the previous chapter, it is difficult to decide beforehand how many potential word clusters (the number of components $K$ in the acoustic model) we need.
Here we follow the same approach as in~\citep{rasanen+etal_interspeech15}: we choose $K$ as a proportion of the number of discovered syllable tokens.
For the speaker-dependent settings, we set $K$ as $20\%$ of the number of syllables, based on English1 performance.
On average, this amounts to $K = 1549$ on English1, $K = 1195$ on English2, and $K = 298$ on Xitsonga.
Compared to the average number of word types per speaker shown in Table~\ref{tbl:bucktsong_data}, these numbers are higher for the English sets and slightly lower for Xitsonga.
For speaker-independent models, we use $5\%$ of the syllable tokens, amounting to $K = 4647$ on English1, $K = 3584$ on English2, and $K = 1789$ on Xitsonga.
These are lower than the true number of total word types shown in Table~\ref{tbl:bucktsong_data}.
On English1, speaker-independent performance did not improve when using a larger $K$ and inference was much slower.

Although we use a fixed number of components $K$ here, the Bayesian GMM does not need to use all the components.
In many of the experiments in Section~\ref{sec:buckstong_bigram_exp}, the model empties out several of its components, with $K$ acting as a maximum.
Sparsity can be controlled through the covariance parameter $\sigma^2$, with less components being used when the spherical covariance is made larger and vice versa.
We used this property in the small-vocabulary system of Chapter~\ref{chap:tidigits} to, in effect, automatically discover the number of word types in the data (by setting $K$ much larger than the true number).
However, speaker-dependent development on English1 indicated that for the large-vocabulary model here, it is more robust to set $K$ as a proportion of the syllable tokens (as explained in the previous paragraph) and then set $\sigma^2$  small enough that all components are used.

To improve sampler convergence, we use simulated annealing~\citep{goldwater+etal_cognition09}, by raising the boundary probability in~\eqref{eq:backward} to the power $\frac{1}{\xi}$ before sampling, where $\xi$ is a temperature parameter.
As in Chapter~\ref{chap:tidigits}, we found that convergence is improved by first running the sampler without sampling boundaries.
In all experiments we do this for 15 iterations.
Subsequently, the complete sampler is run for $J = 15$ Gibbs sampling iterations with 3 annealing steps in which $\frac{1}{\xi}$ is increased as $[0.01, 0.5, 1]$.
Word boundaries are initialized randomly by setting boundaries at allowed locations with a $0.25$ probability.
\edit{In the speaker-dependent setting, we apply the model in parallel to each of the speakers; this takes around $30~\textrm{minutes}$ for a single speaker on a single $2.8~\textrm{GHz}$ CPU.
Applying the model to all speakers together in the speaker-independent setting takes around $40~\textrm{hours}$ on a single $2.8~\textrm{GHz}$ CPU.}

Given the common setup above, we consider three variants of our approach:

\myemph{BayesSeg} is the most general segmental Bayesian model. In this model, a word segment can be of any duration, as long as it spans less than six syllables.

\myemph{BayesSegMinDur} is the same as BayesSeg, but requires word candidates to be at least 250~ms in duration; on English1, this improved performance on several metrics. Such a minimum duration constraint is also used in most UTD systems~\citep{park+glass_taslp08,jansen+vandurme_asru11}, and was used in the system of Chapter~\ref{chap:tidigits}.

\myemph{SyllableBayesClust} clusters the discovered syllable tokens using the Bayesian GMM, but does not sample word boundaries. It can be seen as a baseline for the two models above, where segmentation is turned off and the detected syllable boundaries are set as initial (and permanent) word boundaries.
All word candidates therefore span a single syllable in this model.

%

\subsection{Intermediate result: Intrinsic evaluation of features and embeddings}

In Section~\ref{sec:background_phondisc_eval}, the \textit{same-different task} was introduced: a word discrimination task which can be used to provide an intrinsic evaluation of either frame-level or acoustic embedding representations without the need to train a full recognition or discovery system~\citep{carlin+etal_icassp11}.
Here we briefly note performance on this task as an intermediate evaluation of the representations used in our full system.
In the same-different task, we are given a pair of acoustic segments, each corresponding to a word, and we must decide whether the segments are instances of the same or different words.
The task can be approached in a number of ways depending on the features being evaluated: for frame-level features, a DTW score between segments is used, while for acoustic word embeddings, the Euclidean or cosine distance between embeddings is used.
Two words can then be classified as being of the same or different type based on some threshold, and a precision-recall curve is obtained. 
To evaluate embeddings across different operating points, the area under the precision-recall curve is used as the final evaluation metric, referred to as the average precision (AP).

\begin{table}[tbp]
    \mytable
    \caption{Same-different performance of different representations. In the first two rows, DTW alignment over all frames is used for segment comparison; dimensionality (dim.) is at the frame level. In the last two rows, cosine distance between fixed-dimensional acoustic word embeddings (obtained by downsampling to ten frames) is used and the size of the embedding vectors is given. Higher AP is better.}
    \begin{tabularx}{0.918\linewidth}{@{}lccCCCC@{}}
        \toprule
        & & & \multicolumn{3}{c}{Average precision (AP)} \\
        \cmidrule{4-6}
        Features & Dim. & Metric & English1 & English2 & Xitsonga \\
        \midrule
        MFCCs with CMVN                 & 39 & DTW & 0.339 & 0.359 & 0.281 \\
        13-dimensional cAE features     & 13 & DTW & 0.474 & 0.429 & 0.552 \\
        \addlinespace        
        Downsampled MFCC embeddings     & 130 & cosine & 0.193 & 0.212 & 0.147 \\
        Downsampled cAE embeddings      & 130 & cosine & 0.251 & 0.228 & 0.299 \\
        \bottomrule
    \end{tabularx}
    \label{tbl:bucktsong_ap}
\end{table}

Table~\ref{tbl:bucktsong_ap} gives the same-different performance of different representations. 
Here each test set contains all words of at least 5 characters and 0.5 seconds in duration from all speakers in the corresponding dataset (i.e.\ speaker-independent evaluation).
For the cAE, we use activations from the 13-unit bottleneck layer as features, as described in the previous section.
The first two rows in Table~\ref{tbl:bucktsong_ap} give performance when representing each test segment using all its frames, and segments are compared using DTW; 
this is a direct evaluation of the output of $f_a$ in Figure~\ref{fig:bucktsong_unsup_repr_wordseg}.
As in Chapter~\ref{chap:cae}, the cAE features provide a large gain over MFCCs with cepstral mean and variance normalization (CMVN).\footnote{For the first line, both first and second order derivatives are used, yielding 39-dimensional MFCCs. This gives better performance than using only the static coefficients. For the third line (downsampled MFCCs), only the static MFCCs are used in order to keep the acoustic embedding dimensionality reasonable (130-dimensional since we are downsampling to ten frames).}
This is especially pronounced on Xitsonga, which has the lowest AP (0.281) for the MFCC features, but the highest AP (0.552) for the cAE features.
The performance when downsampling each test segment to ten frames (the output of $f_e$) is given in the third and fourth rows using the MFCC and cAE features, respectively.
Again, cAE features perform best.
The improvements on English2, however, are not as pronounced as they are on the other two datasets.
The biggest improvement is again on Xitsonga.

\subsection{Results: Error rates and analysis}
\label{sec:bucktsong_results_own}

\subsubsection{Speaker-dependent models}

Table~\ref{tbl:bucktsong_sd_wer} shows one-to-one and many-to-one WERs for the different speaker-dependent models on the three datasets.
The trends in WER using one-to-one and many-to-one mappings are similar, with the absolute performance of the latter consistently better by around 10\% to 20\% absolute.
The performance on Xitsonga varies much more dramatically than on the English datasets, with WER ranging from around 140\% to 75\% and WER$_\text{m}$ from 135\% to 69\%.\footnote{From its definition, WER is more than 100\% if there are more substitutions, deletions and insertions than ground truth tokens.}
Table~\ref{tbl:bucktsong_data} shows that the characteristics of the Xitsonga data are quite different from the English sets.
For the speaker-dependent case here, much less data is available per Xitsonga speaker (just over six minutes on average) than for an English speaker (more than ten minutes), which might (at least partially) explain why error rates vary much more dramatically on Xitsonga.
Moreover, there is a much higher proportion of multisyllabic words in Xitsonga~\citep{rasanen+etal_interspeech15}, as reflected in the average duration of words which is almost twice as long in the Xitsonga than in the English data (Section~\ref{sec:bucktsong_exp_setup}).

\begin{table}[tbp]
    \mytable
    \caption{Performance on the three datasets for speaker-dependent models.}
    \begin{tabularx}{\linewidth}{@{}l@{\ }c@{\ \ }CCCCCC@{}}
        \toprule
        & & \multicolumn{3}{c}{One-to-one WER (\%)} & \multicolumn{3}{c}{Many-to-one WER$_\text{m}$ (\%)} \\
        \cmidrule{3-5} \cmidrule(l){6-8}
        Model & Embeds. & English1 & English2 & Xitsonga & English1 & English2 & Xitsonga \\
        \midrule
        SyllableBayesClust & MFCC & 93.3 & 94.1 & 140.3 & 72.4 & 76.1 & 134.5 \\
        BayesSeg & MFCC & 89.2 & 88.8 & 116.2 & 68.3 & 70.5 & 109.5 \\
        BayesSegMinDur & MFCC & 83.7 & 82.8 & 78.9 & 67.6 & 68.3 & 71.7 \\
        BayesSeg & cAE & 89.3 & 89.3 & 107.9 & 70.0 & 73.0 & 100.5 \\
        BayesSegMinDur & cAE & 85.2 & 84.1 & 75.9 & 70.6 & 71.2 & 68.8 \\
        \bottomrule
    \end{tabularx}    
    \label{tbl:bucktsong_sd_wer}
\end{table}

Comparing the results for the three
systems using MFCC features indicates that, on all three datasets,
allowing the system to infer word boundaries across multiple syllables
(BayesSeg) yields better performance than treating each syllable as a
word candidate (SyllableBayesClust). Incorporating a minimum duration
constraint (BayesSegMinDur) improves performance further. The relative
differences between these systems are much more pronounced in
Xitsonga, presumably due to the higher proportion of multisyllabic
words.
Table~\ref{tbl:bucktsong_sd_wer} also shows that in most cases the cAE features perform
similarly to MFCC features in these speaker-dependent systems,
although there is a large improvement in Xitsonga for the BayesSeg
system when switching to cAE features (from 116.2\% to 107.9\% in WER and from 109.5\% to 100.5\% in WER$_\text{m}$).

To get a better insight into the types of errors that the models make, Tables~\ref{tbl:bucktsong_sd_english2} and~\ref{tbl:bucktsong_sd_tsonga} give a breakdown of word boundary detection scores, individual error rates, and average cluster purity on English2 and Xitsonga, respectively.
A word boundary tolerance of 20~ms is used, as in~\citep{lee+etal_tacl15}, with a greedy one-to-one mapping for calculating error rates.
SyllableBayesClust gives an upper-bound for word boundary recall since every syllable boundary is set as a word boundary.
The low recall (28.9\% and 24.8\%)
could potentially be improved by using a better syllabification method, but we decided to use the method as provided, and leave further investigation for future~work.

Table~\ref{tbl:bucktsong_sd_english2} shows that on English2, the MFCC-based BayesSeg and BayesSegMinDur models under-segment compared to SyllableBayesClust, causing systematically poorer word boundary recall and $F$-scores and an increase in deletion errors.
However, this is accompanied by large reductions in substitution and insertion error rates, resulting in overall WER improvements and 
more accurate clusters when boundaries are inferred (45.1\% purity, BayesSeg-MFCC) rather than using fixed syllable boundaries (42\%, SyllableBayesClust), with further improvements when not allowing short word candidates (56\%, BayesSegMinDur-MFCC).

\begin{table}[tb]
    \mytable
    \caption{A breakdown of the errors on English2 for the speaker-dependent models in Table~\ref{tbl:bucktsong_sd_wer}. The word boundary detection tolerance is 20~ms. The greedy one-to-one cluster mapping is used for error rate computations.}
    \begin{tabularx}{\linewidth}{@{}l@{\ }cCCCCCCCc@{}}
        \toprule
        & & \multicolumn{3}{c}{Word boundary (\%)} & \multicolumn{4}{c}{Errors (\%)} & Purity \\
        \cmidrule{3-5} \cmidrule(l){6-9} \cmidrule(l){10-10}
        Model & Embeds. & Prec. & Rec. & $F$  & Sub. & Del. & Ins. & WER & Avg.~(\%) \\
        \midrule
        SyllableBayesClust & MFCC & 27.7 & 28.9 & 28.3 & 63.8 & 13.6 & 16.7 & 94.1 & 42.0 \\
        BayesSeg & MFCC & 29.3 & 26.3 & 27.7 & 59.3 & 18.3 & 11.2 & 88.8 & 45.1\\
        BayesSegMinDur & MFCC & 31.5 & 12.4 & 17.8 & 38.3 & 43.2 & 1.3 & 82.8 & 56.0 \\
        BayesSeg & cAE & 29.1 & 22.8 & 25.6 & 55.7 & 24.3 & 9.3 & 89.3 & 43.9 \\
        BayesSegMinDur & cAE & 30.9 & 10.0 & 15.1 & 35.4 & 47.7 & 1.0 & 84.1 & 55.5 \\
        \bottomrule
    \end{tabularx}
    \label{tbl:bucktsong_sd_english2}
\end{table}

\begin{table*}[tb]
    \mytable
    \caption{A breakdown of the errors on Xitsonga for the speaker-dependent models in Table~\ref{tbl:bucktsong_sd_wer}. The word boundary detection tolerance is 20~ms. The greedy one-to-one cluster mapping is used for error rate computations.}
    \begin{tabularx}{\linewidth}{@{}l@{\ }cCCCCCCCc@{}}
        \toprule
        & & \multicolumn{3}{c}{Word boundary (\%)} & \multicolumn{4}{c}{Errors (\%)} & Purity \\
        \cmidrule{3-5} \cmidrule(l){6-9} \cmidrule(l){10-10}
        Model & Embeds. & Prec. & Rec. & $F$  & Sub. & Del. & Ins. & WER & Avg.~(\%) \\
        \midrule
        SyllableBayesClust & MFCC & 12.4 & 24.8 & 16.5 & 55.8 & 2.1 & 82.4 & 140.3 & 33.1 \\
        BayesSeg & MFCC & 12.4 & 20.3 & 15.4 & 53.5 & 6.0 & 56.6 & 116.2 & 36.8 \\
        BayesSegMinDur & MFCC & 11.8 & 10.8 & 11.3 & 43.2 & 21.2 & 14.5 & 78.9 & 50.1 \\
        BayesSeg & cAE & 12.4 & 18.3 & 14.8 & 50.2 & 9.7 & 47.9 & 107.9 & 40.0\\
        BayesSegMinDur & cAE & 11.5 & 8.9 & 10.0 & 38.3 & 27.9 & 9.7 & 75.9 & 63.7\\
        \bottomrule
    \end{tabularx}
    \label{tbl:bucktsong_sd_tsonga}
\end{table*}

In contrast to English2, Table~\ref{tbl:bucktsong_sd_tsonga} shows that on Xitsonga, SyllableBayesClust heavily over-segments causing a large number of insertion errors.
This is not surprising since every syllable is treated as a word, while most of the true Xitsonga words are multisyllabic.
At the cost of more deletions and poorer word boundary detection, BayesSeg-MFCC and BayesSegMinDur-MFCC systematically reduces substitution and insertion errors, again resulting in better overall WER and average cluster purity.
Where the cAE-based
models on English2 performed more-or-less on par with their MFCC counterparts, on Xitsonga the cAE embeddings yield large improvements on some metrics: by switching to cAE embeddings, the WER of BayesSeg improves by 8.3\% absolute, while average cluster purity is 13.6\% better for BayesSegMinDur.

\subsubsection{Speaker-independent models}

Table~\ref{tbl:bucktsong_si_wer} gives the performance of different speaker-independent models.
Compared to the speaker-dependent results of Table~\ref{tbl:bucktsong_sd_wer}, performance is worse for all models and datasets.
As in the speaker-dependent case, BayesSegMinDur is the best performing MFCC system, followed by BayesSeg, and SyllableBayesClust performs worst.
In the speaker-dependent experiments, some MFCC-based models slightly outperformed their cAE counterparts.
Here, however, the WERs of cAE models are identical or improved in all cases;
for Xitsonga in particular, improvements are obtained by using cAE features in both BayesSeg (improvement of 26.3\% absolute in WER) and BayesSegMinDur (7.4\%).
The cAE-based BayesSegMinDur model is the only speaker-independent Xitsonga model with a WER less than 100\%.
Again, by allowing more than one cluster to be mapped to the same true
word type, WER$_\text{m}$ scores are lower than WER.
Although the cAE-based models do not result in better scores on the English data, the cAE features 
yield large
improvements on Xitsonga. Word boundary scores and substitution,
deletion and insertion errors (not shown) follow a similar pattern to
that of the speaker-dependent models.


\begin{table*}[tbp]
    \mytable
    \caption{Performance on the three datasets for speaker-independent models.}
    \begin{tabularx}{\linewidth}{@{}l@{\ }c@{\ \ }CCCCCC@{}}
        \toprule
        & & \multicolumn{3}{c}{One-to-one WER (\%)} & \multicolumn{3}{c}{Many-to-one WER$_\text{m}$ (\%)} \\
        \cmidrule{3-5} \cmidrule(l){6-8}
        Model & Embeds. & English1 & English2 & Xitsonga & English1 & English2 & Xitsonga \\
        \midrule
        SyllableBayesClust & MFCC & 105.1 & 106.5 & 167.2 & 86.4 & 89.6 & 149.2 \\
        BayesSeg & MFCC & 101.7 & 102.1 & 148.3 & 83.4 & 85.6 & 131.3 \\
        BayesSegMinDur & MFCC & 93.9 & 93.7 & 102.4 & 81.4 & 82.0 & 89.8 \\
        BayesSeg       & cAE & 99.0 & 99.9 & 122.0 & 82.6 & 85.4 & 104.7 \\
        BayesSegMinDur & cAE & 94.0 & 93.7 & 95.0 & 82.4 & 83.3 & 81.1 \\
        \bottomrule
    \end{tabularx}
    \label{tbl:bucktsong_si_wer}
\end{table*}

\begin{table}[tbp]
    \mytable
    \caption{Average speaker-independent cluster (clust.), speaker (spk.), and gender (gndr) purity for BayesSegMinDur on the three datasets.}
    \begin{tabularx}{\linewidth}{@{}cCCCCCCCCC@{}}
        \toprule
        & \multicolumn{3}{c}{English1 (\%)} & \multicolumn{3}{c}{English2 (\%)} & \multicolumn{3}{c}{Xitsonga (\%)} \\
        \cmidrule{2-4} \cmidrule(l){5-7} \cmidrule(l){8-10}
        Embeds. & Clust. & Spk. & Gndr & Clust. & Spk. & Gndr & Clust. & Spk. & Gndr \\
        \midrule
        MFCC & 30.3 & 56.7 & 86.8 & 29.9 & 55.9 & 87.6 & 24.5 & 43.1 & 87.1 \\
        cAE & 31.5 & 37.9 & 77.0 & 30.0 & 35.7 & 73.8 & 33.1 & 29.3 & 76.6 \\
        \bottomrule
    \end{tabularx}
    \label{tbl:bucktsong_si_purity}
\end{table}

To better illustrate the benefits of 
unsupervised representation learning, Table~\ref{tbl:bucktsong_si_purity} shows general purity measures 
for the speaker-independent MFCC- and cAE-based BayesSegMinDur models.
Average cluster purity is as defined before.
Average speaker purity is similarly defined, but instead of considering the mapped ground truth label of a segmented token, it considers the speaker who produced it: speaker purity is 100\% if every cluster contains tokens from a single speaker, while it is $1/12 = 8.3\%$ if all clusters are completely speaker balanced 
for the English sets and $1/24 = 4.2\%$ for Xitsonga.
Average gender purity is similarly defined: it is 100\% if every cluster contains tokens from a single gender, while $1/2 = 50\%$ indicates a perfectly gender-balanced cluster.
Ideally, a speaker-independent system should have high cluster purity and low speaker and gender purities.
Table~\ref{tbl:bucktsong_si_purity} indicates that for all three datasets, cAE-based embeddings are less speaker and gender discriminative, and have higher or similar cluster purity compared to the MFCC-based embeddings.

\begin{figure}[!p]
    \centering
    \begin{minipage}[c]{0.918\linewidth}
        \centering


\includegraphics[scale=0.625]{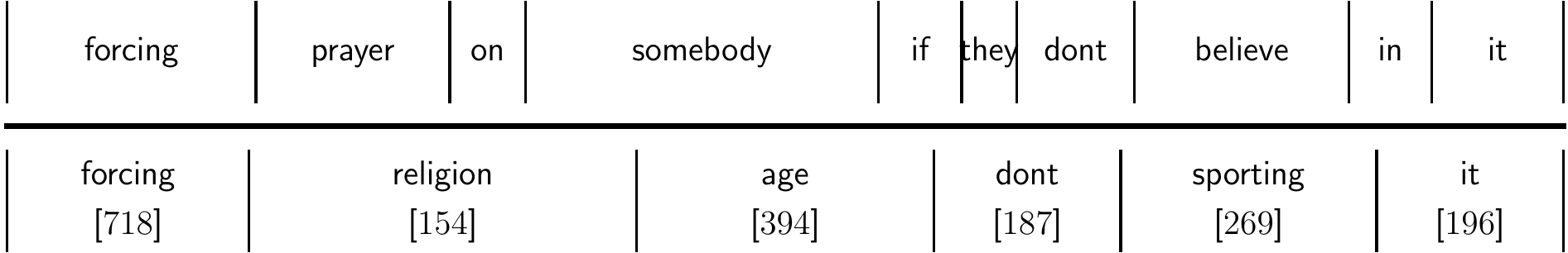}

\vspace{10pt}
\includegraphics[scale=0.625]{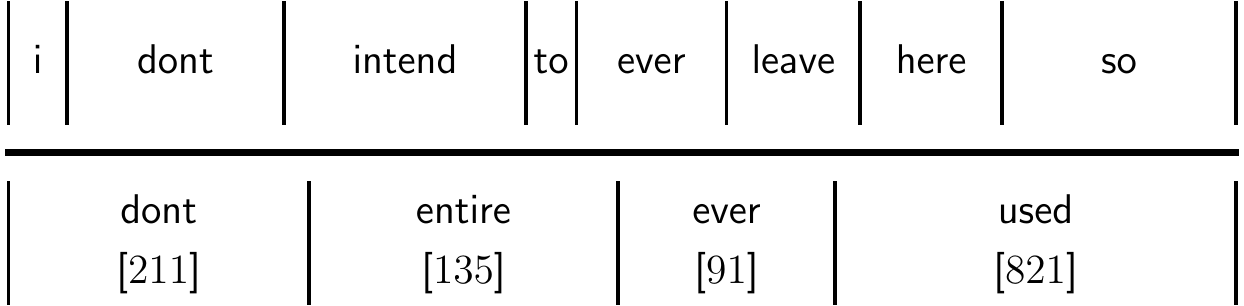}

\vspace{10pt}
\includegraphics[scale=0.625]{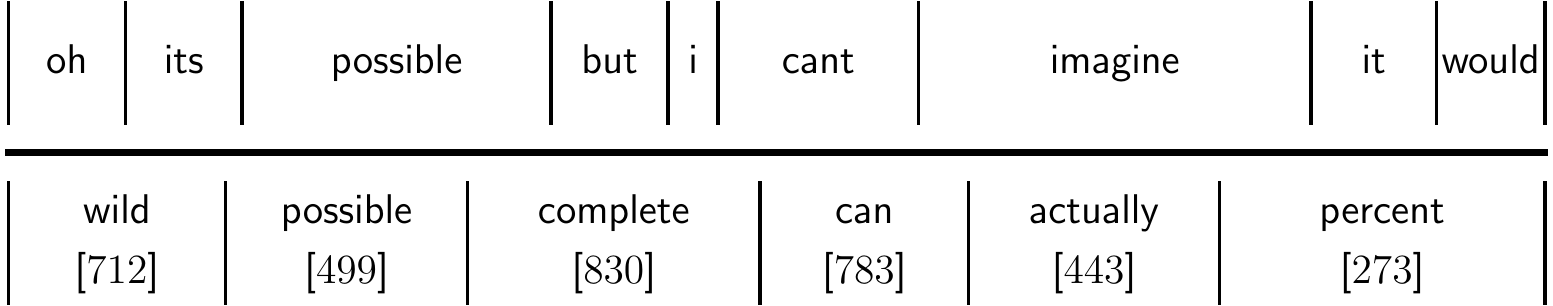}

\vspace{10pt}
\includegraphics[scale=0.625]{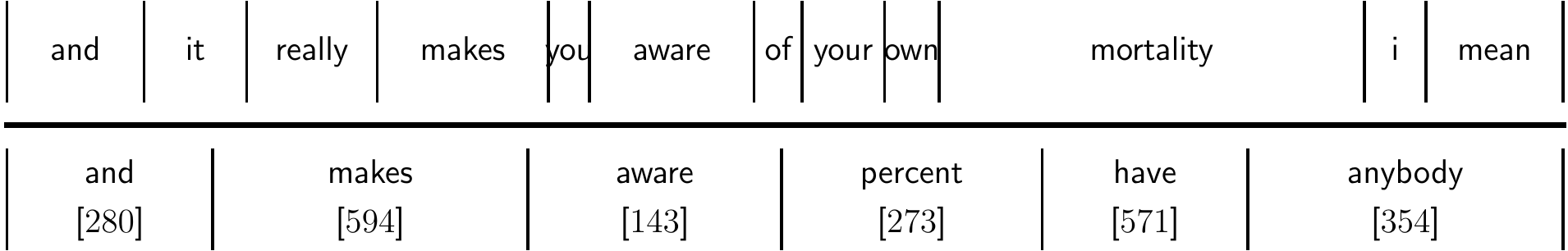}
    \end{minipage}
    \caption{Randomly selected example output of the speaker-dependent BayesSegMinDur models for speaker S01 from English2 and. On top of each horizontal line (representing time), ground truth alignments are given. Discovered patterns are shown below the line: in brackets the cluster IDs, with the many-to-one-mapped labels above the clusters IDs.}
    \label{fig:example_output_buckeye}
\end{figure}

\begin{figure}[!p]
    \centering
    \begin{minipage}[c]{0.918\linewidth}
        \centering
\includegraphics[scale=0.625]{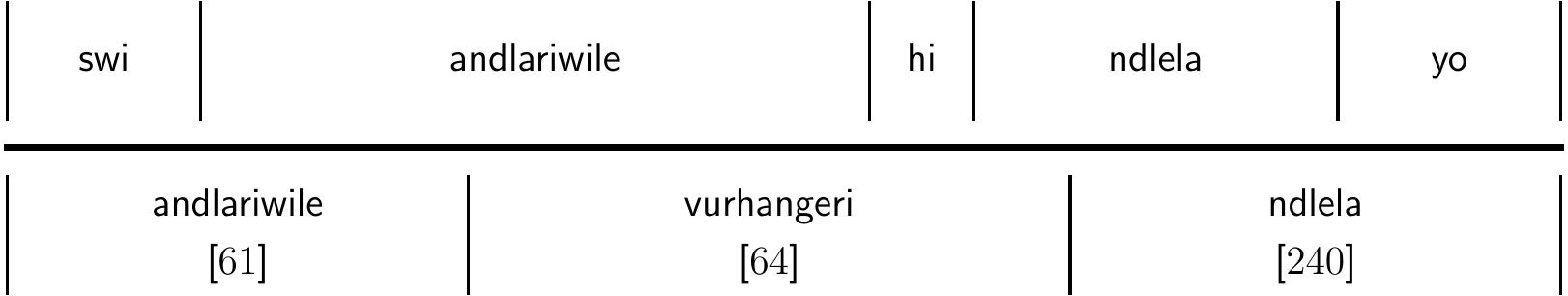}

\vspace{10pt}
\includegraphics[scale=0.625]{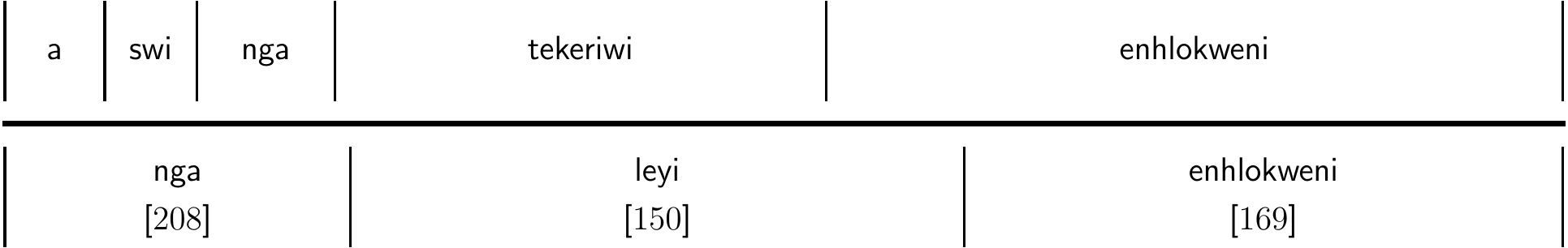}

\vspace{10pt}
\includegraphics[scale=0.625]{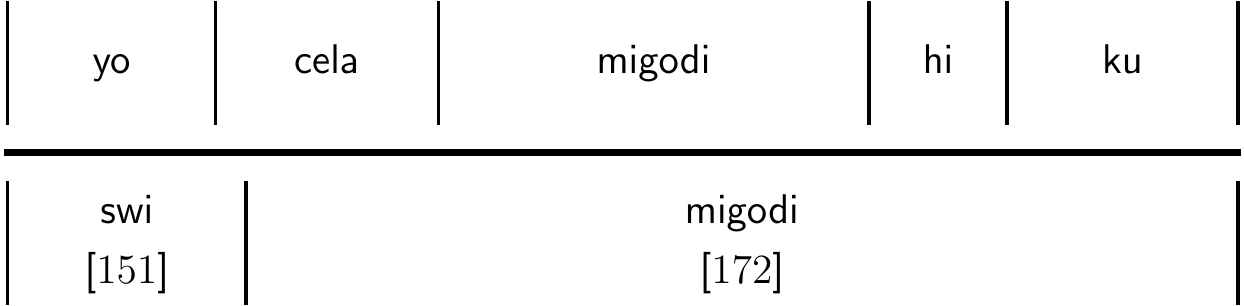}


\vspace{10pt}
\includegraphics[scale=0.625]{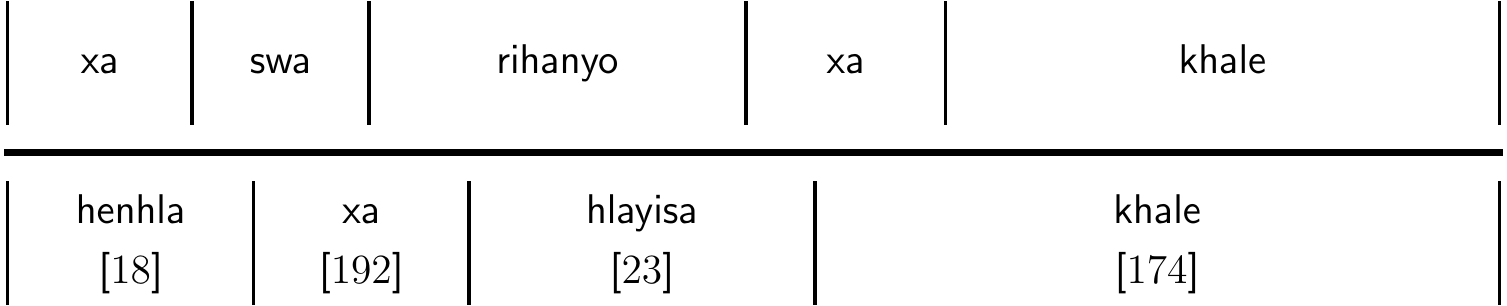}

%
%
%
%
    \end{minipage} 
%
    \caption{Randomly selected example output of the speaker-dependent BayesSegMinDur models for speaker 001m from Xitsonga. On top of each horizontal line (representing time), ground truth alignments are given. Discovered patterns are shown below the line: in brackets the cluster IDs, with the many-to-one-mapped labels above the cluster IDs.}
    \label{fig:example_output_tsonga}
\end{figure}

\subsubsection{Qualitative analysis and summary}

Qualitative analysis involved concatenating and listening to the audio from the tokens in some of the biggest clusters of the best speaker-dependent and -independent models.
Apart from the trends mentioned already, others also became immediately apparent.
Despite the low average cluster purity ranging from 30\% to 60\% in the analyses above, we found that most of the clusters are acoustically very pure: often tokens correspond to the same syllable or partial word, but occur within different ground truth words. For example, a cluster with the word `day' had the corresponding portions from `daycare' and `Tuesday'.
These are marked as errors for cluster purity and WER calculations.
In the next section, we use NED as metric, which does not penalize such partial word matches.
The biggest clusters often correspond to filler-words.
As an example, speaker S38 from English1 had several clusters corresponding to `yeah' and `you know'.  But the BayesSegMinDur-MFCC model applied to S38 also discovered pure clusters corresponding to `different', `people' and `five'.
For the speaker-independent BayesSegMinDur-cAE system, the biggest clusters consisted of instances of `um', `uh', `oh', `so' and `yeah'.
Randomly selected example output from English and Xistonga speaker-dependent BayesSegMinDur systems (Table~\ref{tbl:bucktsong_sd_wer}) are shown respectively in Figures~\ref{fig:example_output_buckeye} and~\ref{fig:example_output_tsonga}, together with the ground truth forced alignments and the many-to-one mapped labels of each cluster.

In summary, although under-segmentation occurs in the BayesSeg and BayesSegMinDur models,
these models yield more accurate clusters and thereby improve overall purity and WER. This highlights the benefit of including sensible bottom-level constraints within a full-coverage segmentation system.
In most cases, cAE embeddings either yield similar or improved performance compared to MFCCs.
In particular in the speaker-independent case, cAE-based models discover clusters that are more speaker- and gender-independent.
This illustrates the benefit of incorporating weak top-down supervision for unsupervised representation learning within a zero-resource system.

\subsection{Results: Comparison to other systems}
\label{sec:bucktsong_comparison}

We now compare our approach to others using
the evaluation framework provided as part of the ZRS challenge~\citep{versteegh+etal_interspeech15}.
We compare to three systems:

\myemph{ZRSBaselineUTD} is the UTD system provided as official baseline in the challenge, as described in Section~\ref{sec:bucktsong_utd}~\citep{versteegh+etal_interspeech15}. 
It is based on the original system of~\cite{jansen+vandurme_asru11}.

\myemph{UTDGraphCC} is the best UTD system of~\cite{lyzinski+etal_interspeech15}, using a connected component graph clustering algorithm to group discovered segments (Section~\ref{sec:bucktsong_utd}).

\myemph{SyllableSegOsc$^{\text{+}}$} uses oscillator-based syllabification followed by speaker-dependent clustering and word discovery, as described in Sections~\ref{sec:background_fullcoverage} and~\ref{sec:bucktsong_utd}~\citep{rasanen+etal_interspeech15}.
We add the superscript + since, after publication of~\citep{rasanen+etal_interspeech15}, the authors further refined their syllable boundary detection method~\citep{rasanen+etal_submission16}. 
We use this updated version for presegmentation in our system. The authors 
kindly regenerated their full ZRS results using the updated method for comparison here.
The original results are included in Appendix~\ref{appen:results_zrs} (denoted without the superscript).

For our approach, we focus on systems that performed best on English1 in the previous section: for the speaker-dependent setting we use the MFCC-based BayesSegMinDur system, while for the speaker-independent setting we use the cAE-based BayesSegMinDur model.
The performance of all our system variants using all of the ZRS metrics are given in Appendix~\ref{appen:results_zrs}.

\begin{figure}[!b]
    \centering
    \includegraphics[scale=0.825]{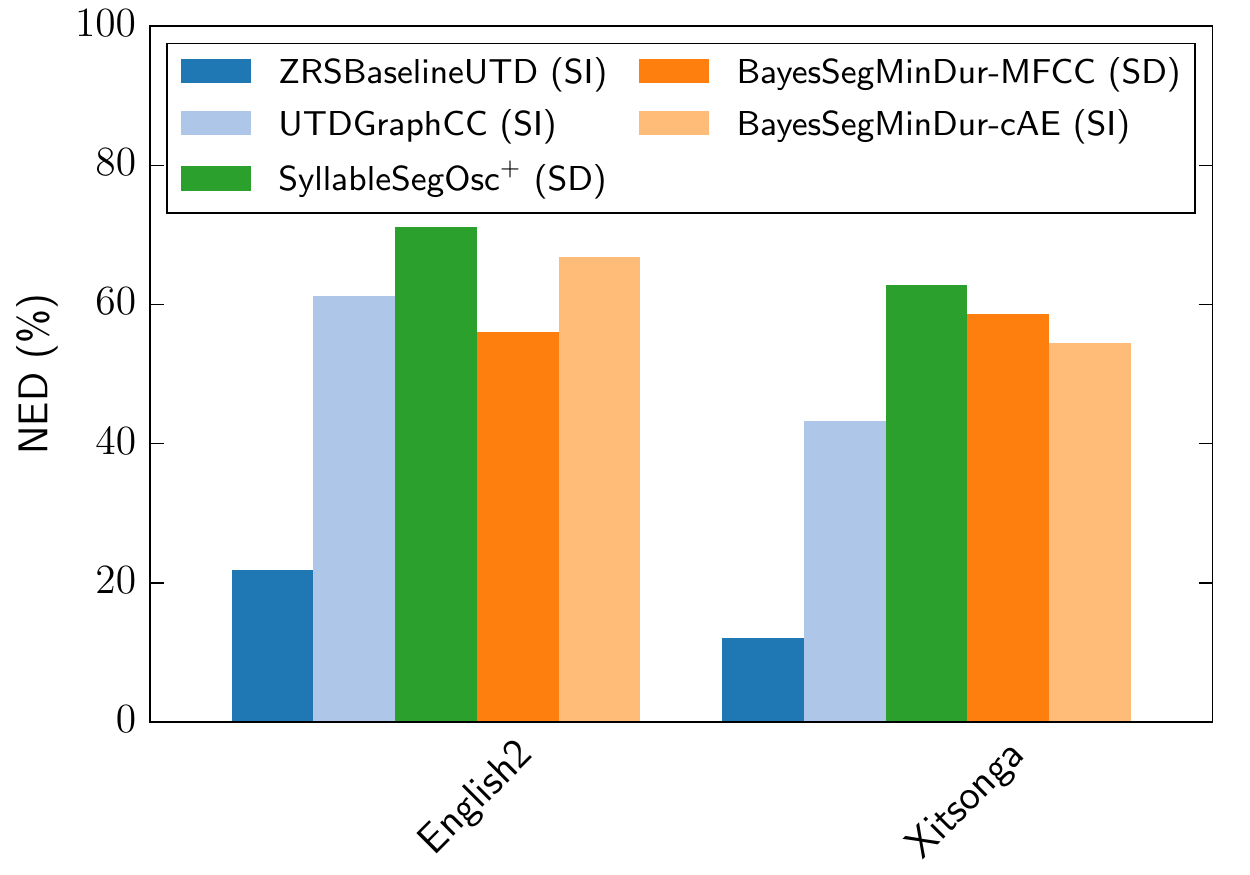}
    \caption{Normalized edit distance (NED) on English2 and Xitsonga. Lower NED is better. Scores are only computed on the analyzed portion of data (so the lower-coverage UTD systems have an advantage). SD/SI indicates that a system is speaker-dependent/speaker-independent.}
    \label{fig:ned_zrs}
\end{figure}

\begin{figure}[tbp]
    \centering
    \includegraphics[scale=0.825]{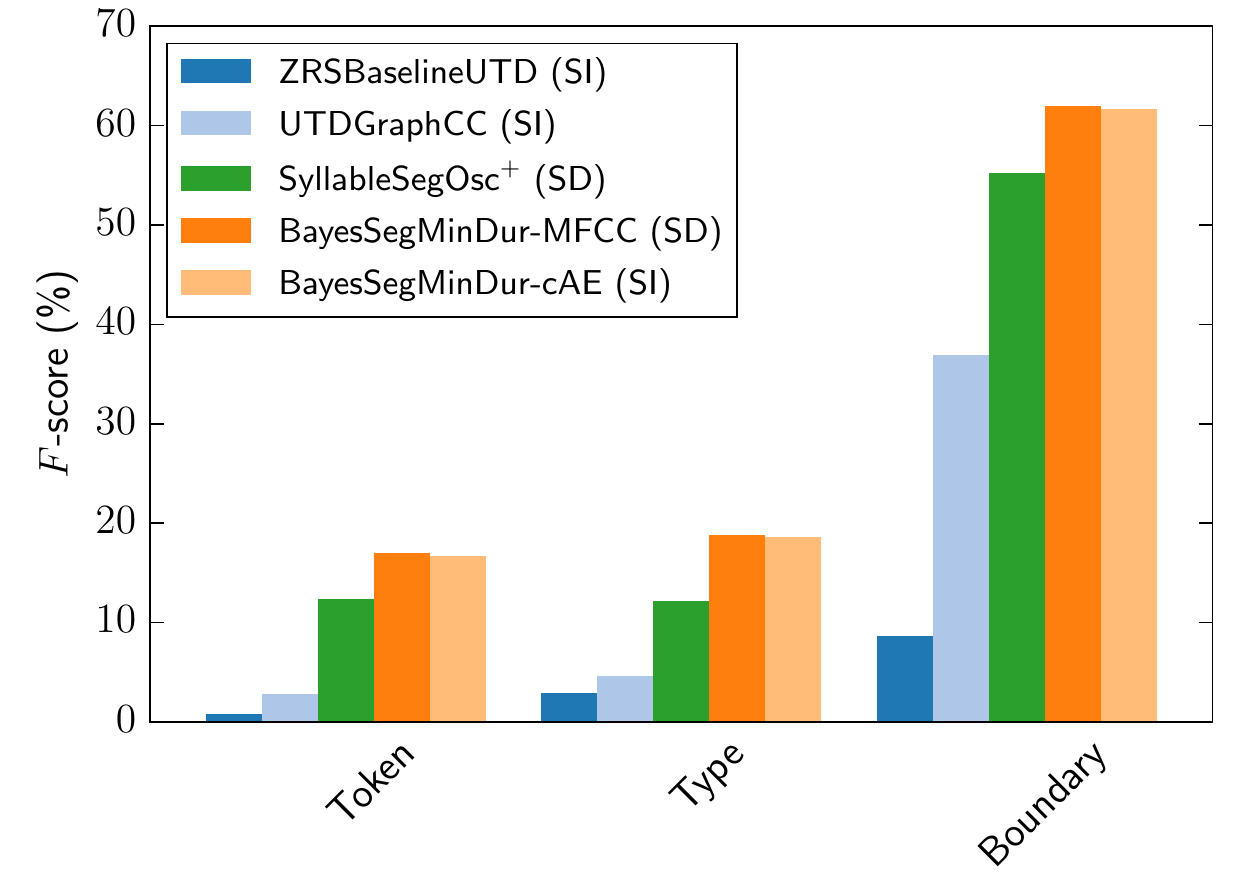}
    \caption{Word token, type and boundary $F$-scores on English2.
    SD/SI indicates that a system is speaker-dependent/speaker-independent.
    The word boundary detection tolerance is 30~ms or 50\% of a phoneme.
    }
    \label{fig:english_zrs}
\end{figure}

\begin{figure}[tbp]
    \centering
    \includegraphics[scale=0.825]{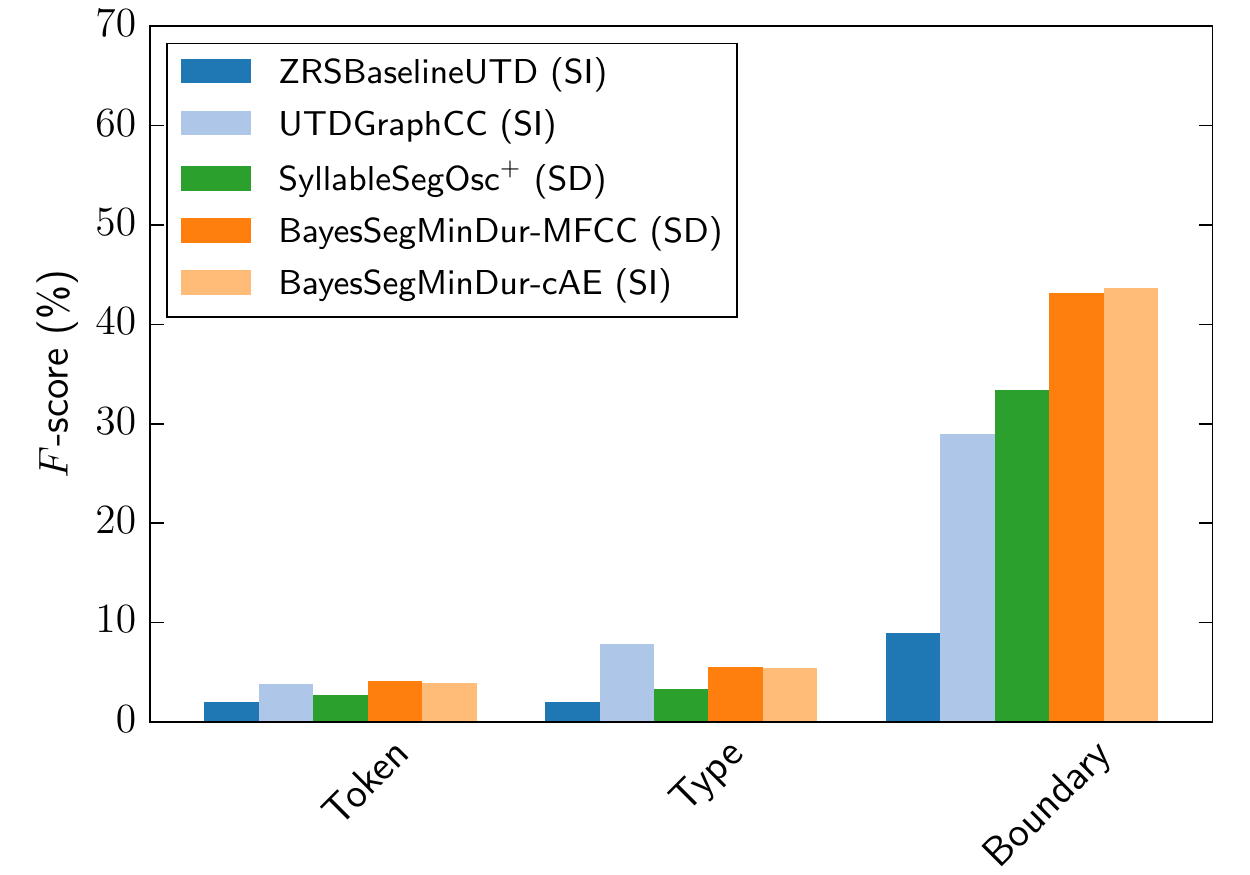}
    \caption{Word token, type and boundary $F$-scores on Xitsonga.
    SD/SI indicates that a system is speaker-dependent/speaker-independent.
    The word boundary detection tolerance is 30~ms or 50\% of a phoneme.}
    \label{fig:xitsonga_zrs}
\end{figure}

Figure~\ref{fig:ned_zrs} shows the NED scores of the different systems on English2 and Xitsonga.
Lower NED scores are better.
ZRSBaselineUTD yields the best NED on both languages, with UTDGraphCC also performing well.
UTD systems like these explicitly aim to discover high-precision clusters of isolated segments, but do not cover all the data.
They are therefore tailored to NED, which only evaluates the patterns discovered by the method and does not evaluate recall on the rest of the data.
In contrast, SyllableSegOsc$^{\text{+}}$ and our own systems perform full-coverage segmentation.
Of these, our systems achieve better NED than SyllableSegOsc$^{\text{+}}$ on both languages, indicating that the discovered clusters in our approach are more consistent.
Even when running the system in a speaker-independent setting (BayesSegMinDur-cAE in the figure), our approach outperforms the speaker-dependent SyllableSegOsc$^{\text{+}}$.

Figures~\ref{fig:english_zrs} and~\ref{fig:xitsonga_zrs} show the token, type and boundary $F$-scores on the two languages.
Apart from word type $F$-score on Xitsonga, our models outperform all other approaches.
The UTD systems struggle on these metrics since the $F$-scores are based on precision and recall over the entire input.
The full-coverage SyllableSegOsc$^{\text{+}}$ is therefore our strongest competitor in most cases.
The prediction of word candidates from reoccurring cluster sequences in SyllableSegOsc$^{\text{+}}$ is done greedily and bottom-up, without regard to other word mappings in an utterance.
In contrast, BayesSegMinDur samples word boundaries and cluster assignments together by taking a whole utterance into account; it imposes a consistent top-down segmentation, while simultaneously adhering to bottom-up syllable boundary detection and minimum duration constraints.
The result is a more accurate segmentation of the data.
Note that in BayesSeg it is easy to incorporate additional bottom-up constraints (such as a minimum duration) and these are considered jointly with segmentation.
In contrast, such a minimum duration constraint would require additional heuristics in the pure bottom-up approach of~\cite{rasanen+etal_interspeech15}.

The results in Figures~\ref{fig:english_zrs} and~\ref{fig:xitsonga_zrs} also indicate that our speaker-independent system performs on par with the speaker-dependent system on these metrics; despite less accurate clusters (in terms of purity, WER and NED), the speaker-independent models still yields an accurate segmentation of the data, outperforming both speaker-independent UTD baselines and the speaker-dependent SyllableSegOsc$^{\text{+}}$.

We conclude that by hypothesizing word boundaries consistently over an utterance rather than taking these decisions in isolation,
our approach yields more accurate clusters (NED) that correspond better to true words (word type $F$-score) than the bottom-up full-coverage syllable-based approach of~\cite{rasanen+etal_interspeech15}.
It also segments the data more accurately (word token and boundary $F$-scores), even when applying the model to data from multiple speakers.
However, despite the benefits of our model, the algorithm of~\cite{rasanen+etal_interspeech15} is much simpler in terms of computational complexity and implementation.
Compared to UTD systems which aim to find high-quality reoccurring patterns but do not cover all the data, the items in our clusters have a poorer match to each other (NED), but correspond better to true words on the English data (word type $F$-score). On both languages, our full-coverage method also segments the data better into word-like units (word boundary and token $F$-scores) than the UTD systems.

\subsection{Results: Towards bigram modelling}
\label{sec:buckstong_bigram_exp}

Up to this point we have been using the unigram segmental Bayesian model (first introduced in Section~\ref{sec:tidigits_segmental_bayesian_model}).
Here we present a set of experiments using a bigram model 
to sample the cluster assignments of hypothesized word segments, as outlined in Section~\ref{sec:bucktsong_bisampling_assignments}.
All the experiments presented here were carried out on a single male speaker (S38) from English1 using MFCC embeddings and the bigram version of BayesSegMinDur.

\begin{figure}[tbp]
    \centering
    \includegraphics[scale=0.85]{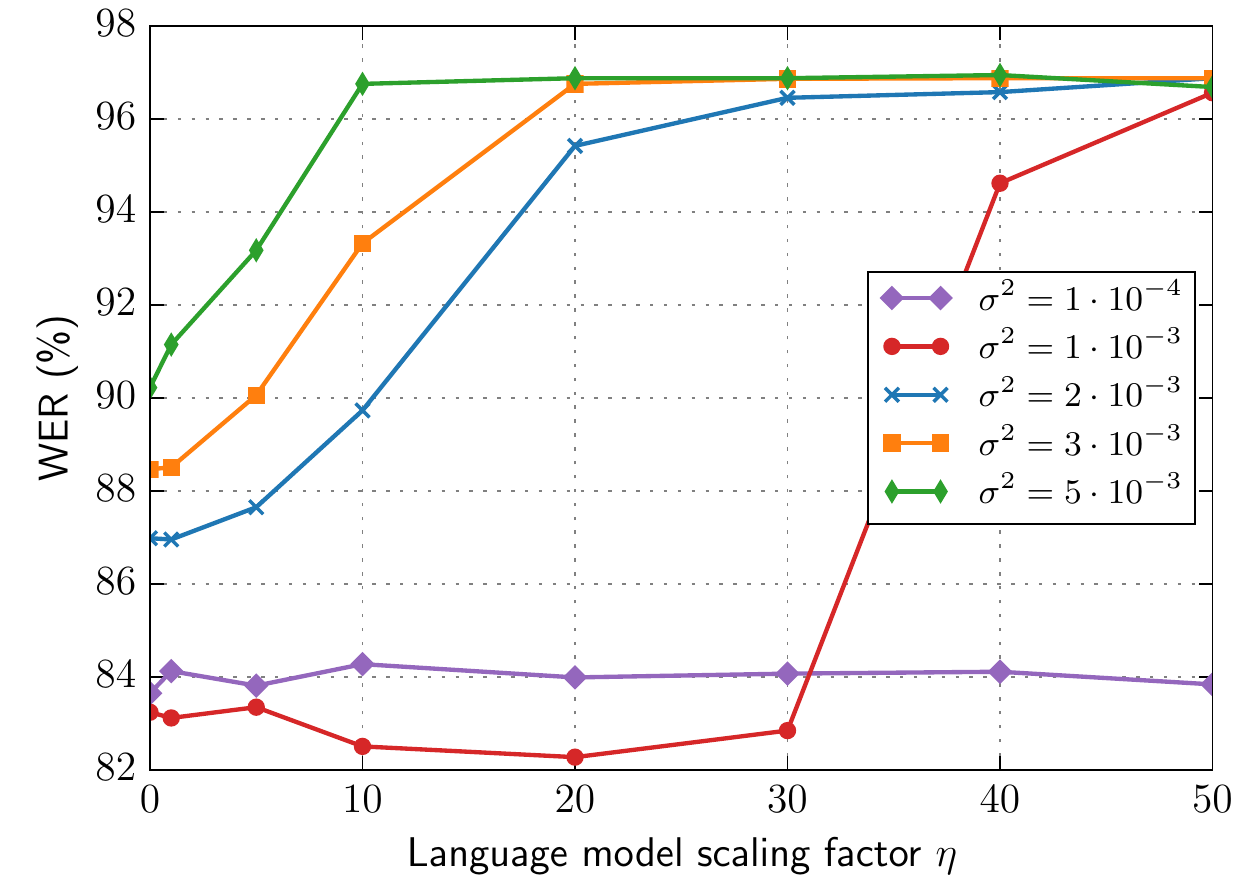}
    \caption{Unsupervised one-to-one WER of bigram BayesSegMinDur-MFCC models applied to speech from a single speaker, as the covariance $\sigma^2$ and language model scaling factor $\eta$ are varied.}
    \label{fig:bucktsong_wer_lms}
\end{figure}

\begin{figure}[tbp]
    \centering
    \includegraphics[scale=0.85]{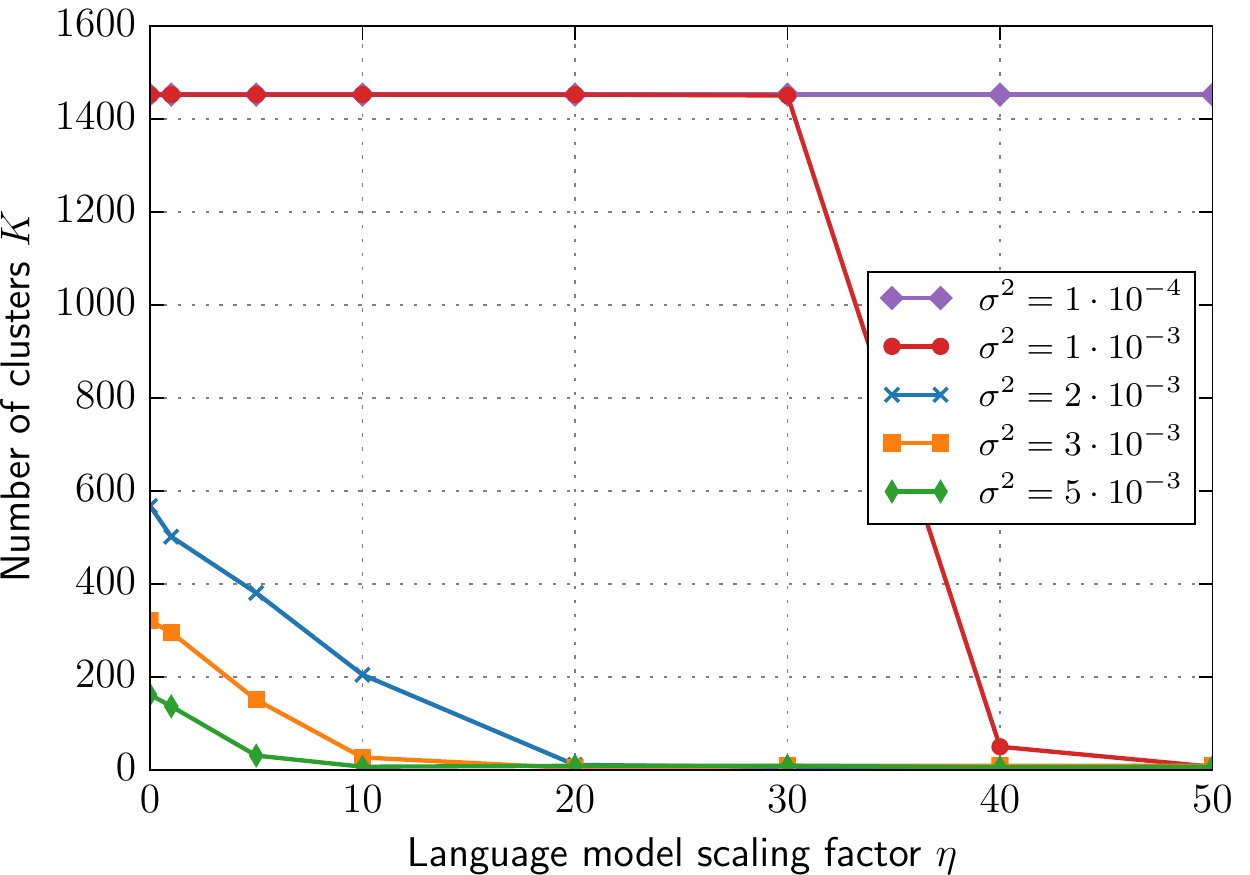}
    \caption{The number of Gaussian components $K$ of bigram BayesSegMinDur-MFCC models applied to speech from a single speaker, as the covariance $\sigma^2$ and language model scaling factor $\eta$ are varied.
    The maximum number of components is $K = 1453$.}
    \label{fig:bucktsong_n_clusters_lms}
\end{figure}

We first consider this model across a range of language model scaling factors $\eta$ using the same covariance parameter $\sigma^2 = 1\cdot10^{-3}$ used throughout for the MFCC systems so far.
As described in Section~\ref{sec:bucktsong_bisampling_assignments}, $\eta$ controls the relative importance of the language and acoustic modelling terms in~\eqref{eq:bucktsong_bicollapsed2}, with larger $\eta$ assigning more weight to the language model.
The WERs for this range of models are shown with the red circle-dotted line in Figure~\ref{fig:bucktsong_wer_lms}.
For comparison, the unigram model achieved $83.3\%$ WER on this speaker (with $\eta = 1$).
The red-circle line in the figure shows that the bigram model has a similar WER for $\eta$ smaller than 30.
At $\eta=40$, however, WER increases dramatically to around 95\%.
The corresponding red circle-dotted line in Figure~\ref{fig:bucktsong_n_clusters_lms} indicates that this drop in WER coincides with a large drop in the number of clusters used by the model: up to $\eta=30$, the model uses the maximum of $K = 1453$ clusters, but at $\eta=40$, this drops to only 50.
Qualitative listening reveals that these 50 clusters are very noisy, having an average cluster purity of only 9.8\%.

As the scaling factor $\eta$ is increased, it is natural that fewer clusters would be preferred since the language modelling term $P(z_i = k|\vec{z}_{\backslash i} ; \vec{\gamma})$ in~\eqref{eq:bucktsong_bicollapsed2} would be higher for those clusters that are occupied.
The analysis for the $\sigma^2 = 1\cdot10^{-3}$ model in Figure~\ref{fig:bucktsong_wer_lms} indicates, however, that the point at which $P(z_i = k|\vec{z}_{\backslash i} ; \vec{\gamma})$ has an effect over the acoustic term $p(\vec{x}_i | \mathcal{X}_{k \backslash i}; \vec{\beta})$ in~\eqref{eq:bucktsong_bicollapsed2} is sudden and leads to dramatic consequences in terms of the number of clusters and accuracy.

We attribute this to the very peaked nature of the acoustic score $p(\vec{x}_i | \mathcal{X}_{k \backslash i}; \vec{\beta})$, which we hope to illustrate in the rest of this section.
If every potential word segment has a particular component $k$ which it prefers acoustically almost certainly over all others---i.e.\ there is a single $k$ for which the PDF $p(\vec{x}_i | \mathcal{X}_{k \backslash i}; \vec{\beta})$ is several orders higher than all others---then when the bigram language modelling term $P(z_i = k|\vec{z}_{\backslash i} ; \vec{\gamma})$ disagrees with the acoustic assignment, the model is required to move tokens to clusters where they are significantly less likely.
If the scaling factor $\eta$ is set high enough, this results in the sudden creation of a small number of large clusters with very poor purity.

Since we explicitly set the spherical covariance parameter $\sigma^2$ for $p(\vec{x}_i | \mathcal{X}_{k \backslash i}; \vec{\beta})$, we can explicitly control how peaked this posterior predictive distribution is.
We therefore performed an analysis where we varied $\sigma^2$. WERs and the number of clusters for several settings of $\sigma^2$ are illustrated in Figures~\ref{fig:bucktsong_wer_lms} and~\ref{fig:bucktsong_n_clusters_lms}.
As $\sigma^2$ is increased above $1\cdot10^{-3}$, Figure~\ref{fig:bucktsong_n_clusters_lms} shows that fewer clusters are used since the acoustic model now allows more diverse embedded segments to be clustered together.
This, however, is accompanied by a gradual increase in WER in Figure~\ref{fig:bucktsong_wer_lms}, showing that clusters become less accurate.
Although the effect of the bigram language model is less sudden for the higher-$\sigma^2$ models than for $\sigma^2 = 1\cdot10^{-3}$, WERs still gradually increase and the number of clusters decrease as the scaling factor $\eta$ is raised.

\begin{figure}[p]
    \centering
    \includegraphics[scale=0.85]{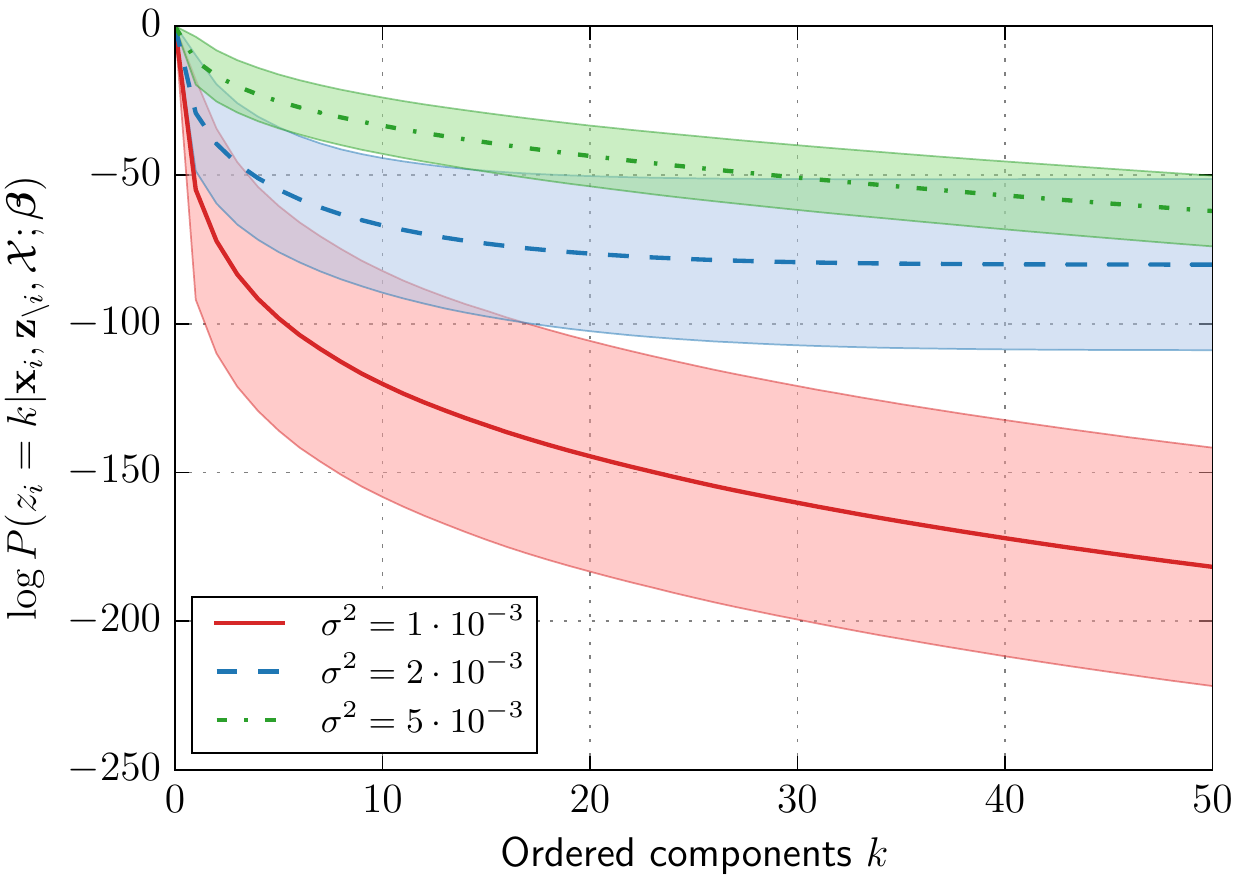}
    \caption{The average log posterior predictives for components ordered from most to least probable for BayesSegMinDur-MFCC applied to speech from a single speaker.  The first $50$ out of $1453$ components are shown. No language model is used: $\eta = 0$. The shaded area indicates a single standard deviation from the mean.}
    \label{fig:bucktsong_post_pred_assignments_3}
\end{figure}

\begin{figure}[p]
    \centering
    \includegraphics[scale=0.85]{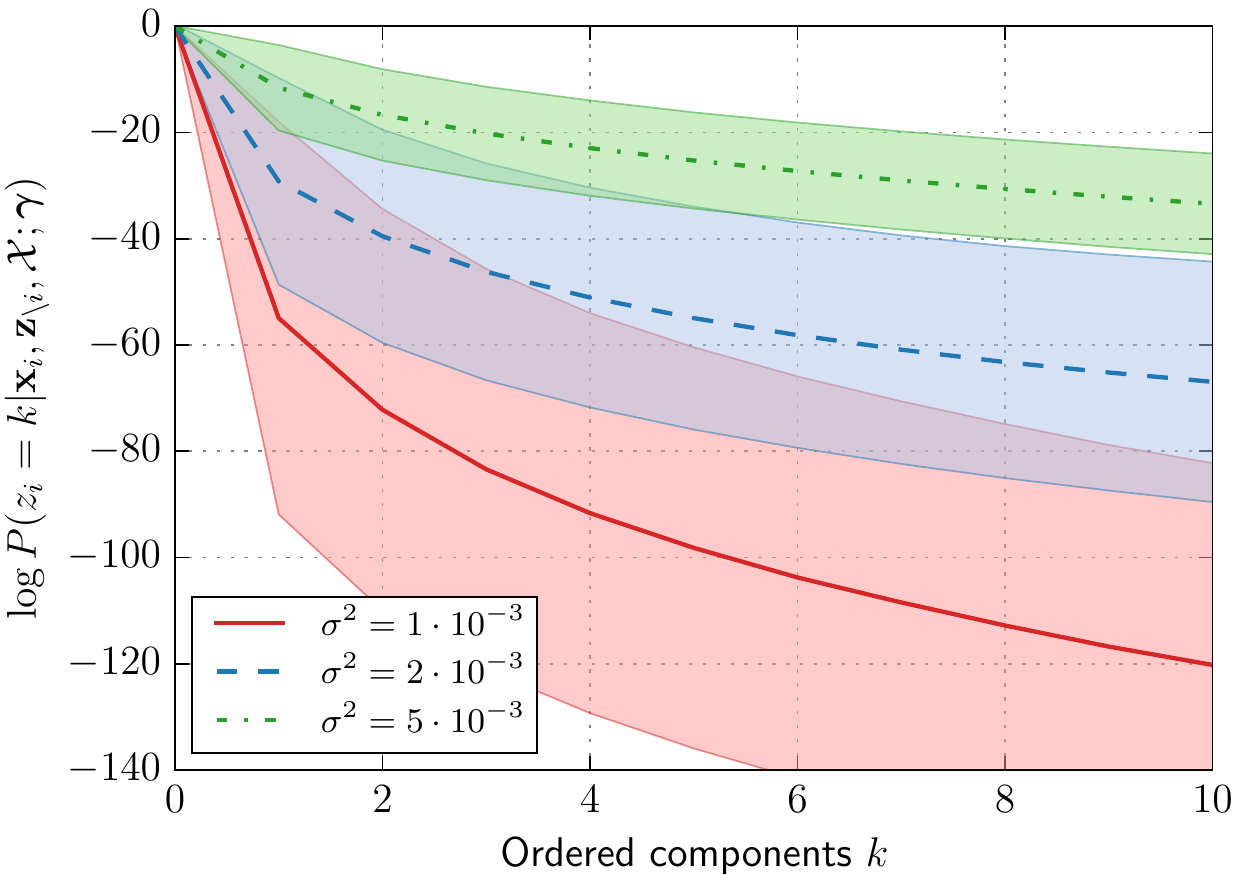}
    \caption{A focused view of Figure~\ref{fig:bucktsong_post_pred_assignments_3} showing only the first ten components.}
    \label{fig:bucktsong_post_pred_assignments_3_zoom}
\end{figure}

To show that the posterior predictive $p(\vec{x}_i | \mathcal{X}_{k \backslash i}; \vec{\beta})$ is indeed peaked around a single value for most embeddings, we performed the following analysis. 
We set $\eta = 0$ to remove the language model and only consider the acoustic term.
For a particular embedded segment $\vec{x}_i$, the component assignment in line~\ref{alg_line:fbgmm_inside_loop} of Algorithm~\ref{alg:gibbs_wordseg} is now sampled according to $P(z_i = k|\vec{z}_{\backslash i}, \mathcal{X} ; a, \vec{\beta} ) \propto p(\vec{x}_i | \mathcal{X}_{k \backslash i}; \vec{\beta})$.  We use vector $\vec{p}$, with elements $p_k = P(z_i = k|\vec{z}_{\backslash i}, \mathcal{X} ; a, \vec{\beta})$, to denote the categorical distribution used to sample the assignment.
We are interested in determining the expected `peakedness' of $\vec{p}$ for an arbitrary embedding.
For all the embeddings considered in the last Gibbs sampling iteration of this model, we therefore calculate $\vec{p}$ and then reorder the elements within $\vec{p}$ from largest to smallest.
The ordered vectors are then averaged over all embeddings, and the result is shown on the $\log$ scale in Figures~\ref{fig:bucktsong_post_pred_assignments_3} and~\ref{fig:bucktsong_post_pred_assignments_3_zoom}.

\begin{figure}[tbp]
    \centering
    \includegraphics[scale=0.85]{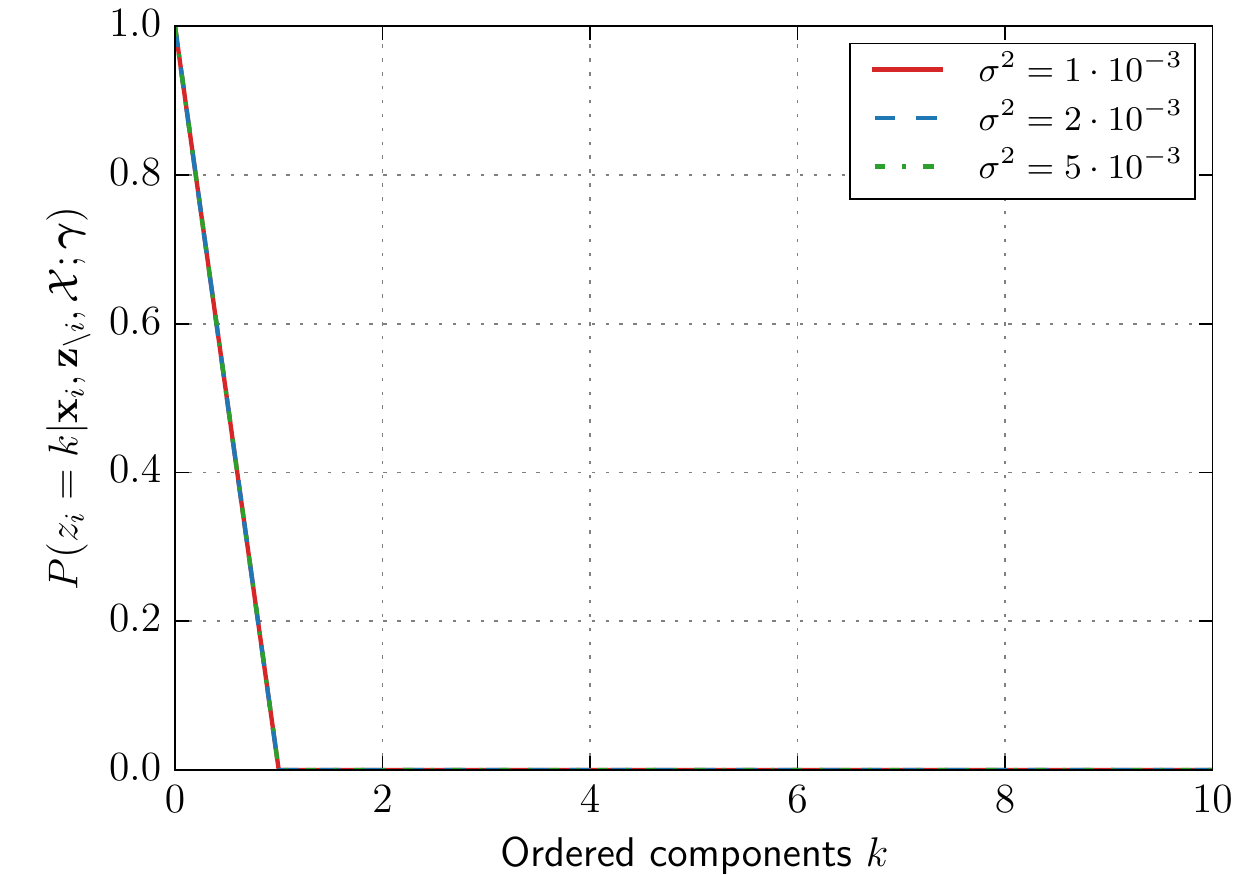}
    \caption{The average posterior predictives for components ordered from most to least probable for BayesSegMinDur-MFCC applied to speech from a single speaker.
    In contrast to Figure~\ref{fig:bucktsong_post_pred_assignments_3_zoom}, the log scale is not used here.
    As in Figure~\ref{fig:bucktsong_post_pred_assignments_3_zoom}, the first $10$ out of $1453$ components are shown. No language model is used: $\eta = 0$.
    }
    \label{fig:bucktsong_post_pred_assignments_3_nonlog_zoom}
\end{figure}

For all three settings of $\sigma^2$ considered, the average of $\log P(z_i = k|\vec{z}_{\backslash i}, \mathcal{X} ; a, \vec{\beta})$ is close to zero for the most likely component.
When focusing in on only the first ten most probable components as in Figure~\ref{fig:bucktsong_post_pred_assignments_3_zoom}, it is evident that even the second-most likely component has a much lower probability on average; this is the case even for the $\sigma^2=5\cdot10^{-3}$ system, which has a much flatter posterior predictive distribution because of the larger covariance.
The peakedness is very clear when considering the same analysis but on the scale at which sampling is actually performed (rather than the log scale): this is shown in Figure~\ref{fig:bucktsong_post_pred_assignments_3_nonlog_zoom} which indicates that in most cases a single component is almost deterministically preferred over all others.

In summary, since the acoustic distribution is so peaked, there is little for the bigram language model to contribute since it would need to force tokens into clusters that are acoustically a much poorer match in cases where it does not agree with the acoustic model.
Flattening the acoustic distribution through the covariance parameter $\sigma^2$ results in much less accurate clusters, which the bigram language model cannot take advantage of either.
A qualitative analysis of the best unigram models reveals that different instances of the same ground truth word type are often split across different clusters.
This happens because their embeddings are different.
If different instances of the same word were truly mapped to similar areas in the embedding space, but still split into different clusters, then these clusters would be confusable and result in a flatter acoustic distribution.
We therefore speculate that a more accuracte embedding function $f_e$ would be necessary to take advantage of word-word dependencies in the data.

Here we considered a limited implementation of the model using bigram modelling only in cluster assignment and still using a unigram assumption for segmentation (Section~\ref{sec:bucktsong_bisampling_assignments}).
As a result of the analysis above, the full bigram model was not implemented.
However, a complete description of the full model is given in Appendix~\ref{appen:derivations_bigramseg}, with a discussion of possible approximations and other potential pitfalls.

\section{Summary and conclusion}

This chapter presented a segmental Bayesian model which segments and clusters conversational speech audio---the first full-coverage zero-resource system to be evaluated
on multi-speaker large-vocabulary data.
The system limits word boundary positions by using a bottom-up presegmentation method to detect syllable-like units, and relies on a downsampling approach to represent word candidates as fixed-dimensional acoustic word embeddings.

Our speaker-dependent systems achieve WERs of around 84\% on English and 76\% on Xitsonga data, outperforming a purely bottom-up method that treats each syllable as a word candidate.
Despite much worse speaker-independent performance, here we achieve improvements by incorporating frame-level features from the correspondence autoencoder (cAE), trained using weak top-down constraints (Chapter~\ref{chap:cae}).
This results in clusters that are purer and less speaker- and gender-specific than when using MFCCs.
The cAE incorporates both bottom-up and top-down knowledge, and the results here show that this combination results in an unsupervised representation learning method that extrinsically improves speaker-independence.

We compared our approach to state-of-the-art baselines on both languages.
We found that, although the isolated patterns discovered by term discovery systems are more consistent, the clusters of our full-coverage approach are better matched to true words, measured in terms of word token, type and boundary $F$-scores.
We also found that by proposing a consistent top-down segmentation and clustering over whole utterances while simultaneously taking into account bottom-level constraints, the approach outperforms a purely bottom-up syllable-based full-coverage system on these metrics.

In a final set of experiments, we investigated an extension of the approach which uses a bigram language model  for sampling cluster assignments.
Because of the very peaked nature of the acoustic component of the model, a small set of inaccurate garbage clusters form when the language model is scaled such that it has an effect.
We speculate that a more accurate embedding function would be required in order to obtain smoother acoustic likelihoods over the different clusters.
Both the analysis here and in Chapter~\ref{chap:tidigits} highlighted the reliance of our approach on accurate embeddings; in the next chapter, this aspect is noted as a crucial direction for future work.

\graphicspath{{conclusion/fig/}}

\chapter{Summary and conclusions}
\label{chap:conclusion}

Zero-resource speech processing is essential for allowing speech technology to be developed without transcriptions, pronunciation dictionaries or language modelling text in settings where unlabelled speech audio is the only available resource.
This thesis made contributions in both of the two main focus-areas of zero-resource speech processing: (i)~unsupervised representation learning, and (ii)~segmentation and clustering of unlabelled speech.
The overarching claim of the thesis is that 
both bottom-up and top-down modelling are beneficial for solving these problems.
Below, we first show how this claim is supported by giving a summary of the presented work.
We then describe how the work could be extended and applied in future research, and discuss challenges and research questions for zero-resource speech processing in general.

\section{Main findings}
\label{sec:conclusion_main_findings}

A summary of the main findings of the thesis is given below, showing
that 
a combination of top-down and bottom-up modelling are advantageous for zero-resource speech processing.
As explained in Section~\ref{sec:intro_topdown_bottomup}, \textit{top-down modelling} refers to a process where knowledge of higher-level units are used to gain insight into their lower-level constituents, while \textit{bottom-up modelling} uses knowledge of lower-level features to give rise to more complex higher-level structures.

\subsection{Unsupervised representation learning}

Chapter~\ref{chap:cae} introduced a new model for unsupervised representation learning, where the aim is to use unlabelled speech audio to learn a feature extracting function that is beneficial for discriminating between linguistic units.
We proposed a novel deep neural network (DNN) model, the \textit{correspondence autoencoder} (cAE), which uses noisy top-down supervision from a first-pass unsupervised term discovery (UTD) system.
Concretely, dynamic time warping (DTW) is used to align feature frames from discovered words that are predicted to be of the same type, and these are presented as input-output pairs to the cAE.
The intuition is that the aligned frames should share common aspects from the underlying subword unit from which they originate, and that this should be captured in intermediate layers of the cAE.

We compared features from the cAE (obtained from a middle layer) to several state-of-the-art approaches in an isolated word discrimination task, giving an intrinsic measure of the quality of the produced representations. 
Two of the baselines (a Gaussian mixture model (GMM) and a stacked autoencoder) were trained purely bottom-up directly on the input speech features.
The third baseline used top-down constraints to partition a GMM-based universal background model (UBM).
As already found in~\citep{jansen+etal_icassp13b}, the top-down partitioned GMM-UBM outperformed both bottom-up baselines.
However, the cAE in turn outperformed the partitioned GMM-UBM by 64\% relative, achieving the best reported result on this word discrimination task.

This shows that DNN-based modelling---which has proved so successful in supervised speech recognition---can also lead to major improvements for representation learning in zero-resource settings, specifically by making effective use of top-down knowledge of longer-spanning patterns in the data.
Crucially, however, the cAE does not rely on top-down modelling alone: it was initialized using weights from a bottom-up stacked AE trained on raw speech. 
Although the stacked AE itself did not provide better features than the input acoustic features, 
this bottom-up initialization was essential in order for the cAE to learn useful representations.

\subsection{Unsupervised segmentation and clustering of speech}

In Chapters~\ref{chap:tidigits} and~\ref{chap:bucktsong} we introduced a novel segmental Bayesian model which completely segments and clusters raw speech audio into word-like units---effectively performing unsupervised speech recognition.
Most previous work on full-coverage segmentation and clustering followed an approach of bottom-up subword discovery with subsequent or joint word discovery, working directly on the frame-wise acoustic speech features.
In contrast, our model uses a fixed-dimensional representation of whole segments: any potential word segment of arbitrary length is mapped to a fixed-length vector, its \textit{acoustic word embedding}.
Using this embedding representation, the model jointly segments speech data into word-like segments and then clusters
these segments using a whole-word Bayesian acoustic model in the embedding space.


In Chapter~\ref{chap:tidigits} we applied this approach to a small-vocabulary unsupervised connected-digit recognition task using the English TIDigits corpus.
As a baseline, we used the more traditional HMM-based system of \cite{walter+etal_asru13}.
By mapping the unsupervised decoded output to ground truth transcriptions, unsupervised word error rates (WERs) were reported.
Our approach achieved around 20\% WER, outperforming the HMM-based system by about 10\% absolute.
On this small-vocabulary task, our method was also able to automatically discover an appropriate number of clusters without being specified the vocabulary size beforehand (in contrast to the HMM baseline).

The small-vocabulary model of Chapter~\ref{chap:tidigits} used a reference vector method to obtain the fixed-dimensional acoustic word embeddings (Sections~\ref{sec:background_segmental} and~\ref{sec:tidigits_embeddings}): 
for a target speech segment, a reference vector consists of the DTW alignment cost to every exemplar in a reference set; 
applying dimensionality reduction to the reference vector yields the embedding in the desired fixed-dimensional space.
The intuition is that the content of a speech segment should be characterized well through its similarity to other segments.
In the original work that proposed this embedding method~\citep{levin+etal_asru13}, the reference set consisted of true word segments, which is not available in truly zero-resource settings.
We therefore started with exemplars extracted randomly from the data, ran the segmentation system, and then included discovered terms in the reference set; this was repeated for a number of iterations, each further refining the reference set and improving performance (see Sections~\ref{sec:tidigits_iterating} and~\ref{sec:tidigits_experiments}, Figure~\ref{fig:flow_diagram} and Table~\ref{tbl:exemplars}).
Again, these improvements can be seen as the result of a combination of top-down and bottom-up modelling: initial embedding extraction using the random reference set is a bottom-up process (relying only on the lowest-level acoustic features and not using any knowledge of higher-level units); by then refining the reference set using the terms discovered in an outer discovery loop, the embedding method starts to incorporate top-down knowledge of longer-spanning patterns to obtain better segmental representations.

In Chapter~\ref{chap:bucktsong}, we extended the system of Chapter~\ref{chap:tidigits} in order to handle large-vocabularies.
Apart from an acoustic model with many more components, a number of additional changes were made to deal with computational constraints.
To improve efficiency, we used a bottom-up unsupervised syllable boundary detection method~\citep{rasanen+etal_interspeech15} to eliminate unlikely word boundaries, reducing the number of potential word segments that need to be considered.
We also used a computationally much simpler embedding approach based on downsampling (a purely bottom-up approach).
While only traditional acoustic features were used in Chapter~\ref{chap:tidigits}, here we considered both MFCCs and cAE features as input to the embedding function. 
We evaluated the approach in both speaker-dependent and speaker-independent settings on English and Xitsonga conversational speech datasets, and compared performance to state-of-the-art baselines using a variety of measures.
We showed that by imposing a consistent top-down segmentation while also using bottom-up knowledge from detected syllable boundaries, both single-speaker and multi-speaker versions of our system outperformed a purely bottom-up single-speaker syllable-based approach.
We also showed that the discovered clusters can be made less speaker- and gender-specific by using the cAE to obtain better frame-level features (prior to embedding); this shows that cAE features (which incorporates both top-down and bottom-up knowledge) are useful, not only intrinsically, but also extrinsically in downstream tasks.

\section{Looking forward}

Future work could further apply the principle of combined top-down and bottom-up modelling to improve, extend, and address the shortcomings of the models presented in this thesis.
There are also a number of important future challenges for zero-resource speech processing in general.

\subsection{Extensions of this work}
\label{sec:conclusion_future}

In Chapter~\ref{chap:cae}, the cAE was always trained on discovered words pairs from a UTD system operating on traditional acoustic features.
Future work 
could use the cAE to provide improved features for UTD, which in turn can improve the word pairs used as weak supervision, and so forth.
Such a joint approach would be a principled way to use top-down knowledge of discovered words to improve bottom-level features, which in turn could improve word discovery, and so on.
Part of the argument in Chapter~\ref{chap:background} was that unsupervised representation learning should ideally be performed jointly with word segmentation and clustering, and this proposal would be a step in this direction for the task of UTD (discovering high-quality clusters with terms spread out over the data).

Although the segmental acoustic word embedding approach has several benefits (Section~\ref{sec:tidigits_related_studies_full_coverage}), it is also the biggest limitation of the full-coverage segmental Bayesian framework presented in Chapters~\ref{chap:tidigits} and~\ref{chap:bucktsong}.
Section~\ref{sec:tidigits_scaling} explained that both the efficiency and accuracy of the segmental framework is largely determined by the embedding method.
Despite this, we used existing methods proposed by others as black box components in our systems, and did not thoroughly investigate any potential improvements. 
Fortunately, the segmental approach is not tied to a specific embedding method; more accurate and efficient embeddings should therefore be an important focus of future work. 
In very recent work, also mentioned in Section~\ref{sec:background_segmental}, an autoencoding encoder-decoder neural network was used as acoustic word embedding function~\citep{chung+etal_arxiv16}, and such an approach could be incorporated into the segmentation model.  

More fundamentally, these observations again point back to Chapter~\ref{chap:background} where we argued for joint representation learning with segmentation and clustering.
Our systems followed this principle only to a limited extent.
For example, the joint refinement of the reference vector method in Sections~\ref{sec:tidigits_iterating} and~\ref{sec:tidigits_results} showed the advantage of feeding back top-down information into the embedding approach, but were  prohibitively slow for the large-vocabulary setting in Chapter~\ref{chap:bucktsong}.
In Chapter~\ref{chap:bucktsong} we showed that top-down information incorporated through the use of the cAE features lead to improvements, but this cAE was trained using a separate first-pass UTD system.
Nevertheless, the results of these imperfect experiments seem to further suggest that (ideally) frame-level representation learning, acoustic embedding representation learning, and segmentation and clustering should be performed jointly, and that both top-down and bottom-up methodologies would be beneficial in such a joint approach.
Applying these principles within the segmental Bayesian framework could be the long-term goal of future~work.

\edit{Throughout the thesis 
we noted that while most previous work in unsupervised speech processing followed an unsupervised subword modelling approach, we instead follow direct whole-word modelling approach. We do not argue that the latter is necessarily superior, since there are both advantages and disadvantages to the whole-word approach compared to subword modelling, as outlined in Section 4.1.3. A direct comparison of subword and whole-word unsupervised segmentation and clustering systems would be very valuable in future work. More importantly, future work could consider how subword and whole-word modelling can complement each other in a combined system. One natural extension of our system would be to incorporate subword modelling within the acoustic word embedding function. In supervised speech recognition there has been a long line of work (stretching from the early nineties to the present) attempting to automatically discover the lexicon and subword inventory of a language in settings where a pronunciation dictionary is not available~\citep{paliwal_icassp90,bahl+etal_tsap93,bacchiani+etal_icslp98,bacchiani+ostendorf_speccom99,goussard+niesler_prasa10,lee+etal_emnlp13}; this line of very relevant research would also give a thorough foundation for incorporating subword discovery into the unsupervised whole-word segmental system.}

\edit{The run-times noted in Sections~\ref{sec:tidigits_model_implementation} and~\ref{sec:bucktsong_modeldev} indicate that it would be impractical to apply the segmental Bayesian model to large speech corpora (potentially containing hundreds of hours of speech) because of its computational complexity.
It would therefore be essential in future work to consider how the model could be simplified and how approximations could be used in order to improve efficiency. \cite{shum+etal_taslp16} was able to dramatically improve the efficiency of the original phonetic discovery model of \cite{lee+glass_acl12} (see Section~\ref{sec:background_bottomup}) by using approximations which allowed the inference algorithm to be parallelized; similar modifications could be applied to the segmental Bayesian model.}

Neither the cAE nor the segmental Bayesian framework is strictly constrained to the zero-resource setting.
Low-resource supervised speech recognition, where small amounts of labelled speech data are available, is of huge practical relevance~\citep{besacier+etal_speechcom14,kamper+etal_speechcom12,kamper+etal_csl14}.
A cAE trained on a small amount of ground truth word pairs could easily be used to replace standard acoustic features as the input to traditional HMM- or DNN-based supervised speech recognition models.
Our segmental Bayesian framework could be used in a semi-supervised setting, where the model is applied to a large set of unlabelled data together with a small amount of labelled data; the labelled data could then be used to infer the identities of some of the discovered~clusters.

Throughout the thesis it was noted that the type of speech processing models developed here could be useful in cognitive models of language acquisition. As explained in Section~\ref{sec:background_symbolic_wordseg}, cognitive scientists use computational models as one way to specify and test particular theories~\citep{goldwater+etal_cognition09,rasanen_speechcom12}; the work presented here could form the basis of such a model for infant language acquisition from continuous speech input.
Despite using this 
motivation, we do not make any claims about the cognitive plausibility of the models which we developed here, nor that these are directly useful for testing cognitive theories in humans.
An investigation of these aspects and how our models could be improved by using insights from human studies are exciting and important avenues for further research.
In particular, at the outset of the thesis (Section~\ref{sec:intro_topdown_bottomup}) observations from human studies were used as an analogy to illustrate the potential benefits of combined top-down and bottom-up modelling~\citep{feldman+etal_ccss09}.
We showed that this claim is justified in computational models, but it remains to be seen how our models can be used to shed light on how humans use and combine bottom-up and top-down processing.

\subsection{Challenges in zero-resource speech processing}
\label{sec:conclusion_zerospeech}

In this thesis we focused on discovering representations and linguistic structure from raw audio data alone, as is done in many of the studies reviewed in Chapter~\ref{chap:background}.
Infants, however, have access to more than just the audio modality when learning their native language.
Similarly, robots that aim to acquire language in new linguistic environments are typically equipped with several different sensors---a microphone is just one of these.
As explained in Section~\ref{sec:background_multimodal}, there has therefore also been a different line of research considering 
language acquisition as a multi-modal problem, where unlabelled speech is coupled with some other input form.
Future work could consider extending the models presented here in situations where speech utterances are paired with other modalities such as images or touch sensors.
For the growing zero-resource processing community in general, this is also a crucial line of research that warrants further investigation (see the work of~\cite{harwath+glass_asru15} and \cite{harwath+etal_nips16} for recent supervised and unsupervised work coupling images and speech using a very promising neural approach).

Section~\ref{sec:background_eval} mentioned that the evaluation of zero-resource speech processing models remains a challenge.
The same-different and ABX tasks (Section~\ref{sec:background_phondisc_eval}) have allowed for comparisons between different frame-level representation learning methods.
But for zero-resource segmentation and clustering systems, different metrics are often used in different studies, making it very hard to compare approaches.
The problem of agreeing on a set of metrics is understandable since it is often not clear upfront what the output of a zero-resource system should be, and will often depend on the final use-case.
The metrics used as part of the Zero Resource Speech Challenge~\citep{ludusan+etal_lrec14,versteegh+etal_interspeech15} are a good step since they evaluate several different aspects of segmentation and clustering. But because of the large number of metrics, it becomes difficult to understand the relative performance of one system over another (consider the tables in Appendix~\ref{appen:results_zrs}, for example).
This thesis did not make contributions in solving this problem, but used existing metrics in order to compare to other studies as far as possible.
Future work could aim towards a smaller set of standardized metrics by investigating how and whether existing metrics correlate with each other, in order to identify redundancies.
One option could be to investigate the performance of supervised systems, and compare how zero-resource metrics correlate with the well-established phone and word error rates of such systems. 

\section{Conclusion}

This thesis made contributions in both unsupervised representation learning, and in unsupervised segmentation and clustering of unlabelled speech---the two main areas of focus in zero-resource speech processing.
For unsupervised representation learning, we proposed a new autoencoder-like neural network that uses automatically discovered terms in the speech data as a weak top-down supervision signal.
This model was the first neural network model of its kind, resulting in major improvements over previous state-of-the-art approaches.
For segmentation and clustering, we proposed a novel segmental Bayesian framework in which potential word candidates of variable duration are modelled as fixed-dimensional acoustic word embedding vectors.
We showed that this model outperformed previous systems in both small- and large-vocabulary unsupervised speech recognition tasks.
This was the first time that a zero-resource system that entirely segments its input was applied to multi-speaker large-vocabulary data.

The theme running through this work is that both top-down and bottom-up methodologies are beneficial for zero-resource speech processing.
Our autoencoder-like neural network used weak top-down constraints to produce features that were much more discriminative than purely bottom-up methods.
However, the model was only able to do so when using bottom-up pretraining on unlabelled speech. 
Similarly, we found that our full-coverage segmental system, which imposes a complete top-down segmentation of its entire input, could only be applied to larger vocabularies when taking into account bottom-up knowledge of automatically detected syllable boundaries, thereby outperforming a purely bottom-up syllable-based approach.
By using the autoencoder features (trained using both bottom-up and top-down knowledge) within the segmentation system, we showed that clusters can be made less speaker- and gender-specific.

This work has shown that a combination of top-down and bottom-up modelling is greatly beneficial in tackling zero-resource problems.
Further research could apply this principle in extending the models presented here, or in addressing the many exciting future challenges in zero-resource speech processing.



\bibliography{mybib}

\begin{thebibliography}{}

\bibitem[Abdel-Hamid et~al., 2013]{ossama+etal_interspeech13}
Abdel-Hamid, O., Deng, L., Yu, D., and Jiang, H. (2013).
\newblock Deep segmental neural networks for speech recognition.
\newblock In {\em Proc. Interspeech}.

\bibitem[Aimetti et~al., 2010]{aimetti+etal_interspeech10}
Aimetti, G., Moore, R.~K., and ten Bosch, L. (2010).
\newblock Discovering an optimal set of minimally contrasting acoustic speech
  units: A point of focus for whole-word pattern matching.
\newblock In {\em Proc. Interspeech}.

\bibitem[Anguera, 2012]{anguera_icassp12}
Anguera, X. (2012).
\newblock Speaker independent discriminant feature extraction for acoustic
  pattern-matching.
\newblock In {\em Proc. ICASSP}.

\bibitem[Athineos and Ellis, 2003]{athineos+ellis_asru03}
Athineos, M. and Ellis, D. P.~W. (2003).
\newblock Frequency-domain linear prediction for temporal features.
\newblock In {\em Proc. ASRU}.

\bibitem[Bacchiani and Ostendorf, 1999]{bacchiani+ostendorf_speccom99}
Bacchiani, M. and Ostendorf, M. (1999).
\newblock Joint lexicon, acoustic unit inventory and model design.
\newblock {\em Speech Commun.}, 29(2):99--114.

\bibitem[Bacchiani et~al., 1998]{bacchiani+etal_icslp98}
Bacchiani, M., Ostendorf, M., Sagisaka, Y., and Paliwal, K. (1998).
\newblock Using automatically-derived acoustic sub-word units in large
  vocabulary speech recognition.
\newblock In {\em Proc. ICSLP}.

\bibitem[Badino et~al., 2014]{badino+etal_icassp14}
Badino, L., Canevari, C., Fadiga, L., and Metta, G. (2014).
\newblock An auto-encoder based approach to unsupervised learning of subword
  units.
\newblock In {\em Proc. ICASSP}.

\bibitem[Badino et~al., 2015]{badino+etal_interspeech15}
Badino, L., Mereta, A., and Rosasco, L. (2015).
\newblock Discovering discrete subword units with binarized autoencoders and
  hidden-{Markov}-model encoders.
\newblock In {\em Proc. Interspeech}.

\bibitem[Bahl et~al., 1993]{bahl+etal_tsap93}
Bahl, L.~R., Brown, P.~F., de~Souza, P.~V., Mercer, R.~L., and Picheny, M.~A.
  (1993).
\newblock A method for the construction of acoustic {M}arkov models for words.
\newblock {\em IEEE Trans. Speech Audio Process.}, 1(4):443--452.

\bibitem[Barber, 2013]{barber}
Barber, D. (2013).
\newblock {\em Bayesian Reasoning and Machine Learning}.
\newblock Cambridge University Press, Cambridge, UK.

\bibitem[Belkin and Niyogi, 2003]{belkin+niyogi_neurocomp03}
Belkin, M. and Niyogi, P. (2003).
\newblock Laplacian eigenmaps for dimensionality reduction and data
  representation.
\newblock {\em Neural Comput.}, 15(6):1373--1396.

\bibitem[Belkin et~al., 2006]{belkin+etal_jmlr06}
Belkin, M., Niyogi, P., and Sindhwani, V. (2006).
\newblock Manifold regularization: A geometric framework for learning from
  labeled and unlabeled examples.
\newblock {\em J. Mach. Learn. Res.}, 7:2399--2434.

\bibitem[Bell et~al., 1990]{bell+etal_book46}
Bell, T.~C., Cleary, J.~G., and Witten, I.~H. (1990).
\newblock {\em Text Compression}.
\newblock Prentice Hall, Upper Saddle River, NJ.

\bibitem[Bengio, 2009]{bengio_ftml09}
Bengio, Y. (2009).
\newblock Learning deep architectures for {AI}.
\newblock {\em Found. Trends Mach. Learning}, 2(1):1--127.

\bibitem[Bengio et~al., 2007]{bengio+etal_nips07}
Bengio, Y., Lamblin, P., Popovici, D., and Larochelle, H. (2007).
\newblock Greedy layer-wise training of deep networks.
\newblock In {\em Proc. NIPS}.

\bibitem[Besacier et~al., 2014]{besacier+etal_speechcom14}
Besacier, L., Barnard, E., Karpov, A., and Schultz, T. (2014).
\newblock Automatic speech recognition for under-resourced languages: A survey.
\newblock {\em Speech Commun.}, 56:85--100.

\bibitem[Bortfeld et~al., 2005]{bortfeld+etal_psychol05}
Bortfeld, H., Morgan, J.~L., Golinkoff, R.~M., and Rathbun, K. (2005).
\newblock Mommy and me: Familiar names help launch babies into speech-stream
  segmentation.
\newblock {\em Psychol. Sci.}, 16(4):298--304.

\bibitem[Brent, 1999]{brent_ml99}
Brent, M.~R. (1999).
\newblock An efficient, probabilistically sound algorithm for segmentation and
  word discovery.
\newblock {\em Mach. Learn.}, 34(1-3):71--105.

\bibitem[Carlin et~al., 2011]{carlin+etal_icassp11}
Carlin, M.~A., Thomas, S., Jansen, A., and Hermansky, H. (2011).
\newblock Rapid evaluation of speech representations for spoken term discovery.
\newblock In {\em Proc. Interspeech}.

\bibitem[Chen et~al., 2015]{chen+etal_interspeech15}
Chen, H., Leung, C.-C., Xie, L., Ma, B., and Li, H. (2015).
\newblock Parallel inference of {Dirichlet} process {Gaussian} mixture models
  for unsupervised acoustic modeling: A feasibility study.
\newblock In {\em Proc. Interspeech}.

\bibitem[Christiansen et~al., 1998]{christiansen+etal_lcp98}
Christiansen, M.~H., Allen, J., and Seidenberg, M.~S. (1998).
\newblock Learning to segment speech using multiple cues: A connectionist
  model.
\newblock {\em Lang. Cognitive Proc.}, 13(2-3):221--268.

\bibitem[Chung et~al., 2013]{chung+etal_icassp13}
Chung, C.-T., Chan, C.-a., and Lee, L.-s. (2013).
\newblock Unsupervised discovery of linguistic structure including two-level
  acoustic patterns using three cascaded stages of iterative optimization.
\newblock In {\em Proc. ICASSP}.

\bibitem[Chung et~al., 2016]{chung+etal_arxiv16}
Chung, Y.-A., Wu, C.-C., Shen, C.-H., and Lee, H.-Y. (2016).
\newblock Audio word2vec: Unsupervised learning of audio segment
  representations using sequence-to-sequence recurrent neural networks.
\newblock {\em arXiv preprint arXiv:1603.00982}.

\bibitem[Dahl et~al., 2012]{dahl+etal_taslp12}
Dahl, G.~E., Yu, D., Deng, L., and Acero, A. (2012).
\newblock Context-dependent pre-trained deep neural networks for
  large-vocabulary speech recognition.
\newblock {\em IEEE Trans. Audio, Speech, Language Process.}, 20(1):30--42.

\bibitem[De~Vries et~al., 2014]{devries+etal_speechcom14}
De~Vries, N.~J., Davel, M.~H., Badenhorst, J., Basson, W.~D., De~Wet, F.,
  Barnard, E., and De~Waal, A. (2014).
\newblock A smartphone-based {ASR} data collection tool for under-resourced
  languages.
\newblock {\em Speech Commun.}, 56:119--131.

\bibitem[Eimas, 1999]{emias_jasm99}
Eimas, P.~D. (1999).
\newblock Segmental and syllabic representations in the perception of speech by
  young infants.
\newblock {\em J. Acoust. Soc. Am.}, 105(3):1901--1911.

\bibitem[Elman and Zipser, 1987]{elman+zipser_jasa88}
Elman, J.~L. and Zipser, D. (1987).
\newblock Learning the hidden structure of speech.
\newblock {\em J. Acoust. Soc. Am.}, 83:1615--1626.

\bibitem[Elsner et~al., 2013]{elsner+etal_emnlp13}
Elsner, M., Goldwater, S.~J., Feldman, N., and Wood, F. (2013).
\newblock A joint learning model of word segmentation, lexical acquisition and
  phonetic variability.
\newblock In {\em Proc. EMNLP}.

\bibitem[Feldman et~al., 2009]{feldman+etal_ccss09}
Feldman, N.~H., Griffiths, T.~L., and Morgan, J.~L. (2009).
\newblock Learning phonetic categories by learning a lexicon.
\newblock In {\em Proc. CCSS}.

\bibitem[Gillick et~al., 2011]{gillick+etal_asru11}
Gillick, D., Gillick, L., and Wegmann, S. (2011).
\newblock Don't multiply lightly: Quantifying problems with the acoustic model
  assumptions in speech recognition.
\newblock In {\em Proc. ASRU}.

\bibitem[Gish et~al., 2009]{gish+etal_interspeech09}
Gish, H., Siu, M.-H., Chan, A., and Belfield, B. (2009).
\newblock Unsupervised training of an {HMM}-based speech recognizer for topic
  classification.
\newblock In {\em Proc. Interspeech}.

\bibitem[Goldwater and Griffiths, 2007]{goldwater+griffiths_acl07}
Goldwater, S.~J. and Griffiths, T.~L. (2007).
\newblock A fully {Bayesian} approach to unsupervised part-of-speech tagging.
\newblock In {\em Proc. ACL}.

\bibitem[Goldwater et~al., 2009]{goldwater+etal_cognition09}
Goldwater, S.~J., Griffiths, T.~L., and Johnson, M. (2009).
\newblock A {B}ayesian framework for word segmentation: Exploring the effects
  of context.
\newblock {\em Cognition}, 112(1):21--54.

\bibitem[Goodfellow et~al., 2013]{goodfellow+etal_arxiv13}
Goodfellow, I.~J., Warde-Farley, D., Lamblin, P., Dumoulin, V., Mirza, M.,
  Pascanu, R., Bergstra, J., Bastien, F., and Bengio, Y. (2013).
\newblock Pylearn2: A machine learning research library.
\newblock {\em arXiv:1308.4214}.

\bibitem[Goussard and Niesler, 2010]{goussard+niesler_prasa10}
Goussard, G. and Niesler, T. (2010).
\newblock Automatic discovery of subword units and pronunciations for automatic
  speech recognition using {TIMIT}.
\newblock In {\em Proc. PRASA}.

\bibitem[Harwath and Glass, 2015]{harwath+glass_asru15}
Harwath, D. and Glass, J. (2015).
\newblock Deep multimodal semantic embeddings for speech and images.
\newblock In {\em Proc. ASRU}.

\bibitem[Harwath et~al., 2016]{harwath+etal_nips16}
Harwath, D., Torralba, A., and Glass, J. (2016).
\newblock Unsupervised learning of spoken language with visual context.
\newblock In {\em Proc. NIPS}.

\bibitem[Heymann et~al., 2013]{heymann+etal_asru13}
Heymann, J., Walter, O., Haeb-Umbach, R., and Raj, B. (2013).
\newblock Unsupervised word segmentation from noisy input.
\newblock In {\em Proc. ASRU}.

\bibitem[Hinton et~al., 2012]{hinton+etal_spm2012}
Hinton, G., Deng, L., Yu, D., Dahl, G.~E., Mohamed, A.-R., Jaitly, N., Senior,
  A., Vanhoucke, V., Nguyen, P., Sainath, T.~N., and Kingsbury, B. (2012).
\newblock Deep neural networks for acoustic modeling in speech recognition: The
  shared views of four research groups.
\newblock {\em {IEEE} Signal Process. Mag.}, 29(6):82--97.

\bibitem[Hinton et~al., 2006]{hinton+etal_neurocomp06}
Hinton, G.~E., Osindero, S., and Teh, Y.-W. (2006).
\newblock A fast learning algorithm for deep belief nets.
\newblock {\em Neural Comput.}, 18(7):1527--1554.

\bibitem[Hinton and Salakhutdinov, 2006]{hinton+salakhutdinov_science06}
Hinton, G.~E. and Salakhutdinov, R.~R. (2006).
\newblock Reducing the dimensionality of data with neural networks.
\newblock {\em Science}, 313(5786):504--507.

\bibitem[Hunt et~al., 1991]{hunt+etal_icassp91}
Hunt, M., Richardson, S.~M., Bateman, D.~C., and Piau, A. (1991).
\newblock An investigation of {PLP} and {IMELDA} acoustic representations and
  of their potential for combination.
\newblock In {\em Proc. ICASSP}.

\bibitem[Jansen and Church, 2011]{jansen+church_interspeech11}
Jansen, A. and Church, K. (2011).
\newblock Towards unsupervised training of speaker independent acoustic models.
\newblock In {\em Proc. Interspeech}.

\bibitem[Jansen et~al., 2010]{jansen+etal_interspeech10}
Jansen, A., Church, K., and Hermansky, H. (2010).
\newblock Towards spoken term discovery at scale with zero resources.
\newblock In {\em Proc. Interspeech}.

\bibitem[Jansen et~al., 2013a]{jansen+etal_icassp13}
Jansen, A., Dupoux, E., Goldwater, S.~J., Johnson, M., Khudanpur, S., Church,
  K., Feldman, N., Hermansky, H., Metze, F., Rose, R., Seltzer, M., Clark, P.,
  McGraw, I., Varadarajan, B., Bennett, E., Borschinger, B., Chiu, J., Dunbar,
  E., Fourtassi, A., Harwath, D., Lee, C.-y., Levin, K., Norouzian, A.,
  Peddinti, V., Richardson, R., Schatz, T., and Thomas, S. (2013a).
\newblock A summary of the 2012 {JHU CLSP} workshop on zero resource speech
  technologies and models of early language acquisition.
\newblock In {\em Proc. ICASSP}.

\bibitem[Jansen et~al., 2013b]{jansen+etal_icassp13b}
Jansen, A., Thomas, S., and Hermansky, H. (2013b).
\newblock Weak top-down constraints for unsupervised acoustic model training.
\newblock In {\em Proc. ICASSP}.

\bibitem[Jansen and Van~Durme, 2011]{jansen+vandurme_asru11}
Jansen, A. and Van~Durme, B. (2011).
\newblock Efficient spoken term discovery using randomized algorithms.
\newblock In {\em Proc. ASRU}.

\bibitem[Jelinek and Mercer, 1980]{jelinek+mercer_prp80}
Jelinek, F. and Mercer, R.~L. (1980).
\newblock Interpolated estimation of {M}arkov source parameters from sparse
  data.
\newblock {\em Pattern Recogn. Pract.}, pages 381--402.

\bibitem[Kamper, 2015]{kamper_bayesgmm15}
Kamper, H. (2015).
\newblock Gibbs sampling for fitting finite and infinite {G}aussian mixture
  models.

\bibitem[Kamper et~al., 2014a]{kamper+etal_csl14}
Kamper, H., De~Wet, F., Hain, T., and Niesler, T.~R. (2014a).
\newblock Capitalising on {North American} speech resources for the development
  of a {South African English} large vocabulary speech recognition system.
\newblock {\em Comput. Speech Lang.}, 28(6):1255--1268.

\bibitem[Kamper et~al., 2015a]{kamper+etal_icassp15}
Kamper, H., Elsner, M., Jansen, A., and Goldwater, S.~J. (2015a).
\newblock Unsupervised neural network based feature extraction using weak
  top-down constraints.
\newblock In {\em Proc. ICASSP}.

\bibitem[Kamper et~al., 2015b]{kamper+etal_interspeech15}
Kamper, H., Goldwater, S.~J., and Jansen, A. (2015b).
\newblock Fully unsupervised small-vocabulary speech recognition using a
  segmental {B}ayesian model.
\newblock In {\em Proc. Interspeech}.

\bibitem[Kamper et~al., 2016a]{kamper+etal_arxiv16}
Kamper, H., Jansen, A., and Goldwater, S. (2016a).
\newblock A segmental framework for fully-unsupervised large-vocabulary speech
  recognition.
\newblock {\em arXiv preprint arXiv:1606.06950}.

\bibitem[Kamper et~al., 2016b]{kamper+etal_taslp16}
Kamper, H., Jansen, A., and Goldwater, S.~J. (2016b).
\newblock Unsupervised word segmentation and lexicon discovery using acoustic
  word embeddings.
\newblock {\em IEEE/ACM Trans. Audio, Speech, Language Process.},
  24(4):669--679.

\bibitem[Kamper et~al., 2014b]{kamper+etal_slt14}
Kamper, H., Jansen, A., King, S., and Goldwater, S.~J. (2014b).
\newblock Unsupervised lexical clustering of speech segments using
  fixed-dimensional acoustic embeddings.
\newblock In {\em Proc. SLT}.

\bibitem[Kamper et~al., 2012]{kamper+etal_speechcom12}
Kamper, H., Mukanya, F. J.~M., and Niesler, T. (2012).
\newblock Multi-accent acoustic modelling of {South African English}.
\newblock {\em Speech Commun.}, 54(6):801--813.

\bibitem[Kamper et~al., 2016c]{kamper+etal_icassp16}
Kamper, H., Wang, W., and Livescu, K. (2016c).
\newblock Deep convolutional acoustic word embeddings using word-pair side
  information.
\newblock In {\em Proc. ICASSP}.

\bibitem[Kuhl, 2004]{kuhl_nature04}
Kuhl, P.~K. (2004).
\newblock Early language acquisition: Cracking the speech code.
\newblock {\em Nature Neurosci.}, 5(11):831--843.

\bibitem[Lee, 2014]{lee_phd14}
Lee, C.-y. (2014).
\newblock {\em Discovering Linguistic Structures in Speech: Models and
  Applications}.
\newblock PhD thesis, Massachusetts Institute of Technology, Cambridge, MA.

\bibitem[Lee and Glass, 2012]{lee+glass_acl12}
Lee, C.-y. and Glass, J.~R. (2012).
\newblock A nonparametric {B}ayesian approach to acoustic model discovery.
\newblock In {\em Proc. ACL}.

\bibitem[Lee et~al., 2015]{lee+etal_tacl15}
Lee, C.-y., O'Donnell, T., and Glass, J.~R. (2015).
\newblock Unsupervised lexicon discovery from acoustic input.
\newblock {\em Trans. ACL}, 3:389--403.

\bibitem[Lee et~al., 2013]{lee+etal_emnlp13}
Lee, C.-y., Zhang, Y., and Glass, J.~R. (2013).
\newblock Joint learning of phonetic units and word pronunciations for {ASR}.
\newblock In {\em Proc. EMNLP}.

\bibitem[Lee and Lee, 2013]{lee+lee_taslp13}
Lee, H.-y. and Lee, L.-s. (2013).
\newblock Enhanced spoken term detection using support vector machines and
  weighted pseudo examples.
\newblock {\em IEEE Trans. Audio, Speech, Language Process.}, 21(6):1272--1284.

\bibitem[Leonard, 1984]{leonard_icassp84}
Leonard, R.~G. (1984).
\newblock A database for speaker-independent digit recognition.
\newblock In {\em Proc. ICASSP}.

\bibitem[Levin et~al., 2013]{levin+etal_asru13}
Levin, K., Henry, K., Jansen, A., and Livescu, K. (2013).
\newblock Fixed-dimensional acoustic embeddings of variable-length segments in
  low-resource settings.
\newblock In {\em Proc. ASRU}.

\bibitem[Levin et~al., 2015]{levin+etal_icassp15}
Levin, K., Jansen, A., and Van~Durme, B. (2015).
\newblock Segmental acoustic indexing for zero resource keyword search.
\newblock In {\em Proc. ICASSP}.

\bibitem[Ludusan et~al., 2014]{ludusan+etal_lrec14}
Ludusan, B., Versteegh, M., Jansen, A., Gravier, G., Cao, X.-N., Johnson, M.,
  and Dupoux, E. (2014).
\newblock Bridging the gap between speech technology and natural language
  processing: An evaluation toolbox for term discovery systems.
\newblock In {\em Proc. LREC}.

\bibitem[Lyzinski et~al., 2015]{lyzinski+etal_interspeech15}
Lyzinski, V., Sell, G., and Jansen, A. (2015).
\newblock An evaluation of graph clustering methods for unsupervised term
  discovery.
\newblock In {\em Proc. Interspeech}.

\bibitem[MacKay and Peto, 1995]{mackay+peto_nle95}
MacKay, D.~J. and Peto, L. C.~B. (1995).
\newblock A hierarchical {D}irichlet language model.
\newblock {\em Nat. Lang. Eng.}, 1(3):289--308.

\bibitem[McInnes and Goldwater, 2011]{mcinnes+goldwater_ccss11}
McInnes, F.~R. and Goldwater, S.~J. (2011).
\newblock Unsupervised extraction of recurring words from infant-directed
  speech.
\newblock In {\em Proc. CCSS}.

\bibitem[McQueen, 1998]{mcqueen_memory98}
McQueen, J.~M. (1998).
\newblock Segmentation of continuous speech using phonotactics.
\newblock {\em J. Memory Lang.}, 39(1):21--46.

\bibitem[Metze et~al., 2013]{metze+etal_icassp13}
Metze, F., Anguera, X., Barnard, E., Davel, M., and Gravier, G. (2013).
\newblock The spoken web search task at {MediaEval} 2012.
\newblock In {\em Proc. ICASSP}.

\bibitem[Mochihashi et~al., 2009]{mochihashi+etal_acl09}
Mochihashi, D., Yamada, T., and Ueda, N. (2009).
\newblock Bayesian unsupervised word segmentation with nested {P}itman-{Y}or
  language modeling.
\newblock In {\em Proc. ACL}.

\bibitem[Murphy, 2007]{murphy_bayesgauss07}
Murphy, K.~P. (2007).
\newblock Conjugate {B}ayesian analysis of the {G}aussian distribution.

\bibitem[Murphy, 2012]{murphy}
Murphy, K.~P. (2012).
\newblock {\em Machine Learning: A Probabilistic Perspective}.
\newblock MIT Press, Cambridge, MA.

\bibitem[Neubig et~al., 2010]{neubig+etal_interspeech10}
Neubig, G., Mimura, M., Mori, S., and Kawahara, T. (2010).
\newblock Learning a language model from continuous speech.
\newblock In {\em Proc. Interspeech}.

\bibitem[O'Grady and Pearlmutter, 2008]{ogrady+pearlmutter_neurocomp08}
O'Grady, P.~D. and Pearlmutter, B.~A. (2008).
\newblock Discovering speech phones using convolutive non-negative matrix
  factorisation with a sparseness constraint.
\newblock {\em Neurocomputing}, 72(1):88--101.

\bibitem[Paliwal, 1990]{paliwal_icassp90}
Paliwal, K.~K. (1990).
\newblock Lexicon-building methods for an acoustic sub-word based speech
  recognizer.
\newblock In {\em Proc. ICASSP}.

\bibitem[Park and Glass, 2008]{park+glass_taslp08}
Park, A.~S. and Glass, J.~R. (2008).
\newblock Unsupervised pattern discovery in speech.
\newblock {\em IEEE Trans. Audio, Speech, Language Process.}, 16(1):186--197.

\bibitem[Pitt et~al., 2005]{pitt+etal_speechcom05}
Pitt, M.~A., Johnson, K., Hume, E., Kiesling, S., and Raymond, W. (2005).
\newblock The {Buckeye} corpus of conversational speech: Labeling conventions
  and a test of transcriber reliability.
\newblock {\em Speech Commun.}, 45(1):89--95.

\bibitem[{R{\"a}s{\"a}nen}, 2012]{rasanen_speechcom12}
{R{\"a}s{\"a}nen}, O.~J. (2012).
\newblock Computational modeling of phonetic and lexical learning in early
  language acquisition: Existing models and future directions.
\newblock {\em Speech Commun.}, 54:975--997.

\bibitem[{R{\"a}s{\"a}nen} et~al., 2015]{rasanen+etal_interspeech15}
{R{\"a}s{\"a}nen}, O.~J., Doyle, G., and Frank, M.~C. (2015).
\newblock Unsupervised word discovery from speech using automatic segmentation
  into syllable-like units.
\newblock In {\em Proc. Interspeech}.

\bibitem[R{\"a}s{\"a}nen et~al., 2016]{rasanen+etal_submission16}
R{\"a}s{\"a}nen, O.~J., Doyle, G., and Frank, M.~C. (2016).
\newblock Pre-linguistic rhythmic segmentation of speech into syllabic units.
\newblock {\em In submission}.

\bibitem[Rasmussen, 1999]{rasmussen_nips99}
Rasmussen, C.~E. (1999).
\newblock The infinite {G}aussian mixture model.
\newblock In {\em Proc. NIPS}.

\bibitem[Renkens et~al., 2014]{renkens+etal_slt14}
Renkens, V., Janssens, S., Ons, B., Gemmeke, J.~F., and Van~hamme, H. (2014).
\newblock Acquisition of ordinal words using weakly supervised {NMF}.
\newblock In {\em Proc. SLT}.

\bibitem[Renkens and Van~hamme, 2015]{renkens+vanhamme_interspeech15}
Renkens, V. and Van~hamme, H. (2015).
\newblock Mutually exclusive grounding for weakly supervised non-negative
  matrix factorisation.
\newblock In {\em Proc. Interspeech}.

\bibitem[Renshaw, 2016]{renshaw_masters16}
Renshaw, D. (2016).
\newblock Representation learning for unsupervised speech processing.
\newblock Master's thesis, University of Edinburgh.

\bibitem[Renshaw et~al., 2015]{renshaw+etal_interspeech15}
Renshaw, D., Kamper, H., Jansen, A., and Goldwater, S.~J. (2015).
\newblock A comparison of neural network methods for unsupervised
  representation learning on the {Zero Resource Speech Challenge}.
\newblock In {\em Proc. Interspeech}.

\bibitem[Resnik and Hardisty, 2010]{resnik+hardisty_gibbs_tutorial10}
Resnik, P. and Hardisty, E. (2010).
\newblock Gibbs sampling for the uninitiated.
\newblock Technical report, University of Maryland, College Park, MD.

\bibitem[Rumelhart et~al., 1986]{rumelhart+etal_pdp86}
Rumelhart, D.~E., Hinton, G.~E., and Williams, R.~J. (1986).
\newblock Learning internal representations by error propagation.
\newblock {\em Parallel Distributed Processing}, 1:318--362.
\newblock MIT Press.

\bibitem[Sakoe and Chiba, 1978]{sakoe+chiba_assp78}
Sakoe, H. and Chiba, S. (1978).
\newblock Dynamic programming algorithm optimization for spoken word
  recognition.
\newblock {\em IEEE Trans. Acoust., Speech, Signal Process.}, 26(1):43--49.

\bibitem[Schatz et~al., 2013]{schatz+etal_interspeech13}
Schatz, T., Peddinti, V., Bach, F., Jansen, A., Hermansky, H., and Dupoux, E.
  (2013).
\newblock Evaluating speech features with the minimal-pair {ABX} task: Analysis
  of the classical {MFC}/{PLP} pipeline.
\newblock In {\em Proc. Interspeech}.

\bibitem[Scott, 2002]{scott_jasa02}
Scott, S.~L. (2002).
\newblock Bayesian methods for hidden {M}arkov models.
\newblock {\em J. Am. Stat. Assoc.}, 97(457):337--351.

\bibitem[Shum et~al., 2016]{shum+etal_taslp16}
Shum, S.~H., Harwath, D.~F., Dehak, N., and Glass, J.~R. (2016).
\newblock On the use of acoustic unit discovery for language recognition.
\newblock {\em IEEE Trans. Acoust., Speech, Signal Process.}, 24(9):1665--1676.

\bibitem[Siu et~al., 2010]{siu+etal_interspeech10}
Siu, M.-H., Gish, H., Chan, A., and Belfield, W. (2010).
\newblock Improved topic classification and keyword discovery using an
  {HMM}-based speech recognizer trained without supervision.
\newblock In {\em Proc. Interspeech}.

\bibitem[Siu et~al., 2014]{siu+etal_csl14}
Siu, M.-H., Gish, H., Chan, A., Belfield, W., and Lowe, S. (2014).
\newblock Unsupervised training of an {HMM}-based self-organizing unit
  recognizer with applications to topic classification and keyword discovery.
\newblock {\em Comput. Speech Lang.}, 28(1):210--223.

\bibitem[Siu et~al., 2011]{siu+etal_interspeech11}
Siu, M.-H., Gish, H., Lowe, S., and Chan, A. (2011).
\newblock Unsupervised audio patterns discovery using {HMM}-based
  self-organized units.
\newblock In {\em Proc. Interspeech}.

\bibitem[Stouten et~al., 2008]{stouten+etal_ieee08}
Stouten, V., Demuynck, K., and Van~hamme, H. (2008).
\newblock Discovering phone patterns in spoken utterances by non-negative
  matrix factorization.
\newblock {\em IEEE Signal Proc. Let.}, 15:131--134.

\bibitem[Sun and Van~hamme, 2013]{sun+vanhamme_csl13}
Sun, M. and Van~hamme, H. (2013).
\newblock Joint training of non-negative {T}ucker decomposition and discrete
  density hidden {M}arkov models.
\newblock {\em Comput. Speech Lang.}, 27(4):969--988.

\bibitem[Synnaeve et~al., 2014]{synnaeve+etal_slt14}
Synnaeve, G., Schatz, T., and Dupoux, E. (2014).
\newblock Phonetics embedding learning with side information.
\newblock In {\em Proc. SLT}.

\bibitem[Taniguchi et~al., 2015]{taniguchi+etal_arxiv15}
Taniguchi, T., Nagai, T., Nakamura, T., Iwahashi, N., Ogata, T., and Asoh, H.
  (2015).
\newblock Symbol emergence in robotics: A survey.
\newblock {\em arXiv preprint arXiv:1509.08973}.

\bibitem[ten Bosch and Cranen, 2007]{tenbosch+cranen_interspeech07}
ten Bosch, L. and Cranen, B. (2007).
\newblock A computational model for unsupervised word discovery.
\newblock In {\em Proc. Interspeech}.

\bibitem[Thiolli{\`e}re et~al., 2015]{thiolliere+etal_interspeech15}
Thiolli{\`e}re, R., Dunbar, E., Synnaeve, G., Versteegh, M., and Dupoux, E.
  (2015).
\newblock A hybrid dynamic time warping-deep neural network architecture for
  unsupervised acoustic modeling.
\newblock In {\em Proc. Interspeech}.

\bibitem[Vanhainen and Salvi, 2014]{vanhainen+salvi_icassp14}
Vanhainen, N. and Salvi, G. (2014).
\newblock Pattern discovery in continuous speech using block diagonal infinite
  {HMM}.
\newblock In {\em Proc. ICASSP}.

\bibitem[Varadarajan et~al., 2008]{varadarajan+etal_acl08}
Varadarajan, B., Khudanpur, S., and Dupoux, E. (2008).
\newblock Unsupervised learning of acoustic sub-word units.
\newblock In {\em Proc. ACL}.

\bibitem[Venkataraman, 2001]{venkataraman_cl01}
Venkataraman, A. (2001).
\newblock A statistical model for word discovery in transcribed speech.
\newblock {\em Comput. Linguist.}, 27(3):351--372.

\bibitem[Versteegh et~al., 2016]{versteegh+etal_sltu16}
Versteegh, M., Anguera, X., Jansen, A., and Dupoux, E. (2016).
\newblock The {Zero Resource Speech Challenge 2015}: Proposed approaches and
  results.
\newblock In {\em Proc. SLTU}.

\bibitem[Versteegh et~al., 2015]{versteegh+etal_interspeech15}
Versteegh, M., Thiolli{\`e}re, R., Schatz, T., Cao, X.~N., Anguera, X., Jansen,
  A., and Dupoux, E. (2015).
\newblock The {Zero Resource Speech Challenge} 2015.
\newblock In {\em Proc. Interspeech}.

\bibitem[Vincent et~al., 2008]{vincent+etal_icml08}
Vincent, P., Larochelle, H., Bengio, Y., and Manzagol, P.-A. (2008).
\newblock Extracting and composing robust features with denoising autoencoders.
\newblock In {\em Proc. ICML}.

\bibitem[Walter et~al., 2013]{walter+etal_asru13}
Walter, O., Korthals, T., Haeb-Umbach, R., and Raj, B. (2013).
\newblock A hierarchical system for word discovery exploiting {DTW}-based
  initialization.
\newblock In {\em Proc. ASRU}.

\bibitem[Weng et~al., 2014]{weng+etal_icassp14}
Weng, C., Yu, D., Watanabe, S., and Juang, B.-H. (2014).
\newblock Recurrent deep neural networks for robust speech recognition.
\newblock In {\em Proc. ICASSP}.

\bibitem[Wood and Black, 2012]{wood+black_jnm08}
Wood, F. and Black, M.~J. (2012).
\newblock A nonparametric {B}ayesian alternative to spike sorting.
\newblock {\em J. Neurosci. Methods}, 173(1):1--12.

\bibitem[Young et~al., 2009]{htk_book}
Young, S.~J., Evermann, G., Gales, M. J.~F., Hain, T., Kershaw, D., Liu, X.,
  Moore, G.~L., Odell, J.~J., Ollason, D., Povey, D., Valtchev, V., and
  Woodland, P.~C. (2009).
\newblock {\em The {HTK} Book (for {HTK} Version 3.4)}.
\newblock Cambridge University Engineering Department.

\bibitem[Yu et~al., 2013]{yu+etal_iclr13}
Yu, D., Seltzer, M., Li, J., Huang, J.-T., and Seide, F. (2013).
\newblock Feature learning in deep neural networks: Studies on speech
  recognition.
\newblock In {\em Proc. ICLR}.

\bibitem[Yuan et~al., 2016]{yuan+etal_interspeech16}
Yuan, Y., Leung, C.-C., Xie, L., Ma, B., and Li, H. (2016).
\newblock Learning neural network representations using cross-lingual
  bottleneck features with word-pair information.
\newblock In {\em Proc. Interspeech}.

\bibitem[Zeghidour et~al., 2016]{zeghidour+etal_icassp16}
Zeghidour, N., Synnaeve, G., Versteegh, M., and Dupoux, E. (2016).
\newblock A deep scattering spectrum-deep {Siamese} network pipeline for
  unsupervised acoustic modeling.
\newblock In {\em Proc. ICASSP}.

\bibitem[Zeiler et~al., 2013]{zeiler+etal_icassp13}
Zeiler, M.~D., Ranzato, M., Monga, R., Mao, M., Yang, K., Le, Q.~V., Nguyen,
  P., Senior, A., Vanhoucke, V., Dean, J., and Hinton, G.~E. (2013).
\newblock On rectified linear units for speech processing.
\newblock In {\em Proc. ICASSP}.

\bibitem[Zhang and Glass, 2009]{zhang+glass_asru09}
Zhang, Y. and Glass, J.~R. (2009).
\newblock Unsupervised spoken keyword spotting via segmental {DTW} on
  {G}aussian posteriorgrams.
\newblock In {\em Proc. ASRU}.

\bibitem[Zhang and Glass, 2010]{zhang+glass_icassp10}
Zhang, Y. and Glass, J.~R. (2010).
\newblock Towards multi-speaker unsupervised speech pattern discovery.
\newblock In {\em Proc. ICASSP}.

\bibitem[Zhang et~al., 2012]{zhang+etal_icassp12}
Zhang, Y., Salakhutdinov, R., Chang, H.-A., and Glass, J.~R. (2012).
\newblock Resource configurable spoken query detection using deep {B}oltzmann
  machines.
\newblock In {\em Proc. ICASSP}.

\bibitem[Zweig and Nguyen, 2010]{zweig+nguyen_interspeech10}
Zweig, G. and Nguyen, P. (2010).
\newblock {SCARF}: A segmental conditional random field toolkit for speech
  recognition.
\newblock In {\em Proc. Interspeech}.

\end{thebibliography}

\appendix
\graphicspath{{appendices/fig/}}



\chapter{Sampling the segmentation using a bigram model}
\makeatletter\@mkboth{}{Appendix}\makeatother
\label{appen:derivations_bigramseg}

\begin{figure}[!htbp]
    \centering
    \includegraphics[scale=1.0]{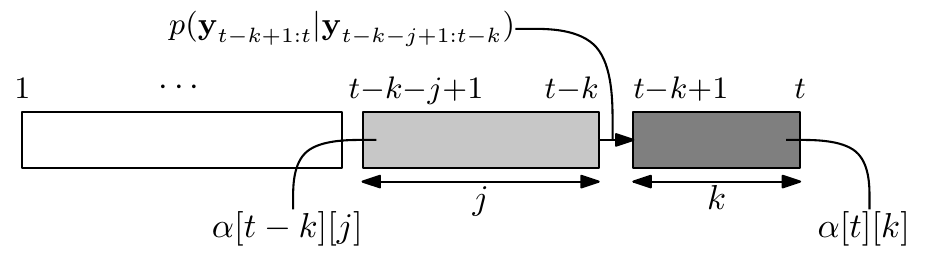}
    \caption{
    Forward filtering for $\alpha[t][k]$, the joint density of the frame sequence $\vec{y}_{1:t}$ with the last $k$ frames being a word in the bigram segmentation model. To derive the recursive expression for $\alpha[t][k]$, we marginalize over the number of frames $j$ of the preceding word by expanding the light grey area ($j$ frames) from $1$ to $t-k$.
    }
    \label{fig:forward_filtering}
\end{figure}

Component assignment in the bigram model is described in Section~\ref{sec:bucktsong_bisampling_assignments}.
Here segmentation is described.
A summary of segmentation in the unigram model is given in Section~\ref{sec:tidigits_word_segmentation}.
For the bigram model, the forward variables are defined differently: $\alpha[t][k]$ is defined as the density of the frame sequence $\vec{y}_{1:t}$ with the last $k$ frames being a word: $\alpha[t][k] \defeq p(\vec{y}_{1:t}, q_t = k)$.
In this definition, the variable $q_t$ is used to indicate the number of acoustic observation frames in the hypothesized word that ends at frame $t$: if $q_t = k$, then $\vec{y}_{t - k + 1:t}$ is a word.
Similarly to the unigram case, we can derive a recursive expression for $\alpha[t][k]$, in this case by marginalizing over the number of frames in the word preceding the last word:
\begin{align}
    \alpha[t][k]
    &= p(\vec{y}_{1:t}, q_t = k | h^-) \nonumber \\
    &= \sum_{j = 1}^{t-k} p(\vec{y}_{1:t}, q_t = k, q_{t-k} = j | h^-) \nonumber \\
    &= \sum_{j = 1}^{t-k} p(\vec{y}_{1:{t-k}}, \vec{y}_{{t - k + 1}:t}, q_t = k, q_{t-k} = j | h^-) \nonumber \\
    &= \sum_{j = 1}^{t-k} p(\vec{y}_{{t - k + 1}:t} | \vec{y}_{1:{t-k}}, q_t = k, q_{t-k} = j, h^-) p(\vec{y}_{1:{t-k}}, q_t = k, q_{t-k} = j | h^-) \nonumber \\
    &= \sum_{j = 1}^{t-k} p(\vec{y}_{{t - k + 1}:t} | \vec{y}_{t-k-j+1:{t-k}}, h^-) p(\vec{y}_{1:{t-k}}, q_{t-k} = j | h^-)  \label{eq:bi_forward1} \\
    &= \sum_{j = 1}^{t-k} p(\vec{y}_{{t - k + 1}:t} | \vec{y}_{t-k-j+1:{t-k}}, h^-) \alpha[t - k][j] \label{eq:bi_forward}
\end{align}
where we use a uniform prior over $q_t$ in~\eqref{eq:bi_forward1}.
This marginalization is illustrated in Figure~\ref{fig:forward_filtering}.
Recursion starts with $\alpha[0][0] = 1$ and forward variables are calculated for $1 \leq t \leq M$ and $\max(1, t-M) \leq k \leq t$ \citep{mochihashi+etal_acl09}.

\begin{figure}[tbp]
    \centering
    \includegraphics[width=\textwidth]{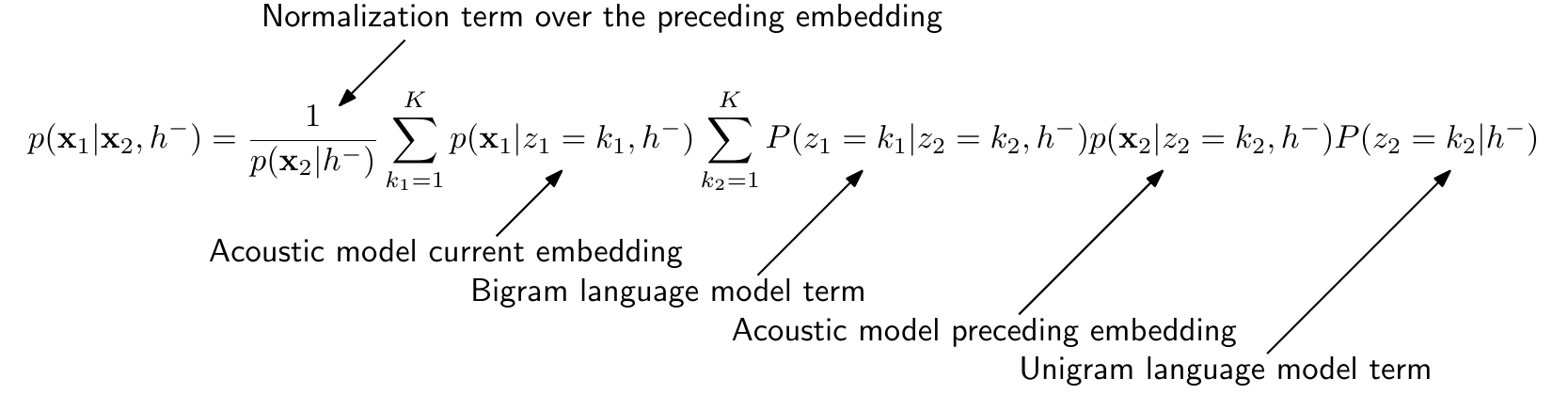}
    \caption{
    An illustration of the different terms in the marginal $p(\vec{x}_1 | \vec{x}_2, h^-)$ derived in~\eqref{eq:bigram2}, which involves marginalizing out the component assignments over both the current embedding $\vec{x}_1$ and the preceding embedding $\vec{x}_2$.
    }
    \label{fig:marginal}
\end{figure}

In analogy to~\eqref{eq:prob_segment}, we treat the required first term in~\eqref{eq:bi_forward} as
\begin{equation}
    p(\vec{y}_{{t - k + 1}:t} | \vec{y}_{t-k-j+1:{t-k}}, h^-) \defeq \left[p \left(\vec{x}_1 | \vec{x}_2, h^- \right) \right]^k 
\end{equation}
with $\vec{x}_1 = f_e(\vec{y}_{{t - k + 1}:t})$ and $\vec{x}_2 = f_e(\vec{y}_{t-k-j+1:{t-k}})$ the acoustic word embeddings calculated on speech frames $\vec{y}_{{t - k + 1}:t}$ and $\vec{y}_{t-k-j+1:{t-k}}$, respectively.
As in~\eqref{eq:prob_segment}, this conditional density is scaled by the number of frames $k$ 
in the speech segment over which the first embedding is calculated.  The inside term is obtained by marginalizing over the component assignments of both $\vec{x}_1$ and $\vec{x}_2$:
\begin{align}
    &p(\vec{x}_1 | \vec{x}_2, h^-) \nonumber \\
    &= \sum_{k_1 = 1}^{K} p(\vec{x}_1, z_1 = k_1 | \vec{x}_2, h^-) \nonumber \\
    &= \sum_{k_1 = 1}^{K} p(\vec{x}_1 | z_1 = k_1, h^-) P(z_1 = k_1 | \vec{x}_2, h^-) \nonumber \\
    &= \sum_{k_1 = 1}^{K} p(\vec{x}_1 | z_1 = k_1, h^-) \sum_{k_2 = 1}^{K} P(z_1 = k_1, z_2 = k_2 | \vec{x}_2, h^-) \nonumber \\
    &= \sum_{k_1 = 1}^{K} p(\vec{x}_1 | z_1 = k_1, h^-) \sum_{k_2 = 1}^{K} P(z_1 = k_1 | z_2 = k_2, h^-) P(z_2 = k_2 | \vec{x}_2, h^-) \label{eq:bigram1} \\
    &= \sum_{k_1 = 1}^{K} p(\vec{x}_1 | z_1 = k_1, h^-) \sum_{k_2 = 1}^{K} P(z_1 = k_1 | z_2 = k_2, h^-) \frac{p(\vec{x}_2 | z_2 = k_2, h^-) P(z_2 = k_2 | h^-)}{p(\vec{x}_2 | h^-)} \nonumber \\
    &= \sum_{k_1 = 1}^{K} p(\vec{x}_1 | z_1 = k_1, h^-) \sum_{k_2 = 1}^{K} P(z_1 = k_1 | z_2 = k_2, h^-) \frac{p(\vec{x}_2 | z_2 = k_2, h^-) P(z_2 = k_2 | h^-)}{\sum_{k'} p(\vec{x}_2 | z_2 = k', h^-) P(z_2 = k' | h^-)} \nonumber \\
    &=  \frac{1}{p(\vec{x}_2 | h^-)} \sum_{k_1 = 1}^{K} p(\vec{x}_1 | z_1 = k_1, h^-) \sum_{k_2 = 1}^{K} P(z_1 = k_1 | z_2 = k_2, h^-) {p(\vec{x}_2 | z_2 = k_2, h^-) P(z_2 = k_2 | h^-)} \label{eq:bigram2}
\end{align}
The different terms in equation~\eqref{eq:bigram2} are annotated in Figure~\ref{fig:marginal} in order to give some intuition into the `score' that is assigned to $\vec{x}_1$ given that it is preceded by $\vec{x}_2$.

As mentioned in~\citep{kamper+etal_taslp16}, the marginalization involved in~\eqref{eq:bigram2} could prove computationally too costly. For marginalization in the unigram case in~\eqref{eq:likelihood_fbgmm}, we had to sum over $K$ terms; here we need to sum over a $K \times K$ terms.
One approximation we can make is to assume that the posterior $P(z_2 = k_2 | \vec{x}_2, h^-)$ in~\eqref{eq:bigram1} is very peaked (close to $1$) around a single $k_2'$, with all the other possible component assignments giving a posterior probability close to $0$.
This would lead to the approximation
\begin{equation}
    p(\vec{x}_1 | \vec{x}_2, h^-) \approx \sum_{k_1 = 1}^{K} p(\vec{x}_1 | z_1 = k_1, h^-) P(z_1 = k_1 | z_2 = k_2', h^-)
    \label{eq:bigram_approx}
\end{equation}
with
\begin{align}
    k_2' &= \argmax_{k_2} P(z_2 = k_2 | \vec{x}_2, h^-) \nonumber \\
    &= \argmax_{k_2} p(\vec{x}_2 | z_2 = k_2, h^-) P(z_2 = k_2 | h^-)
\end{align}
With this approximation, we could first look up the maximal posterior assignment $z_2 = k_2'$ for $\vec{x}_2$ (or possibly sample it), and then marginalize over $k_1$ as in~\eqref{eq:bigram_approx}.

As explained at the end of Section~\ref{sec:bucktsong_bisampling_assignments}, acoustic and language modelling scores can be at very different scales. We can therefore introduce a language model scaling factor where we raise all the language modelling terms to the power $\eta$, as is done for bigram cluster assignment in~\eqref{eq:bucktsong_bicollapsed2}.

%

\chapter{Expanded segmentation and clustering results}
\makeatletter\@mkboth{}{Appendix}\makeatother
\label{appen:results_zrs}


In Section~\ref{sec:bucktsong_results_own}, several variants of our full-coverage segmentation and clustering approach were considered. In Section~\ref{sec:bucktsong_comparison}, a subset of these were compared to other systems evaluated in the context of the Zero Resource Speech Challenge (ZRS) at Interspeech 2015~\citep{versteegh+etal_interspeech15}, using a subset of the challenge metrics.
Tables~\ref{tbl:appen_zseval_english2} and~\ref{tbl:appen_zseval_tsonga} give the performance of all variants of our system on all the ZRS metrics on the English and Xitsonga data, respectively.
The topline and baseline systems provided as part of the challenge is denoted respectively as ZRSTopline and ZRSBaselineUTD~\citep{versteegh+etal_interspeech15}.
The UTD system of \cite{lyzinski+etal_interspeech15} is denoted as UTDGraphCC.
The original system of \cite{rasanen+etal_interspeech15} is denoted as SyllableSegOsc; they subsequently refined their syllabification method~\citep{rasanen+etal_submission16}, and the system incorporating this updated method is denoted as SyllableSegOsc$^{\text{+}}$.

\begin{table*}[!h]
    \mytable
    \scriptsize
    \caption{Performance of several systems on English2. All scores are given as percentages~(\%). The word boundary detection tolerance is 30~ms or 50\% of a phoneme. Boldface scores indicate the best performing system for that metric, without taking the supervised ZRSTopline into account.
    }
    \begin{tabularx}{\linewidth}{@{}l@{\ }CCCCCCCCCCCCCC@{}}
        \toprule
        & \multicolumn{2}{c}{NLP} & \multicolumn{3}{c}{Grouping} & \multicolumn{3}{c}{Word token} & \multicolumn{3}{c}{Word type} & \multicolumn{3}{c@{}}{Word boundary} \\
        \cmidrule{2-3} \cmidrule(l){4-6} \cmidrule(l){7-9} \cmidrule(l){10-12} \cmidrule(l){13-15}
        Model & NED & Cov. & Prec. & Rec. & $F$ & Prec. & Rec. & $F$ & Prec. & Rec. & $F$ & Prec. & Rec. & $F$  \\
        \midrule
        \multicolumn{15}{@{}l}{\textit{Systems from previous studies:}} \\
        ZRSTopline & 0 & 100 & 99.5 & 100 & 99.7 & 68.2 & 60.8 & 64.3 & 50.3 & 56.2 & 53.1 & 88.4 & 86.7 & 87.5 \\
        ZRSBaselineUTD & \textbf{21.9} & 16.3 & 21.4 & \textbf{84.6} & \textbf{33.3} & 5.5 & 0.4 & 0.8 & 6.2 & 1.9 & 2.9 & 44.1 & 4.7 & 8.6 \\
        UTDGraphCC & 61.2& 80.2& - & - & - & 2.4 & 3.5 & 2.8 & 3.1 & 9.2 & 4.6 & 35.4& 38.5& 36.9 \\
        SyllableSegOsc & 70.8 & 42.4 & 13.4 & 15.7 & 14.2 & 22.6 & 6.1 & 9.6 & 14.1 & 12.9 & 13.5 & 75.7 & 33.7 & 46.7 \\
        SyllableSegOsc$^{\text{+}}$ & 71.1 & 100 & 10.2 & 16.3 & 12.6 & 14.3 & 10.9 & 12.4 & 8.4 & 22.1 & 12.2 & 61.1 & 50.1 & 55.2 \\
        \addlinespace
        \multicolumn{15}{@{}l}{\textit{Speaker-dependent, MFCC embeddings:}} \\
        SyllableBayesClust & 62.2 & 100 & 17.5 & 11.2 & 13.7 & 21.5 & 18.0 & 19.6 & 12.3 & 28.8 & 17.2 & 63.8 & \textbf{59.8} & 61.7 \\
        BayesSeg & 61.5 & 100 & 17.1 & 13.7 & 15.2 & 24.0 &  18.1 & \textbf{20.6} & 13.1 & \textbf{30.1} & 18.2 & 67.3 & 58.3 & \textbf{62.5} \\
        BayesSegMinDur & 56.0 & 100 & 22.7 & 29.6 & 25.5 & 26.6 & 12.5 & 17.0 & 14.0 & 28.6 & \textbf{18.8} & 80.7 & 50.4 & 62.0 \\
        \addlinespace
        \multicolumn{15}{@{}l}{\textit{Speaker-dependent, cAE embeddings:}} \\
        BayesSeg & 62.1 & 100 & 18.0 & 15.0 & 16.3 & 24.8 & 17.0 & 20.2 & 13.3 & 29.1 & 18.3 & 69.4 & 56.3 & 62.2 \\
        BayesSegMinDur & 57.2 & 100 & \textbf{23.7} & 26.3 & 24.9 & \textbf{27.6} & 11.9 & 16.6 & \textbf{14.2} & 26.7 & 18.5 & \textbf{83.1} & 49.0 & 61.6 \\
        \addlinespace
        \multicolumn{15}{@{}l}{\textit{Speaker-independent, MFCC embeddings:}} \\
        SyllableBayesClust & 73.0 & 100 & 9.2 & 5.1 & 6.5 & 21.5 & 18.0 & 19.6 & 12.3 & 28.8 & 17.2 & 63.8 & \textbf{59.8} & 61.7 \\
        BayesSeg & 73.2 & 100 & 9.1 & 5.9 & 7.2 & 23.6 & \textbf{18.2} & \textbf{20.6} & 12.8 & 29.6 & 17.9 & 66.5 & 58.8 & 62.4 \\
        BayesSegMinDur & 72.0 & 100 & 9.9 & 13.0 & 11.2 & 25.9 & 12.6 & 17.0 & 13.7 & 28.9 & 18.6 & 79.7 & 51.4 & 62.1 \\
        \addlinespace
        \multicolumn{15}{@{}l}{\textit{Speaker-independent, cAE embeddings:}} \\
        BayesSeg & 71.1 & 100 & 10.3 & 7.2 & 8.5 & 24.5 & 16.6 & 19.8 & 12.9 & 27.7 & 17.6 & 69.6 & 55.8 & 62.0 \\
        BayesSegMinDur & 66.9 & 100 & 11.9 & 14.0 & 12.8 & 26.9 & 12.2 & 16.7 & 14.1 & 27.5 & 18.6 & 81.7 & 49.6 & 61.7 \\
        \bottomrule
    \end{tabularx}
    \label{tbl:appen_zseval_english2}
\end{table*}

\begin{table*}[!p]
    \mytable
    \scriptsize
    \caption{Performance of several systems on Xitsonga. All scores are given as percentages~(\%). The word boundary detection tolerance is 30~ms or 50\% of a phoneme. Boldface scores indicate the best performing system for that metric, without taking the supervised ZRSTopline into account.
    }
    \begin{tabularx}{\linewidth}{@{}l@{\ }CCCCCCCCCCCCCC@{}}
        \toprule
        & \multicolumn{2}{c}{NLP} & \multicolumn{3}{c}{Grouping} & \multicolumn{3}{c}{Word token} & \multicolumn{3}{c}{Word type} & \multicolumn{3}{c@{}}{Word boundary} \\
        \cmidrule{2-3} \cmidrule(l){4-6} \cmidrule(l){7-9} \cmidrule(l){10-12} \cmidrule(l){13-15}
        Model & NED & Cov. & Prec. & Rec. & $F$ & Prec. & Rec. & $F$ & Prec. & Rec. & $F$ & Prec. & Rec. & $F$  \\
        \midrule
        \multicolumn{15}{@{}l}{\textit{Systems from previous studies:}} \\
        ZRSTopline & 0 & 100 & 100 & 100 & 100 & 34.1 & 49.7 & 40.4 & 15.1 & 18.1 & 16.5 & 66.6 & 91.9 & 77.2 \\
        ZRSBaselineUTD & \textbf{12.0} & 16.2 & \textbf{52.1} & \textbf{77.4} & \textbf{62.2} & 3.2 & 1.4 & 2.0 & 3.2 & 1.4 & 2.0 & 22.3 & 5.6 & 8.9\\
        UTDGraphCC & 43.2 & 89.4 & - & - & - & 2.2 & \textbf{12.6} & 3.8 & \textbf{4.9} & \textbf{18.8} & \textbf{7.8} & 18.8 & \textbf{64.0} & 29.0\\
        SyllableSegOsc & 63.1 & 94.7 & 10.7 & 3.3 & 5.0 & 2.3 & 3.4 & 2.7 & 2.2 & 6.2 & 3.3 & 29.2 & 39.4 & 33.5 \\
        SyllableSegOsc$^{\text{+}}$ & 62.8 & 94.7 & 10.6 & 3.1 & 4.8 & 2.3 & 3.3 & 2.7 & 2.3 & 6.3 & 3.3 & 29.1 & 39.1 & 33.4 \\
        \addlinespace
        \multicolumn{15}{@{}l}{\textit{Speaker-dependent, MFCC embeddings:}} \\
        SyllableBayesClust & 57.7 & 100 & 13.0 & 2.5 & 4.2 & 3.8 & 6.8 & 4.9 & 2.5 & 6.6 & 3.6 & 31.4 & 52.3 & 39.2 \\
        BayesSeg & 56.5 & 100 & 12.7 & 4.1 & 6.2 & 4.1 & 6.2 & 4.9 & 2.9 & 7.8 & 4.2 & 34.5 & 49.0 & 40.5 \\
        BayesSegMinDur & 58.6 & 100 & 8.3 & 10.3 & 9.2 & \textbf{4.3} & 4.0 & 4.1 & 3.8 & 9.8 & 5.5 & 44.5 & 42.0 & 43.2  \\
        \addlinespace
        \multicolumn{15}{@{}l}{\textit{Speaker-dependent, cAE embeddings:}} \\
        BayesSeg & 52.6 & 100 & 16.0 & 5.0 & 7.6 & 4.1 & 5.7 & 4.8 & 3.1 & 8.1 & 4.5 & 36.0 & 47.5 & 41.0 \\
        BayesSegMinDur & 57.0 & 100 & 10.3 & 13.6 & 11.7 & 4.2 & 3.4 & 3.7 & 3.7 & 9.3 & 5.3 & \textbf{47.8} & 40.6 & \textbf{43.9} \\
        \addlinespace
        \multicolumn{15}{@{}l}{\textit{Speaker-independent, MFCC embeddings:}} \\
        SyllableBayesClust & 63.0 & 100 & 8.8 & 3.5 & 5.0 & 3.8 & 6.8 & 4.9 & 2.5 & 6.6 & 3.6 & 31.4 & 52.3 & 39.2 \\
        BayesSeg & 63.6 & 100 & 7.7 & 4.4 & 5.6 & 4.1 & 6.5 & \textbf{5.0} & 2.7  &7.4  &4.0 & 33.5 & 50.0 & 40.1  \\
        BayesSegMinDur & 64.8 & 100 & 4.8 & 8.1 & 6.0 & 3.9 & 3.9 & 3.9 & 3.5 & 9.2 & 5.0 & 42.4 & 42.5 & 42.4 \\
        \addlinespace
        \multicolumn{15}{@{}l}{\textit{Speaker-independent, cAE embeddings:}} \\
        BayesSeg & 55.4 & 100 & 12.6 & 12.8 & 12.7 & 4.2 & 5.3 & 4.7 & 3.1 & 8.1 & 4.5 & 37.6 & 46.2 & 41.5 \\
        BayesSegMinDur & 54.5 & 100 & 9.4 & 21.1 & 13.0 & 4.2 & 3.6 & 3.9 & 3.8 & 9.5 & 5.4 & 46.5 & 41.2 & 43.7 \\
        \bottomrule
    \end{tabularx}
    \label{tbl:appen_zseval_tsonga}
\end{table*}

\end{document}